%% file: arxiv.tex
\documentclass[letterpaper]{article}
\PassOptionsToPackage{hyphens}{url}

\usepackage[T1]{fontenc}
\usepackage{geometry}
\usepackage{setspace}

\usepackage[
  style = chem-acs,
  articletitle = true
]{biblatex}
\usepackage{graphicx}
\graphicspath{{manuscript/}{si/}}
\usepackage{float}
\newfloat{scheme}{htbp}{los}
\floatname{scheme}{Scheme}
\floatname{chart}{Chart}
\newfloat{graph}{htbp}{loh}

\usepackage{chemformula}
\usepackage[version = 4]{mhchem}
\usepackage{amssymb}
\usepackage[
  colorlinks,
  linkcolor=blue,
  citecolor=blue,
  urlcolor=blue,
]{hyperref}
\usepackage{lineno}
\usepackage{setspace}
\usepackage{booktabs}
\usepackage{algpseudocode}
\usepackage{algorithm}
\usepackage[percent]{overpic}
\usepackage{multirow}
\usepackage{docmute}

\usepackage[acronym]{glossaries}
\newacronym{gE}{$g^{\textrm{E}}$}{excess Gibbs energy}
\newacronym{gmix}{$\Delta g^{\textrm{mix}}$}{Gibbs energy of mixing}
\newacronym{ML}{ML}{machine learning}
\newacronym{cem}{CEM}{Convex envelope method}
\newacronym{lle}{LLE}{liquid-liquid equilibrium}
\newacronym{ebms}{EBMs}{Energy-Based Models}

\glsdisablehyper

\usepackage{authblk}
\author[1]{Karim K. {Ben Hicham}}
\author[1]{Moreno Ascani}
\author[1]{Jan G. Rittig}
\author[1,2,3]{Alexander Mitsos\thanks{Corresponding author: amitsos@alum.mit.edu}}

\affil[1]{RWTH Aachen University, Process Systems Engineering (AVT.SVT), 52074 Aachen, Germany}
\affil[2]{Forschungszentrum J\"ulich GmbH, Institute of Climate and Energy Systems ICE-1: Energy Systems Engineering, 52425 J\"ulich, Germany}
\affil[3]{JARA-ENERGY, 52056 Aachen, Germany}

\title{Differentiable Thermodynamic Phase-Equilibria for Machine Learning}
\date{*Email: amitsos@alum.mit.edu}

\makeatletter
\def\input@path{{manuscript/}{si/}}
\makeatother

\newcommand*\arxivcombined{}

\begin{document}

\input{manuscript/main.tex}

\clearpage
\setcounter{section}{0}
\setcounter{subsection}{0}
\setcounter{subsubsection}{0}
\setcounter{figure}{0}
\setcounter{table}{0}
\setcounter{equation}{0}
\setcounter{algorithm}{0}
\setcounter{secnumdepth}{3}
\renewcommand{\thesection}{S\arabic{section}}
\renewcommand{\thesubsection}{S\arabic{section}.\arabic{subsection}}
\renewcommand{\thesubsubsection}{S\arabic{section}.\arabic{subsection}.\arabic{subsubsection}}
\renewcommand{\thefigure}{S\arabic{figure}}
\renewcommand{\thetable}{S\arabic{table}}
\renewcommand{\theequation}{S\arabic{equation}}
\renewcommand{\thealgorithm}{S\arabic{algorithm}}
\renewcommand{\thepage}{S\arabic{page}}
\renewcommand{\theHsection}{supp.\arabic{section}}
\renewcommand{\theHsubsection}{supp.\arabic{section}.\arabic{subsection}}
\renewcommand{\theHsubsubsection}{supp.\arabic{section}.\arabic{subsection}.\arabic{subsubsection}}
\renewcommand{\theHfigure}{supp.\arabic{figure}}
\renewcommand{\theHtable}{supp.\arabic{table}}
\renewcommand{\theHequation}{supp.\arabic{equation}}
\renewcommand{\theHalgorithm}{supp.\arabic{algorithm}}

\input{si/supplement.tex}

\printbibliography

\end{document}

%% file: manuscript/main.tex
\maketitle

\begin{abstract}
    Accurate prediction of phase equilibria remains a central challenge in chemical engineering.
    Physics-consistent machine learning methods that incorporate thermodynamic structure into neural networks have recently shown strong performance for activity-coefficient modeling.
    However, extending such approaches to equilibrium data arising from an extremum principle, such as liquid-liquid equilibria, remains difficult.
    Here we present DISCOMAX, a differentiable algorithm for phase-equilibrium calculation that guarantees thermodynamic consistency at both training and inference, only subject to a user-specified discretization.
    The method combines discrete enumeration of feasible phase states with masked softmax aggregation in the backward pass, with the propagation of the true equilibrium state in the forward pass, using a straight-through gradient estimator to enable physics-consistent end-to-end learning of neural \gls{gE}-models.
    We show that this approach bears analogy to statistical thermodynamics, and we evaluate it on binary liquid-liquid equilibrium data where it outperforms existing surrogate-based methods, while offering a general framework for learning from different kinds of equilibrium data.
\end{abstract}

\section{Introduction}
\label{sec:Intro}
% \linenumbers

Prediction of accurate phase equilibria of multicomponent mixtures is a ubiquitous problem with far-reaching implications in engineering and scientific fields.
Applications of phase equilibrium thermodynamics encompass, e.g., the design of unit operations such as distillation, extraction, crystallization or reactive separations~\cite{redepenning2017pinch,cisternas1993process,skiborowski2022process}, the development of formulations for pharmaceuticals~\cite{barros2023accounting,Sadowski.2025}, the study of geological processes~\cite{paterson1973nonhydrostatic} and even in biology and life science ~\cite{galano2023calculation,Sing.2020,Greinert.2020}.
The prediction of phase equilibria has traditionally relied on \gls{gE} models such as Wilson~\cite{wilson1964vapor}, NRTL~\cite{renonPRAUSNITZLocalCompositionsThermodynamic1968}, and UNIQUAC~\cite{abrams1975statistical}, which are computationally efficient and can model a wide variety of systems, but require extensive parameter fitting to experimental data.
Group-contribution methods such as UNIFAC~\cite{fredenslundPRAUSNITZGroupcontributionEstimationActivity1975} and quantum-chemistry-based approaches such as COSMO-RS~\cite{klamt2000cosmo,klamtKLAMTCOSMOCOSMORSSolvation2011} extend predictive capabilities beyond fitted systems, yet remain limited by applicability, accuracy, and/or computational cost.

More recently, machine-learning models have achieved high predictive accuracy for activity coefficients and mixture nonidealities~\cite{winterBARDOWSmileAllYou2022,medinaSUNDMACHERGraphNeuralNetworks2022,chenSUNDMACHERNeuralRecommenderSystem2021}, but often at the expense of thermodynamic consistency. 
Hybrid and physics-informed ML approaches have therefore been proposed to embed thermodynamic structure directly into learning architectures \cite{dicaprioLEBLEBICIHybridGammaThermodynamicallyConsistent2023,abranchesCOLONActivityCoefficientAcquisition2023,winterBARDOWUnderstandingLanguageMolecules2025,feltonLAPKINMLSAFTMachineLearning2024, alam2026equinet,rittigMITSOSThermodynamicsconsistentGraphNeural2024}, see also our perspective article~\cite{rittig2025molecular}.
After first presenting a soft loss to promote thermodynamic consistency without relying on rigid \gls{gE} models~\cite{rittigMITSOSGibbsDuhemInformed2023}, we introduced a thermodynamically consistent GNN architecture that predicts composition-dependent activity coefficients by introducing a differentiable latent variable for \gls{gE}($x$), and thus enforcing Gibbs-Duhem consistency~\cite{rittigMITSOSThermodynamicsconsistentGraphNeural2024}.
This was further extended by~\cite{spechtJIRASEKHANNAHardconstraintNeural2024} to also ensure \gls{gE} vanishes for pure components.
Enforcing thermodynamic consistency in such a way has been used and discussed in multiple follow-up works by now~\cite{yang2025physics, alam2026equinet}.
When modeling activity coefficients or vapor–liquid equilibria, this $g^{\textrm{E}}$-based architecture is fully sufficient, as the required thermodynamic properties depend only on mole fraction $x^{*}$, the derivatives $\left.\frac{d\Delta g^{\textrm{E}}}{dx}\right|_{x^{*}}$ and $\Delta g^{\textrm{E}}(x^{*})$ itself.
These derivatives can be obtained directly through automatic differentiation, enabling end-to-end training without any additional numerical solvers needed.

However, the situation changes fundamentally when the quantity of interest is itself defined implicitly as the solution of an equation or even as the minimizer of a constrained optimization problem, especially in the presence of many solutions.
In such cases, the learning task naturally assumes the structure of a bilevel optimization problem~\cite{mitsos2009bilevel}, which is formally given by the following equation: 
\begin{equation*}
\begin{aligned}
\min_{\mathbf{\theta}}&\; \mathcal{L} (\mathbf{y^*, \hat{y}^*(\mathbf{\theta})}) \\
\text{s.t.}&\ \
\mathbf{\hat{y}^*(\mathbf{\theta}) \in \arg\min_{y \in \mathcal{Y}}\, }G\mathbf{(y;\mathbf{\theta})}
\end{aligned}
\end{equation*}
where the upper-level objective depends on the solution of a lower-level problem.
Here $\mathcal{L} (\mathbf{y^*}, \hat{\mathbf{y}^*(\mathbf{\theta})})$ denotes the prediction error on the measurements $\mathbf{y^*}$.
The model parameters $\mathbf{\theta}$ affect the thermodynamic model and, as such, the optimal solution to the lower-level problem.
The function $G$ in the lower-level problem typically corresponds to determining the equilibrium state $\mathbf{y}$ of a system by minimizing an appropriate thermodynamic potential, either directly or via specialized reformulations.
The calculation of phase equilibria is a classic problem in computational thermodynamics, and a wide range of methods have been developed for both phase equilibrium calculation and model parameter estimation (see, e.g., \cite{zhang2011review} for a comprehensive overview).
Rigorous approaches, in particular, typically rely on global optimization to ensure that the thermodynamic extremum condition (e.g., global minimum of the Gibbs energy at given $T,p$ and overall composition $\bar{z}$) is satisfied.
These methods are implemented either as large single-level formulations or as specialized formulations~\cite{baker_82_1,michelsen_82_1,michelsen_82_2,mitsosBARTONDualExtremumPrinciple2007}.
Discretization-based methods, such as the Convex Envelope Method (CEM)~\cite{gottlBURGERConvexEnvelopeMethod2025, gottlBURGERConvexEnvelopeMethod2023a}, provide alternative deterministic strategies and operate by constructing a discretized convex envelope of the Gibbs energy landscape over the entire composition space.
Although both continuous optimization formulations and discretization-based convexification methods are, in principle, capable of identifying the thermodynamically correct equilibrium state, the resulting solution mappings are generally not readily differentiable, making gradient computation challenging.
Consequently, integrating such solvers directly into gradient-based learning frameworks is challenging, and training thermodynamically consistent models on phase-equilibrium data, such as equilibrium compositions of multiphase systems, requires solving a lower-level optimization problem for each data point at every training epoch and differentiating through its solution.
For learning model parameters in an end-to-end fashion and using pure component saturation pressures and liquid densities, an approach is presented by~\cite{winterBARDOWUnderstandingLanguageMolecules2025}.
In their approach, the authors maintain differentiability by mapping the neural network outputs to a much simpler parametric model (PC-SAFT implemented in the FeOs calculation framework) to obtain fully converged equilibrium states, and then performing a single Newton update, implemented in PyTorch, on top of the converged solution, which reconstructs the implicit equilibrium dependence in a form compatible with automatic differentiation.
We study the more general case of a fully neural \gls{gE} model, which avoids inheriting the assumptions and limitations of the empirical model. 
The approach by~\cite{winterBARDOWUnderstandingLanguageMolecules2025} is not directly suitable for neural $g^{\mathrm{E}}$ models, because it differentiates through a local solution of a smooth system of implicit equations. However, some thermodynamic properties require finding the global minimum, to ensure physical consistency. Their approach can provide derivatives of a found local solution, but it neither guarantees that this solution is globally stable nor accounts for discontinuous changes when the identity or number of phases changes.
In the field of material science, \cite{guan2022differentiable} gave a perspective on learning the parameters of a thermodynamic potential from phase equilibria and thermochemical properties, only for fitting a single system. 
Similar to~\cite{winterBARDOWUnderstandingLanguageMolecules2025}, an internal iteration step is required to solve for the isopotential condition.

Recently, \cite{hoffmannJIRASEKMachinelearnedExpressionExcess2025} have introduced a \gls{ML} \gls{gE} model that can be trained not only on vapor-liquid equilibrium (VLE) and activity coefficient (AC) data, but also \gls{lle} data.
Their approach follows a sequential strategy: (i) \gls{lle} phase-equilibrium data are generated using a semi-empirical thermodynamic model for binary mixtures over a range of temperatures; (ii) a neural-network surrogate is trained on this pseudo-data to predict equilibrium phase compositions; (iii) the trained surrogate solver is frozen and employed during the training of the machine-learned \gls{gE} model to provide differentiable equilibrium predictions.
While this surrogate-based approach by \cite{hoffmannJIRASEKMachinelearnedExpressionExcess2025} greatly simplifies the problem of learning \gls{lle} behavior, it also introduces two critical limitations in our view.
First, the surrogate solver does not satisfy the following thermodynamic consistency requirements, because it is a learned direct mapping from discretized \gls{gmix} values to phase compositions.
As such, it does not guarantee (i) global optimality of \gls{gmix} and (ii) satisfaction of stability conditions, even in the limit of arbitrarily fine discretization.
Moreover, it always returns a phase split and can be arbitrarily inaccurate for systems out of distribution of the training data.
Secondly, the surrogate training data are, in our opinion, still biased towards the test set because they are generated from the mod. UNIFAC model, whose parameterization was fitted using many of the molecules that appear in the test set.
As a result, prior information from the test domain is implicitly leaked into the surrogate solver, even if the surrogate solver is not directly trained on the test set itself.

In this work, we propose a differentiable multi-phase equilibrium algorithm, which can be integrated into ML model training in an end-to-end fashion.
It ensures thermodynamic consistency by treating the constrained minimization of the \gls{gmix} as a partially masked classification problem, where the \gls{gmix} of a given state is interpreted as an unnormalized score, and the corresponding normalized state weights are obtained via a softmax operator.
The predictions of the solver are thermodynamically consistent up to a user-specified precision, i.e., they guarantee satisfaction of simultaneously: (i) global optimality of \gls{gmix} and (ii) stability conditions.
We do not need any synthetic data in comparison to previous surrogate-based approaches~\cite{dhamankarAcceleratingMulticomponentPhasecoexistence2025, hoffmannJIRASEKMachinelearnedExpressionExcess2025}.
Rather, our approach enables direct learning of the physical relationships, without introducing additional neural network approximation error.
This makes the method significantly simpler and at the same time more general in comparison to previous work.
Nonetheless, the work by~\cite{hoffmannJIRASEKMachinelearnedExpressionExcess2025} indicates that using a trained surrogate neural network in place of a solver works well for binary systems in practice, and we will use it as the main baseline in this work.
We avoid the pitfalls of local optimization methods for this highly non-linear problem by enumerating a discrete grid over the full feasible space.
While this approach scales exponentially with the number of components, it remains practical since most available data involve binary mixtures.
The calculations are implemented with batched tensor operations in PyTorch and can be fully run on the GPU.

We demonstrate our method on a newly generated dataset, which contains 8,597 unique binary \gls{lle} systems, and benchmark our method with the recently proposed surrogate solver by~\cite{hoffmannJIRASEKMachinelearnedExpressionExcess2025}.
Next to the significant qualitative and practical improvements of our algorithm, we show a reduction of approximately 15\% in cross-validation MAE in comparison to the surrogate approach.
\section{Relation to previous work}
\label{sec:RelToPrevWork}
\subsection{Relation to Machine Learning}
In modern deep learning, it is increasingly common to encounter learning problems in which parts of the computational pipeline are inherently non-differentiable~\cite{nesterov2005smooth}.
Typical examples include operations such as selecting a maximizer~\cite{martins2016softmax}, sorting~\cite{blondel2020fast}, sampling elements from a discrete set~\cite{jangCategoricalReparameterizationGumbelSoftmax2017, maddisonConcreteDistributionContinuous2017}, or locating active pixels in an image for a given task~\cite{nibaliNumericalCoordinateRegression2018}.
To address the broad class of discrete optimization problems that arise in these contexts, a substantial body of work in differentiable programming has focused on replacing non-differentiable operations with smooth, continuous relaxations that permit gradient-based optimization.
An introduction to the field is given in \cite{blondel2024elements}.
Almost all the proposed strategies are \emph{probabilistic} in nature and rely on using a neural network score model to predict parameters $\lambda \in \Lambda$ of a distribution function $p_{\lambda}$ and employing this function to reconstruct a continuous output $\hat{y}$ by inference of its expectation or sampling~\cite{blondel2024elements}.
Usually, the employed distribution function $p_{\lambda}$ makes use of a \emph{temperature} parameter $\tau$ that controls the sharpness of the distribution: larger values of $\tau$ yield smoother distributions, while in the limit of $\tau \to 0$ the distribution collapses to the discrete (and non-differentiable) formulation.

Another conceptually related approach is energy-based models (EBMs)~\cite{ackley1985learning, lecun2006ebm}.
These models learn a scalar energy function $E(X,Y)$ that assigns low energy to compatible input-output configurations (e.g., an image and its correct class label) and high energy to incompatible ones~\cite{du2019implicit}.
Prediction is then performed by minimizing this energy, often using approximate optimization or sampling-based inference methods.
\subsection{Relation to Statistical Thermodynamics}
Phase equilibrium thermodynamics is fundamentally probabilistic~\cite{hansen2013theory}: any macroscopic observable $\mathcal{O}$ at equilibrium corresponds to an ensemble average over a large number of accessible microscopic states $i$. Under conditions of constant temperature $T$, this average is given by the Boltzmann distribution in Eq.~\eqref{eq:Boltz_Aver}:
\begin{equation}
    \langle \mathcal{O} \rangle = \sum_i{\mathcal{O}\left(E_i\right)p\left( E_i\right)} =\frac{\sum_i{\mathcal{O}\left(E_i\right)}e^{-\frac{E_i}{k_B T}}}{\sum_j{e^{-\frac{E_j}{k_B T}}}}
\label{eq:Boltz_Aver}
\end{equation}
where $E_i$ denotes the energy of state $i$, $k_B$ is the Boltzmann constant, and $T$ is the temperature. In Eq.~\eqref{eq:Boltz_Prob}, $p\left( E_i\right)$ is the probability that the system occupies the energy state $E_i$:
\begin{equation}
    p\left( E_i\right) =\frac{e^{-\frac{E_i}{k_B T}}}{\sum_j{e^{-\frac{E_j}{k_B T}}}}.
\label{eq:Boltz_Prob}
\end{equation}
In the thermodynamic limit, the resulting probability distribution $p(E_i)\propto \exp(-E_i/k_B T)$ becomes sharply peaked around the equilibrium state, which justifies the classical treatment of thermodynamic equilibrium as the solution of a deterministic minimization problem.
This probabilistic structure coincides with the mathematical formulation of the smooth relaxations of the $\arg\min$ operator over the discrete elements of a vector. 
The softmax function, which rescales the n-elements $\hat{x}_1,...,\hat{x}_n$ of a vector in the range $(0,1)$ and is defined as~\cite{agazzi2025global, nesterov2005smooth}:
\begin{equation}
    \text{softmax}(-\hat{x}_i) =\frac{e^{-\hat{x}_i}}{\sum_j{e^{-\hat{x}_j}}}
\label{eq:softmax_relax}
\end{equation}
Equation~\eqref{eq:softmax_relax} has the same mathematical structure as a Boltzmann-weighted distribution over discrete energy states.
Introducing a temperature-like parameter $\tau$ (which, in the Boltzmann distribution, is equivalent to the $k_BT$ term) controls the sharpness of this distribution and recovers the classical equilibrium solution in the limit $\tau\to 0$, while yielding a regular and differentiable mapping for finite $\tau$.
In our view, the statistical mechanics interpretation of the phase equilibrium problem is complementary to the direct application of relaxing discrete variables in order to obtain meaningful gradients.
In S1 in the SI, we show that the proposed equilibrium solver is mathematically similar to a statistical-thermodynamics ensemble which considers composition fluctuations over a finite set of admissible states at fixed temperature $T$, pressure $p$, and overall composition $z$.
Furthermore, we provide specific case studies using the algorithm to compute pure component vapor pressure as well as a ternary LLLE of an ionic liquid system in S2 in the SI.
\section{Methods}
\label{sec:Methods}
\label{sec:Model}
\begin{figure}[htbp]
    \centering
    \includegraphics[width=1.\linewidth]{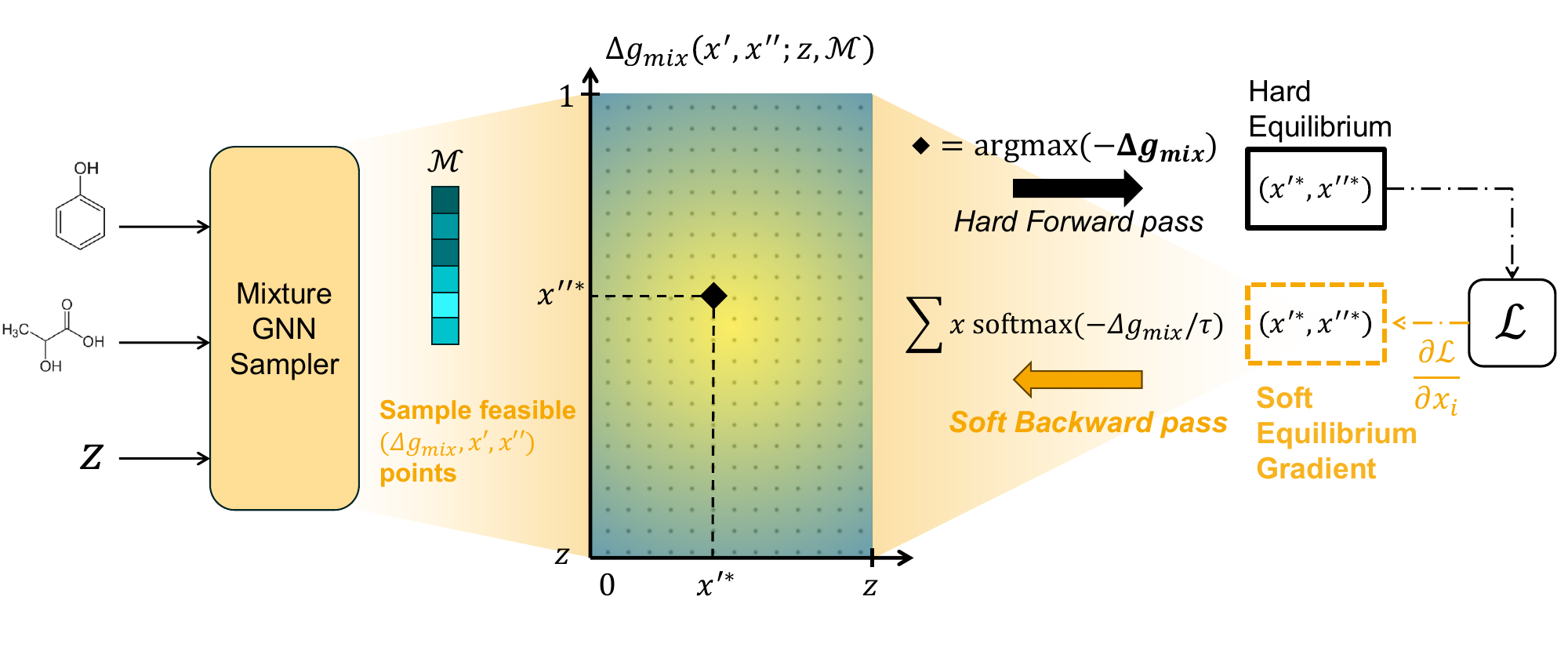}
    \caption{Overview of the proposed end-to-end thermodynamically consistent learning framework. Molecular graphs of the two mixture components are encoded using a graph neural network to form a joint mixture embedding $\mathcal{M}$. The neural model parameterizes the Gibbs energy of mixing $\Delta g^{\mathrm{mix}}(x;\mathcal{M})$, which is evaluated on a discretized composition grid subject to the feed composition $z$. Feasible phase-split candidates $(x',x'')$ are identified via mass-balance constraints and combined to compute the total Gibbs energy of the mixture. In the forward pass, the equilibrium compositions are obtained via a hard minimization of $\Delta g^{\mathrm{mix}}$. During training, gradients are propagated through a Boltzmann-weighted softmax relaxation using a straight-through estimator, enabling differentiable, thermodynamically consistent learning.}

    \label{fig:model_overview}
\end{figure}

The following section presents our end-to-end framework for thermodynamically consistent phase-equilibrium prediction, detailing both the neural \gls{gmix} model and the proposed differentiable equilibrium solver.
The overall approach is illustrated in Figure~\ref{fig:model_overview}.
\subsection{Neural Parameterization of the Gibbs Energy of Mixing}
\label{sec:NeurPar}
Our model predicts \gls{lle} compositions by learning a
parametric representation of the Gibbs Energy of mixing $\Delta g^{\textrm{mix}}$ for binary
liquid systems.
Each molecule pair ($mol_1$ and $mol_2$), is encoded using a
graph neural network (GNN), which maps the corresponding molecular graphs to
continuous embeddings,
\begin{equation*}
    \mathbf{h}_1 = \mathrm{GNN}(mol_1; \mathbf{\theta}), \qquad 
    \mathbf{h}_2 = \mathrm{GNN}(mol_2; \mathbf{\theta}).
\end{equation*}
While any encoder, such as simple fingerprints or transformers, would also work, we use the standard \textit{Chemprop} D-MPNN, which performs message passing over directed bond representations rather than directly over atom representations~\cite{heid2023chemprop}.  
We employ the simplest mixture embedding in which the pure component embeddings $\mathbf{h}_1, \mathbf{h}_2$ are concatenated to form a mixture embedding $\mathcal{M}$ and passed to a neural network $\widetilde{g}^\textrm{E}(\cdot\,;\mathbf{\theta})$, which models the excess Gibbs energy of mixing at
composition $x$ before scaling with $x(1-x)$.
The concatenation uses a fixed component order and therefore does not enforce permutation equivariance with respect to component order.
For real applications in which the two components are exchangeable, an explicitly order-invariant architecture can be preferable, as in \cite{qinM.ZAVALACapturingMolecularInteractions2023, rittigMITSOSThermodynamicsconsistentGraphNeural2024, hoffmannJIRASEKMachinelearnedExpressionExcess2025}.
As we focus exclusively on isothermal binary mixtures at fixed pressure, the only relevant state variable is the composition $x\in[0,1]$.
To ensure thermodynamic consistency at the pure-component boundaries, we multiply the network output by the product of the liquid-phase compositions, such that
\begin{equation}
\label{eq:Gibbs_with_constraints}
\Delta g^{\textrm{mix}}(x;\mathbf{\theta}) = x(1-x)\,\widetilde{g}^{\textrm{E}}(x; \mathcal{M},\mathbf{\theta}) + \Delta g^{\textrm{id}}(x),
\end{equation}
where $\widetilde{g}^{\textrm{E}}$ is the unrestricted network prediction and $\Delta g^{\textrm{id}}(x)=x\ln(x)+(1-x)\ln(1-x)$ is the ideal Gibbs energy of mixing. Equation~\eqref{eq:Gibbs_with_constraints} thus defines the physics-consistent prediction of the Gibbs energy of mixing.
Both ideal and excess Gibbs energies are scaled by $\frac{1}{RT}$, which is constant for our isothermal setting and thus omitted for notational clarity.
The term $x(1-x)$ guarantees $\Delta g^{\textrm{mix}}(0)=\Delta g^{\textrm{mix}}(1)=0$, as required for binary mixtures~\cite{margules1895composition,guan2022differentiable, spechtJIRASEKHANNAHardconstraintNeural2024}.
The composition-dependent prediction module uses smooth ELU activations, so the learned \gls{gmix} surface is smooth with respect to $x$.
Non-smooth ReLU activations in the molecular encoder act only on the fixed molecular graph representation and do not affect smoothness with respect to composition.
All composition derivatives and Hessians used during model training are computed by automatic differentiation.
In the definition of $\Delta g^{\textrm{mix}}(x;\mathbf{\theta})$, the dependence of the mixture embedding $\mathcal{M}$ is omitted for notational clarity.
\subsection{Equilibrium Prediction via Discretized Gibbs Energy Minimization}
\label{sec:EquilPred}
\begin{algorithm}[htpb]
\caption{DISCOMAX: Differentiable Binary Equilibrium Solver with DISCRETE enumeration (fp) and masked SOFTMAX aggregation over feasible states (bp).}
\begin{algorithmic}[1]
\Require Feed composition $z \in (0,1)$, Gibbs energy model $\Delta g^{\textrm{mix}}(x;\mathbf{\theta})$, base discretized grid vector $\mathbf{x}=(x^{(k)})_{k=1}^{N}$, temperature parameter $\tau > 0$
\Ensure ST Prediction of Equilibrium Phase Compositions $(x^{\prime*}, x^{\prime\prime*})$,
\State Define an augmented grid by appending the feed composition.
\[
\mathbf{x}^{\mathrm{aug}}=(x^{(i)})_{i=1}^{N'}=(\mathbf{x},z), \quad N'=N+1.
\]
\State Compute the vector of Gibbs energies on the grid:
\[
\mathbf{\Delta g^{\textrm{mix}}} \in \mathbb{R}^{N'}, \quad \Delta g^{\textrm{mix},i} = \Delta g^{\textrm{mix}}(x^{(i)};\mathbf{\theta}).
\]
\State For each pair of grid indices $(i,j)$ with $1 \le i \le j \le N'$, compute
\[
\Delta g^{\textrm{mix},(i,j)} =
\begin{cases}
    +\infty & \text{if } z < x^{(i)} \text{ or } z > x^{(j)} \\
    \frac{1}{2}\left(\Delta g^{\textrm{mix},(i)}+\Delta g^{\textrm{mix},(j)}\right) - \epsilon
    & \text{if } |x^{(i)}-x^{(j)}|<\epsilon \\
    \Delta g^{\textrm{mix},(i)}\tfrac{z - x^{(j)}}{x^{(i)} - x^{(j)}} 
    + \Delta g^{\textrm{mix},(j)}\tfrac{z - x^{(i)}}{x^{(j)} - x^{(i)}} & \text{otherwise}
\end{cases}
\]
\State Determine hard minimum indices for inference (not differentiable).
\[
(i^*, j^*) \in \arg\min_{1 \le i \le j \le N'} \Delta g^{\textrm{mix},(i,j)}.
\]
\State Set $x^{\prime*}=x^{(i^*)}$, $x^{\prime\prime*}=x^{(j^*)}$.
\State Compute softmax weights:
\[
p^{(i,j)} =
\frac{\exp(-\Delta g^{\textrm{mix},(i,j)}/\tau)}
{\sum_{u=1}^{N'} \sum_{v=u}^{N'} \exp(-\Delta g^{\textrm{mix},(u,v)}/\tau)}.
\]
\State Compute soft estimates for training (differentiable):
\[
x^{\prime*}_{\mathrm{soft}} = \sum_{i=1}^{N'} \sum_{j=i}^{N'} p^{(i,j)}\,x^{(i)}, \quad
x^{\prime\prime*}_{\mathrm{soft}} = \sum_{i=1}^{N'} \sum_{j=i}^{N'} p^{(i,j)}\,x^{(j)}.
\]
\State Straight-Through Gradient Estimation:
\[
x^{\prime*} \leftarrow x^{\prime*} + \left( x^{\prime*}_{\mathrm{soft}} - \operatorname{stopgrad}(x^{\prime*}_{\mathrm{soft}}) \right)
\]
\[
x^{\prime\prime*} \leftarrow x^{\prime\prime*} + \left( x^{\prime\prime*}_{\mathrm{soft}} - \operatorname{stopgrad}(x^{\prime\prime*}_{\mathrm{soft}}) \right)
\]
\State \Return $(x^{\prime*}, x^{\prime\prime*})$
\end{algorithmic}
\label{alg:diff_flash_indexed}
\end{algorithm}
The condition of thermodynamic equilibrium (number and composition of the coexisting phases) of a liquid system at given temperature $T$ and pressure $p$ is dictated by the global minimum of $\Delta g^{\textrm{mix}}$ subject to mass balance for the given overall composition $z$.
For a binary system, the two equilibrium compositions $(x^{\prime*},x^{\prime\prime*})$ satisfy the following equation:
\begin{equation}
\label{eq:actual_min}
(x^{\prime*},x^{\prime\prime*}) \in 
\arg\min_{x',x''}
\left\{
\Delta g^{\textrm{mix}}(x';\mathbf{\theta})\phi'+
\Delta g^{\textrm{mix}}(x'';\mathbf{\theta})\phi''
\right\},
\end{equation}
where $x'$ and $x''$ denote phase composition in phases one and two, respectively, and the molar fractions $\phi',\phi''$ of both phases are given by Eq.~\eqref{eq:phaseamount}.
\begin{equation}
\label{eq:phaseamount}
\phi'=\frac{x''-z}{x''-x'};
\quad
\phi''=\frac{z-x'}{x''-x'}.
\end{equation}
In the continuous setting, the equilibrium problem in Eq.~\eqref{eq:actual_min} constitutes a nonconvex constrained optimization problem, since $\Delta g^{\textrm{mix}}(x;\boldsymbol{\theta})$ may exhibit multiple local minima corresponding to metastable phase states.
Computing the equilibrium state, therefore, requires identifying the global minimizer, which is possible (see e.g.,~\cite{mitsosBARTONDualExtremumPrinciple2007}), but computationally expensive and difficult to integrate into the training.
While local optimization methods could in principle be employed, this might not be without problems: differentiating through a local solver requires either implicit differentiation or unrolled optimization, both of which introduce additional implementation complexity,  computational overhead, and additional hyperparameters, i.e., starting values and number of iterations.
More importantly, local solvers may converge only to stationary points rather than the global optimum, and their convergence behavior can change abruptly under small perturbations of $\boldsymbol{\theta}$.
This will yield incorrect equilibrium predictions and thus wrong and strongly varying gradients, which might destabilize training.

As an alternative, we present the forward pass of DISCOMAX, a discrete optimization procedure which works by enumerating the set of mass-balance feasible candidate equilibrium states and selecting the globally optimal state ($\arg\min$) from that set.

Differentiation of the elements in the set of Eq.~\eqref{eq:actual_min} with respect to the model parameters $\mathbf{\theta}$ does not provide meaningful gradients directly, since the argmin/argmax over a discrete set is a piecewise-constant function~\cite{blondel2020fast}.
We therefore approximate the solution using a relaxation of the
discretized Gibbs energy surface.

Let $\mathbf{x}=(x^{(k)})_{k=1}^{N}$ be a discretized grid vector on $(0,1)$, where $N$ is the number of discretization points, and let $\mathbf{x}^{\mathrm{aug}}=(x^{(i)})_{i=1}^{N'}=(\mathbf{x},z)$ be the augmented grid vector obtained by appending the feed composition $z$, and thus containing $N' = N+1$ entries.
We compute $\Delta g^{\textrm{mix}}$ on this grid, enumerate all feasible pairs $(i,j)$ with $i\le j$ and $x^{(i)} \le z \le x^{(j)}$ (i.e., unique binary phase combinations satisfying the mass balance with non-negative phase amounts), and evaluate the total Gibbs energy of mixing of each pair according to $\Delta g^{\textrm{mix},(i,j)}=\Delta g^{\textrm{mix},(i)}\phi^{(i)}+\Delta g^{\textrm{mix},(j)}\phi^{(j)}$.
This enumeration includes the self-pair $(z,z)$, which represents the quasi-homogeneous single-phase state at the feed composition $x'=x''=z$.
This self-pair is assigned a small energy offset $-\epsilon$ to break exact ties in favor of the quasi-homogeneous state instead of any the possible infinitesimal two-phase splits which have the same \gls{gmix}, i.e., all combinations where $x' = z$ or $x'' = z$.
In the case that the feed self-pair is the global minimum, the mixture can be considered single-phase.
The hard (non-differentiable) equilibrium estimate is obtained as
\begin{equation*}
(i^*,j^*) \in \arg\min_{i\le j} \Delta g^{\textrm{mix},(i,j)},
\qquad
x^{\prime*} = x^{(i^*)},\quad
x^{\prime\prime*} = x^{(j^*)}.
\end{equation*}
\subsection{Differentiable Relaxation via Straight-Through Softmax}
\label{sec:DiffRel}
The equilibrium compositions are defined by a discrete minimization of a vector of total Gibbs energy of mixing values.
While the hard $\arg\min$ satisfies the thermodynamic extremum principle exactly, it is non-differentiable and therefore cannot be used directly in gradient-based training.
We address this by adopting a commonly used straight-through (ST) gradient estimation~\cite{bengio2013estimating, maddisonConcreteDistributionContinuous2017, jangCategoricalReparameterizationGumbelSoftmax2017}, which decouples the \emph{forward} prediction from the \emph{backward} gradient computation. 
The authors of these works use it, to retain discrete decisions in the forward pass while propagating gradients through a differentiable relaxation during backpropagation.
\begin{itemize}
    \item \textbf{Forward pass (fp)} (thermodynamically exact, up to discretization).
    We compute the hard minimizer
    \[
    (i^*,j^*) \in \arg\min_{1\le i \le j \le N'} \Delta g^{\textrm{mix},(i,j)},
    \qquad
    x^{\prime*} = x^{(i^*)},\; x^{\prime\prime*} = x^{(j^*)},
    \]
    and return $(x^{\prime*},x^{\prime\prime*})$. This guarantees that the reported equilibrium state satisfies the extremum principle and the mass-balance feasibility constraints encoded in the
    masking of infeasible pairs.
    \item \textbf{Backward pass (bp)} (smooth surrogate gradients).
    At the same time, we form a differentiable softmax distribution over all candidate pairs,
    \[
    p^{(i,j)} =
    \frac{\exp\!\left(-\Delta g^{\textrm{mix},(i,j)}/\tau\right)}
    {\sum_{u=1}^{N'}\sum_{v=u}^{N'} \exp\!\left(-\Delta g^{\textrm{mix},(u,v)}/\tau\right)},
    \]
    and compute soft equilibrium estimates
    \[
    x^{\prime*}_{\mathrm{soft}} = \sum_{i=1}^{N'}\sum_{j=i}^{N'} p^{(i,j)}x^{(i)},\qquad
    x^{\prime\prime*}_{\mathrm{soft}} = \sum_{i=1}^{N'}\sum_{j=i}^{N'} p^{(i,j)}x^{(j)}.
    \]
    Gradients are then propagated through these soft estimates using a straight-through update,
    \[
    x^{\prime*} \leftarrow x^{\prime*} + \Bigl(x^{\prime*}_{\mathrm{soft}} - \operatorname{stopgrad}(x^{\prime*}_{\mathrm{soft}})\Bigr),\qquad
    x^{\prime\prime*} \leftarrow x^{\prime\prime*} + \Bigl(x^{\prime\prime*}_{\mathrm{soft}} - \operatorname{stopgrad}(x^{\prime\prime*}_{\mathrm{soft}})\Bigr),
    \]
    such that the forward value remains the hard minimizer, but the backward derivative follows
    the differentiable soft surrogate.
\end{itemize}
Concretely, we replace the gradients of $(x',x'')$ with those of $(x_{\mathrm{,soft}}',x_{\mathrm{,soft}}'')$ by invoking \verb|tensor.detach()|.
The temperature $\tau$ controls the sharpness of the soft distribution: as $\tau \to 0$, $p^{(i,j)}$ concentrates on the hard minimizer, recovering the discrete optimum, whereas larger $\tau$ yields smoother gradients that can improve optimization early in training~\cite{jangCategoricalReparameterizationGumbelSoftmax2017, maddisonConcreteDistributionContinuous2017}.

Algorithm~\ref{alg:diff_flash_indexed} defines the full DISCOMAX solver: \emph{discrete} enumeration and optimization in the forward pass and masked \emph{softmax} aggregation over feasible states in the backward pass.
We thus obtain a thermodynamically-consistent binary equilibrium solver formulation that can be integrated with ease into any neural network architecture and thus facilitates full end-to-end training, without needing supplementary data or training steps, in contrast to previous work.
Furthermore, it recovers the thermodynamically exact, globally optimal solution in the limit of infinitely fine discretization.
\subsection{Baseline: Surrogate Solver}
\label{sec:SurrSolv}
As discussed in the Introduction, using a neural network as a surrogate for a solver~\cite{hoffmannJIRASEKMachinelearnedExpressionExcess2025} has several structural limitations, but it remains an important and currently the only directly applicable baseline for this work from the literature.
Because neither the implementation nor the training code is publicly available, we re-implemented a similar surrogate solver and benchmark both with a GNN architecture.
We do not resolve the issues of data leakage mentioned in the Introduction. Moreover, we train the surrogate solver on \emph{all} available UNIFAC molecules in our dataset, which should yield the most optimistic performance of the baseline method.
More details regarding the baseline implementation are given in the SI, and our GitLab repository: \url{https://git.rwth-aachen.de/avt-svt/public/differentiable-thermo-eq-ml}.

Throughout this work we generally refer to the equilibrium prediction module $(x', x'') = f(\mathbf{\Delta g_{\mathrm{mix}}}, (z))$ with the word $\textbf{solver}$ and to the full prediction model $(x', x'') = f(\text{mol}_1, \text{mol}_2, (z))$, with the word $\textbf{model}$.
\subsection*{Loss Function Definitions}
\label{sec:LossFuncDef}
The performance of both the surrogate solver and the DISCOMAX solver depends on the choice of additional loss functions.
We propose a novel non-masked Hessian loss and compare it with the masked ``Gibbs loss'' presented by~\cite{hoffmannJIRASEKMachinelearnedExpressionExcess2025}.
The total loss per data point, including direct and auxiliary losses, can be written as:
\begin{align*}
\mathcal{L}^{k}_{\text{total}} &= 
m^{k}_{G} \times (\mathcal{L}^{k}_{1} + \mathcal{L}^{k}_{2}) +
\lambda_{G}\mathcal{L}^{k}_{G} +
\lambda_{H}\mathcal{L}^{k}_{H}
\end{align*}
where the first two loss terms are standard mean-squared error losses to penalize incorrect predictions directly through the differentiable prediction:
\begin{align*}
\mathcal{L}_{1} &= \| \hat{x}' - x'^{\ast} \|_2^2, \\
\mathcal{L}_{2} &= \| \hat{x}'' - x''^{\ast} \|_2^2.
\end{align*}
Here, $\hat{x}'$ and $\hat{x}''$ denote the differentiable phase-composition predictions used during training. For DISCOMAX, they are the straight-through solver outputs of the differentiable relaxation described above, i.e., hard forward-pass equilibrium compositions with gradients taken from the soft relaxation. For the surrogate baseline, they are the direct neural-network predictions obtained from the discretized \gls{gmix} curve. The targets $x'^{\ast}$ and $x''^{\ast}$ denote the target phase compositions extracted from the synthetic HANNA 2 dataset by discrete Gibbs-energy minimization, as described in the Dataset subsection. The direct prediction loss can be selectively masked by $m^{k}_{G}$, based on conditions explained below. In this work, the mask is only applied when the Gibbs loss is also used, and otherwise set to 1.
The surrogate-only variant without any auxiliary loss is not considered, since the masking strictly requires an alternative loss if the necessary conditions for a phase split are not satisfied. These are never satisfied in practice, because the ideal mixing term dominates the \gls{gmix} prediction, early in training. 

\subsection*{Auxiliary Hessian Loss (H)}\label{hess_loss}
The Hessian loss term promotes thermodynamically consistent stability behavior at the measured data points, specifically, convexity at the target phase compositions $( x^{\prime*}, x^{\prime\prime*})$ and concavity at the feed composition $(z)$.
Convexity at $x^{\prime*}$ and $ x^{\prime\prime*}$ is a necessary, but not sufficient condition for a consistent and correctly predicted equilibrium.
Concavity at $z$ is neither necessary nor sufficient, but we assumed that it might improve training, since at random initialization in the first epoch, \gls{gmix} of each system is convex everywhere as the ideal term dominates. 
We discuss the need for this auxiliary loss in the results. 
\begin{equation*}
\mathbf{H}_{\Delta g_{\mathrm{mix}}}(x^{\prime*}), \mathbf{H}_{\Delta g_{\mathrm{mix}}}(x^{\prime\prime*}) \succ 0 \quad \text{(convex, stable phases)}, 
\qquad
\mathbf{H}_{\Delta g_{\mathrm{mix}}}(z) \prec 0 \quad \text{(concave, unstable mixture)}.
\end{equation*}
It is defined as
\begin{align*}
\mathcal{L}_{\text{H}} &=
\mathcal{L}_{\text{convex}}(x^{\prime*})
+ \mathcal{L}_{\text{convex}}(x^{\prime\prime*})
+ \mathcal{L}_{\text{concave}}(z),
\end{align*}
where each curvature loss penalizes deviations from the desired sign of the Hessian diagonals:
\begin{align*}
\mathcal{L}_{\text{convex}}(x) &=
\frac{1}{N} \sum_i 
\begin{cases}
0, & \text{if } (\mathbf{H}_{\Delta g_{\mathrm{mix}}}(x))_{ii} \ge \epsilon, \\[6pt]
\epsilon - (\mathbf{H}_{\Delta g_{\mathrm{mix}}}(x))_{ii}, & \text{if } (\mathbf{H}_{\Delta g_{\mathrm{mix}}}(x))_{ii} < \epsilon,
\end{cases} \\[8pt]
\mathcal{L}_{\text{concave}}(x) &=
\frac{1}{N} \sum_i 
\begin{cases}
0, & \text{if } (\mathbf{H}_{\Delta g_{\mathrm{mix}}}(x))_{ii} \le -\epsilon, \\[6pt]
(\mathbf{H}_{\Delta g_{\mathrm{mix}}}(x))_{ii} + \epsilon, & \text{if } (\mathbf{H}_{\Delta g_{\mathrm{mix}}}(x))_{ii} > -\epsilon,
\end{cases}
\end{align*}
with a small margin \(\epsilon > 0\) to define a constraint back-off, in order to avoid oscillations during training.
\subsection*{Auxiliary masked Gibbs Loss (G)}\label{gibbs_loss}
The masked Gibbs loss term introduced by~\cite{hoffmannJIRASEKMachinelearnedExpressionExcess2025} promotes the correct qualitative curvature of the molar Gibbs energy of mixing 
\(\Delta g_{\mathrm{mix}}(T, x)\) over the composition domain for data points known to exhibit an \gls{lle}. 
In particular, it penalizes cases where the model fails to predict a concave region in \gls{gmix}, which is a weak but necessary condition for the presence of the correct miscibility gap.
At the same time, it controls whether the surrogate solver loss is applied via a mask $m^{k}_{G} \in \{0,1\}$~\cite{hoffmannJIRASEKMachinelearnedExpressionExcess2025}.
In our experiments, we did not find the mask $m^{k}_{G}$ to be necessary and even potentially harmful, but we stick with the masked version to ensure consistency with the literature baseline.
Comparison between the masked and unmasked versions is given in the SI.
For each \gls{lle} sample \(k\), the curvature metric
\begin{equation*}
S_k(x_d)
= 
\frac{\partial^2 \Delta g_{\mathrm{mix}}(x_d)}{\partial x^2}
\end{equation*}
is evaluated on a discretized composition grid 
\(\{x_d\}_{d=1}^{101}\).   
The minimum curvature value over the grid determines whether the model predicts a phase split.
The Gibbs loss is defined as
\begin{align*}
\mathcal{L}_{\text{Gibbs}}
&=
\max\Bigl( 0, \; \min_d S_k(x_d) \Bigr),
\end{align*}
where negative minimum curvature indicates a concave region and thus a correctly detected miscibility gap.
In the binary case, the concavity part of the Hessian loss and the Gibbs loss can be similar because both depend on the curvature of \gls{gmix}.
They are not redundant, however.
The Hessian concavity term is evaluated specifically at the feed composition $z$ and therefore promotes local instability at the desired split location, whereas the Gibbs loss only requires a negative curvature at any composition.
The Gibbs loss can therefore be satisfied by a concave region outside of the miscibility gap, which is undesirable.
In our case studies, we only consider isothermal \gls{lle}, but the extension is straightforward.
\section{Computational Experiments}
\label{sec:Experim}
We evaluate our proposed differentiable equilibrium solver and compare it with the surrogate solver in
two complementary computational test cases.
The first test case isolates the optimization behavior of the solvers on a single binary mixture, allowing us to assess whether each method can recover the phase equilibrium for a fixed system, with no learned dependence on the molecular structure.
The second test case extends the comparison to a chemically diverse dataset of binary mixtures, with the aim of evaluating generalization performance when molecular structure information is included.
\subsection{Dataset}
\label{sec:Dataset}
\begin{figure}[htbp]
    \centering
    \begin{minipage}{0.31\textwidth}
        \centering
        \includegraphics[width=\textwidth]{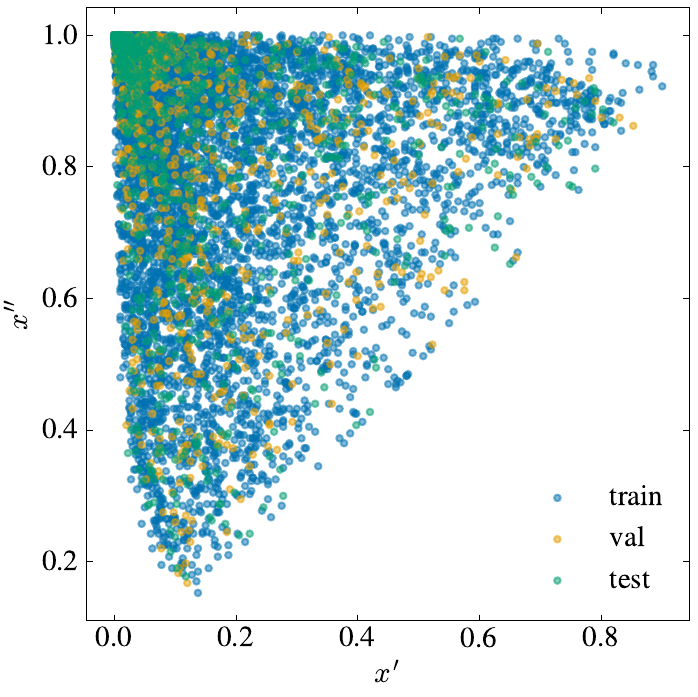}
    \end{minipage}
    \hfill
    \begin{minipage}{0.31\textwidth}
        \centering
        \includegraphics[width=\textwidth]{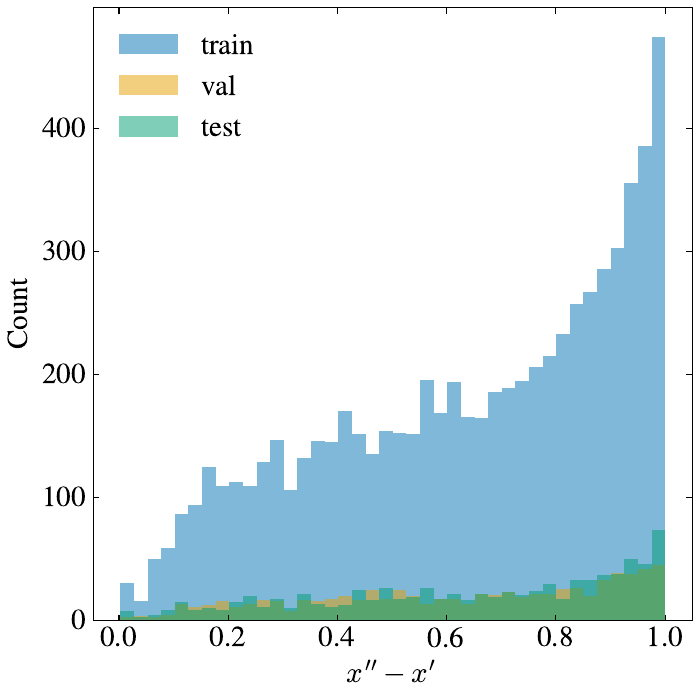}
    \end{minipage}
    \hfill
    \begin{minipage}{0.31\textwidth}
        \centering
        \includegraphics[width=\textwidth]{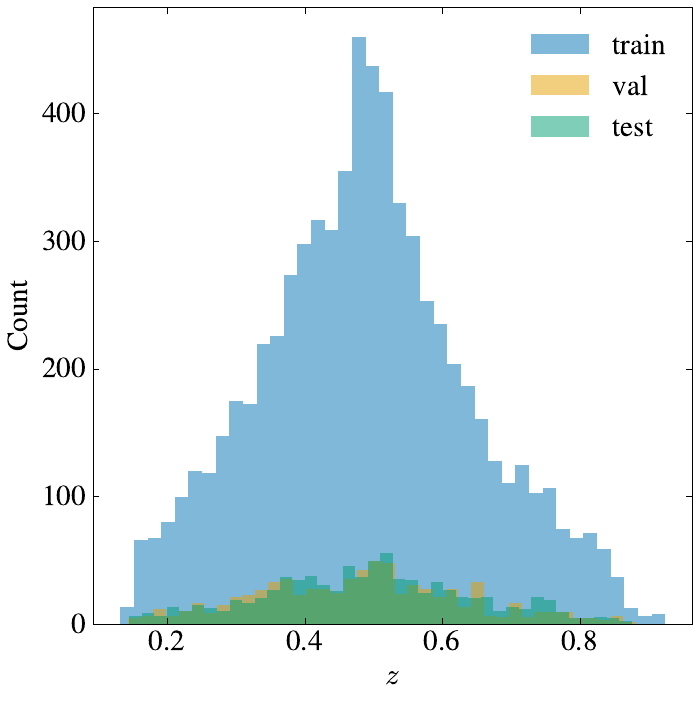}
    \end{minipage}
     \caption{Scatter plot of equilibrium composition pairs (left). Histogram of the miscibility gap width (middle). Histogram of the feed composition $z$ (right). Colored by train, validation, and test split for the first fold.}
     \label{fig:dataset}
\end{figure}

We create a comprehensive binary \gls{lle} dataset based on the Dortmund Database trained HANNA 2 model~\cite{hoffmannJIRASEKMachinelearnedExpressionExcess2025}.
We do this by forming all possible binary mixtures of 526 unique solvents, resulting in 138,601 candidate systems.
This is the subset of solvents first published by~\cite{qinM.ZAVALACapturingMolecularInteractions2023}, which were also found in the COSMObase2021.
From this pool, we extract only those mixtures that exhibit a \gls{lle} at 298.15 K.
This yields 8,597 unique binary systems with two liquid phases, which we then use as the ground truth data set in this work.
The ground truth equilibrium data for each system is obtained by evaluating the HANNA 2 model on a 401-point grid of compositions, estimating a point inside the LLE via curvature, forming all mass-balance feasible phase combinations, and searching for the lowest \gls{gmix} pair.
Throughout this work, we use the term ``LLE system'' as shorthand for a mixture whose isothermal HANNA 2 \gls{gmix} curve exhibits a two-phase split under this discrete construction. This negative-curvature and Gibbs-minimization criterion does not prove that the corresponding real physical mixture would necessarily exhibit experimentally observable \gls{lle}.
Depending on additional properties such as vapor pressure or melting behavior, the same chemistry could also display, e.g., VLE or SLE, which should be taken into account in future work.
The published dataset should therefore be interpreted as a synthetic benchmark.
Details on the dataset generation are given in the SI, and the full code to generate the dataset is available in our GitLab repository.
Dataset statistics for the first fold are visualized in Figure~\ref{fig:dataset}. 
Each data point consists of features: $mol_1, mol_2, z$ and labels: $x^{\prime*}, x^{\prime\prime*}$. 
For training, the feed $z$ is chosen to be in the center of the miscibility gap, i.e. $z = 0.5*(x^{\prime*} + x^{\prime\prime*})$.
This construction uses complete tie-line data, for which both phase compositions are available.
If only one phase composition is measured, but any overall composition inside the immiscible region is known or can be estimated, DISCOMAX can still be used.
This works by predicting both phases and evaluating the loss only on the observed target phase composition.
Since estimating a composition inside the miscibility gap, and determining which side of the tie-line the measured phase belongs to (also needed for the surrogate approach) can be challenging, it is not covered here and left for future work. 
It is important to note that the feed composition is not passed to the neural \gls{gE} model and, as such, can also not leak information.
It is only used in the DISCOMAX solver module, which does not contain any trainable parameters, and is used to define the mass balance and feasible candidates.
For a predicted binary \gls{gmix} curve with a miscibility gap, any feed inside the same gap yields, by construction, the same equilibrium phase compositions in the forward pass.
This also means that a system with multiple disconnected miscibility gaps can be learned by providing corresponding feed compositions and target phase compositions as separate data points in training.
Initially, we ran all experiments with the equivalent COSMO-RS-generated dataset, but we opted to use HANNA 2 (MIT License) in order to be able to publish the full dataset and enable full reproducibility of our results.
For the same reason, experimental equilibrium data from NIST TDE or Dortmund Database were not used in this work, since it is not publicly available in their entirety and thus cannot be used for reproducible benchmarking.
Since the model used for generating the HANNA 2 dataset was trained using the surrogate solver, there may be some bias in favor of the surrogate solver, the main baseline of this work, versus our differentiable solver.
We report key results of our method for the COSMO-RS dataset in the SI.
We publish the full dataset as well as the code used to compute the full dataset using the HANNA 2 model.
\subsection{Fitting single systems}
\label{sec:single_system}
In the first test case, we consider a single binary mixture and train a system-specific neural \gls{gE} model to reproduce its \gls{lle} behavior. 
The purpose of this setup is to probe the optimization properties of the equilibrium solvers in isolation, without the confounding effects of molecular feature extraction.
In particular, this test examines whether a solver can drive the model towards exact overfitting of a known \gls{lle} system and whether the optimization converges to thermodynamically consistent solutions.
Since performance depends only on the target equilibrium compositions, we group the available systems into 50 bins based on the size of their miscibility gap between zero and one. One system is selected from each bin, ensuring that the evaluation covers the full range of miscibility behaviors and avoids bias toward any particular miscibility gap width.

The neural \gls{gE} model used in this test case is not conditioned on molecular structure and takes only the scalar composition $x$ as input.
It consists of three hidden layers with 64 units each and also ELU activations, to ensure smoothness of the predicted \gls{gmix} curve.
The model is trained for 200 epochs using the AdamW optimizer in combination with a OneCycle learning rate schedule with cosine annealing and a maximum learning rate of $0.001$.
The softmax temperature $\tau$ is initialized with $0.1$ and multiplied by a factor of $0.98$ each epoch.
Early stopping is intentionally omitted, as the goal of this test is to assess the ability of each solver to achieve exact fitting.
The weights in the loss function were chosen to be $\lambda_G=0.01$ and $\lambda_H=0.05$.
Based on a small grid search, we chose a single learning rate based on the best Surrogate G solver configuration, the choice for our DISCOMAX H was largely robust to the different learning rates tested (MAE $\le 0.025$) and was hence chosen freely; details are provided in the SI.
We chose 101 discretization points for both the DISCOMAX and surrogate solvers in line with \gls{lle} measurement accuracy and to keep comparability to the surrogate solver literature baseline.

\subsection{Generalizing to new systems}
\label{sec:full_dataset}
The second test case evaluates solver performance in a practical learning scenario, where the model must generalize across chemically distinct binary mixtures.
This setup assesses whether the solvers enable stable end-to-end training of a molecular-structure informed \gls{gE} model and whether the learned representations transfer to previously unseen systems.
We focus our analysis on the comparison between DISCOMAX, DISCOMAX with Hessian loss (DISCOMAX H), and the surrogate solver with Gibbs loss (Surrogate G).

For this case study, we train a neural \gls{gE} model with ten-fold cross-validation, which is conditioned on both molecular graphs and the mixture composition. The encoder is a default Chemprop MPNN, which produces a 300-dimensional molecular embedding for each component, which is then concatenated together with the composition and passed through an ELU-activated feedforward network.
The dataset is split into training, validation, and test subsets (80/10/10) comprising different binary mixtures, ensuring that all test systems are unseen during training, i.e., a system-wise split.
We use the AdamW optimizer in combination with a OneCycle learning rate schedule with cosine annealing and a maximum learning rate of $0.001$, with early stopping enabled and a patience of 50 epochs.
Training the neural \gls{gE} model through the surrogate-solver baseline diverged multiple times at higher learning rates.
The auxiliary loss weights are chosen to be $\lambda_G = 0.01$, and $\lambda_H = 0.05$.
We screened batch sizes from 16 to 512, and for a single fold and evaluated each model with the best configuration for the cross-validation runs. Details for all batch sizes and other model configurations are given in the SI.
The softmax temperature $\tau$ is initialized with $0.1$ and multiplied by a factor of $0.98$ each epoch.
However, performance with a fixed $\tau$ of $0.01$ performed similarly as shown for single fold experiments in the SI.
\subsection{Performance metrics}
\noindent In both test cases, performance is evaluated using system-level mean absolute error (MAE), root mean squared error (RMSE), and coefficient of determination $R^2$.
Let $x'^{\ast}_i, x''^{\ast}_i$ denote the ground-truth values and $x'_i, x''_i$ the corresponding predictions of the actual tie line, for sample $i$, with $N$ samples per output.
This tie line is obtained by evaluating the learned \gls{gE} model to compute the predicted \gls{gmix} curve, and computing the lowest \gls{gmix} pair of phase compositions of the discretized system with a grid of 101 points, and hence a discretization error of $0.01$.
The MAE is computed as the sum of the individual output-wise errors,
\begin{equation*}
\text{MAE}
=
\frac{1}{N} \sum_{i=1}^{N} \left(  \left| x'_i - x'^{\ast}_i \right|
+ \left| x''_i - x''^{\ast}_i \right| \right).
\end{equation*}
Similarly, the system-level RMSE is given by
\begin{equation*}
\text{RMSE}
=
\sqrt{
\frac{1}{N}
\sum_{i=1}^{N}
\left[
\left( x'_i - x'^{\ast}_i \right)^2
+
\left( x''_i - x''^{\ast}_i \right)^2
\right]}.
\end{equation*}
The coefficient of determination $R^2$ is calculated as usual, by concatenating the predictions of both phase compositions.
\section{Results}
\label{sec:Results}
In this section, we present the results of our evaluation on binary \gls{lle} phase-composition prediction.
We begin in the Fitting single systems subsection by benchmarking the surrogate solver against our differentiable DISCOMAX solver on the task of fitting a single system, highlighting the advantages of a thermodynamically consistent forward pass.

The Generalizing to new systems subsection then reports results of the generalization to unseen systems case and compares the performance of both solver approaches for a full GNN-based \gls{gE} model trained solely on \gls{lle} composition data, demonstrating how our proposed approach enables end-to-end learning of equilibrium behavior.
\subsection{Fitting single systems}\label{sec:single_sys}
\begin{figure}[htbp]
    \centering
    \includegraphics[width=1.\textwidth]{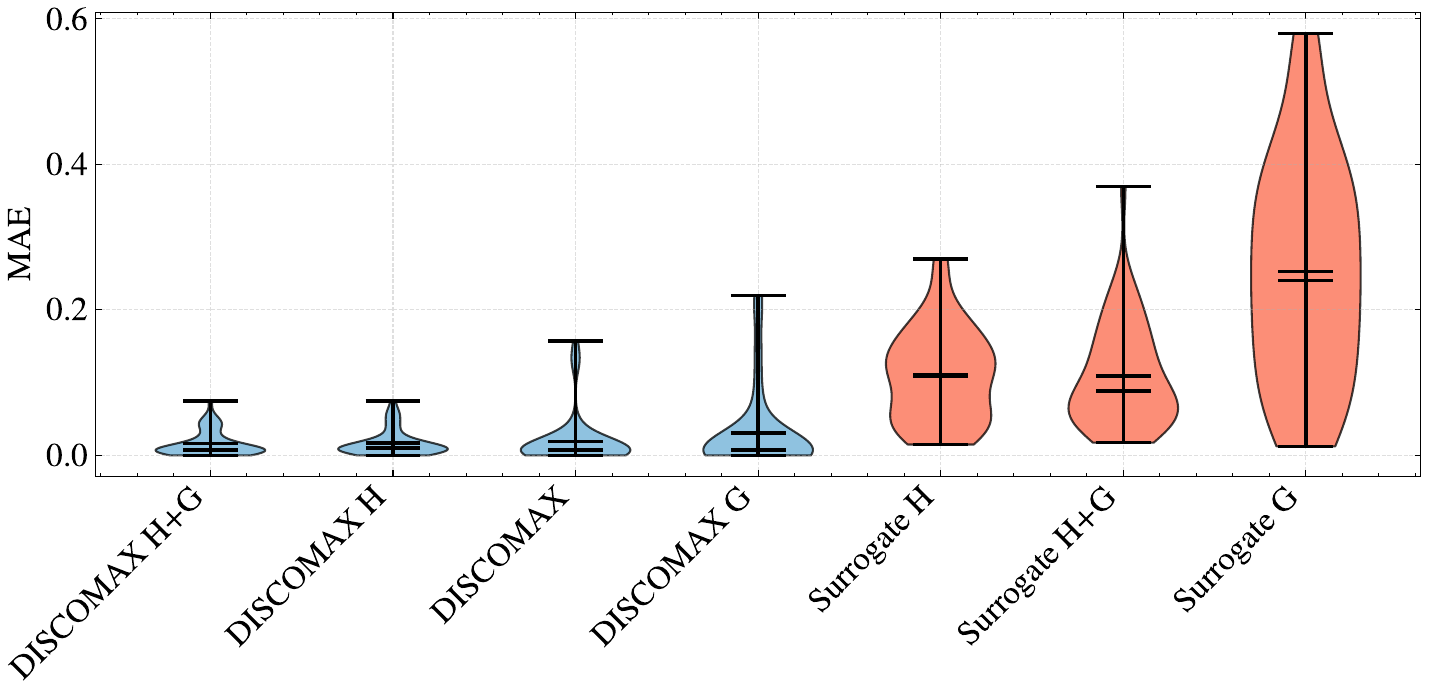}
    \caption{Violin plot of the MAE obtained by fitting 50 systems with different gap widths one by one (batch size = 1).
Each violin represents a different loss configuration; black markers indicate the mean, median, and extrema. H stands for the auxiliary Hessian loss, and G stands for the auxiliary Gibbs loss.}
    \label{fig:single-sys-violin}
\end{figure}
\begin{figure}[htbp]
    \centering
    \includegraphics[width=1.\linewidth]{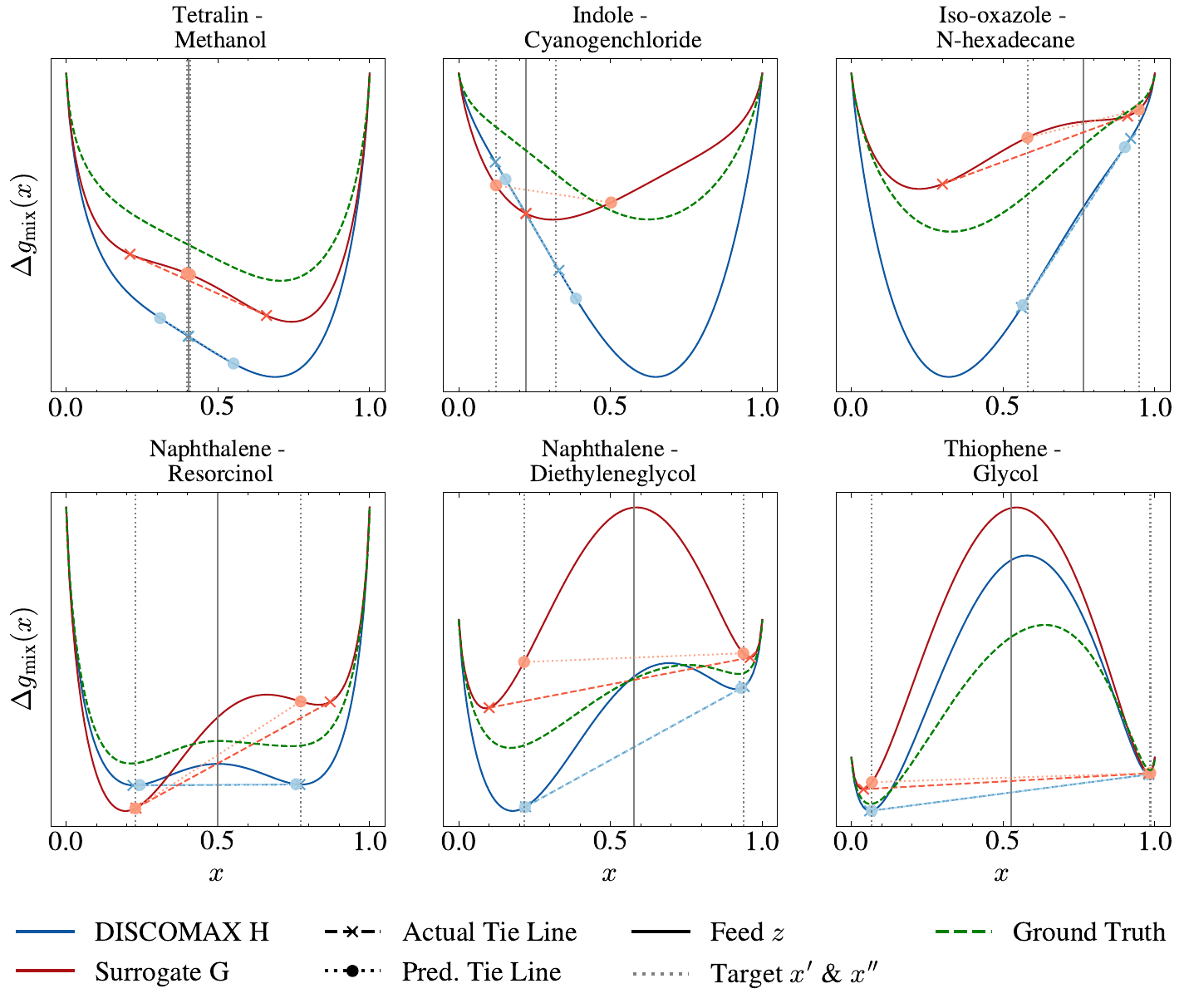}
    \caption{Six systems with increasing miscibility gap and comparison of the DISCOMAX H model (blue) and the Surrogate G model (red). The ground-truth \gls{gmix} profile is shown in green. Solid vertical lines indicate the feed composition, dotted vertical lines indicate the target phase compositions, dashed lines indicate actual tie lines, and dotted tie lines indicate predicted tie lines. Vertical scales are omitted because each panel is shifted independently for visualization. Absolute \gls{gmix} differences are not relevant for phase equilibrium.}
    \label{fig:single-sys-systems-gap-width}
\end{figure}
\begin{table}[b!]
    \centering
        \caption{Metrics of fitting 50 systems with different gap widths one by one (batch size = 1). H stands for the auxiliary Hessian loss, and G stands for the auxiliary Gibbs loss.}
   \begin{tabular}{lccc}
    \hline
    Model & MAE (mean $\pm$ std) & RMSE & $R^2$ \\
    \hline
    DISCOMAX H+G & 0.016 $\pm$ 0.017 & 0.018 & 0.998 \\
    DISCOMAX H & 0.017 $\pm$ 0.018 & 0.019 & 0.998 \\
    DISCOMAX & 0.019 $\pm$ 0.036 & 0.030 & 0.996 \\
    DISCOMAX G & 0.031 $\pm$ 0.057 & 0.046 & 0.990 \\
    \hline
    Surrogate H & 0.109 $\pm$ 0.064 & 0.102 & 0.952 \\
    Surrogate H+G & 0.109 $\pm$ 0.071 & 0.108 & 0.946 \\
    Surrogate G & 0.253 $\pm$ 0.153 & 0.237 & 0.740 \\
    \hline
    \end{tabular}
    \label{tab:single-sys-metrics}
\end{table}
 Table~\ref{tab:single-sys-metrics} and Figure~\ref{fig:single-sys-violin} present the mean absolute error (MAE) obtained when fitting 50 individual \gls{lle} systems spanning a broad range of miscibility-gap widths.
The MAE reflects the summed absolute deviation between the predicted and true compositions of both coexisting phases.
Each violin in Figure~\ref{fig:single-sys-violin} represents a kernel density estimate over all system MAEs for a given solver configuration, including both our thermodynamically consistent DISCOMAX solver (shown in blue) and the surrogate neural-network solver (shown in red), each trained either with the Hessian loss, the Gibbs loss, or both, as described in the Loss Function Definitions subsection.

Across all configurations, the proposed differentiable DISCOMAX solver consistently outperforms the surrogate baseline by a large margin. 
The best-performing configurations are DISCOMAX~H+G with an MAE of $0.016 \pm 0.017$ and DISCOMAX~H with $0.017 \pm 0.018$, compared to $0.109 \pm 0.064$ for Surrogate~H and $0.109 \pm 0.071$ for Surrogate~H+G, the strongest surrogate variants. The best surrogate variant has an approximately $6.8\times$ higher MAE relative to DISCOMAX~H+G, the best DISCOMAX variant.
Thus, even when both auxiliary loss terms are applied, the surrogate solver remains substantially less accurate.
The gap is even more pronounced without Hessian loss; the surrogate model trained with the Gibbs loss alone (Surrogate~G, as proposed by~\cite{hoffmannJIRASEKMachinelearnedExpressionExcess2025}) yields $\mathrm{MAE}$ of $0.253 \pm 0.153$ (approximately $16\times$ higher than DISCOMAX~H+G).
By contrast, all DISCOMAX variants deliver good predictions, and incorporating the Hessian loss yields the greatest improvements, as reflected by consistently lower MAE/RMSE and near-unity $R^2$ values, as also shown in Table~\ref{tab:single-sys-metrics}.
Even the pure DISCOMAX method without any auxiliary loss performs about $5.7\times$ better than the best surrogate methods and about $13\times$ better than the previously proposed Surrogate~G method, in terms of MAE.
Especially considering the discretization width of $0.01$, these results can be considered satisfactory for the DISCOMAX solver.

Figure~\ref{fig:single-sys-systems-gap-width} provides a qualitative comparison of the fitted phase-composition profiles for six representative systems with different target miscibility gap widths for each model, respectively.
The smallest-gap examples can appear close to single-phase behavior at the plotted scale, but they still contain a finite phase split.
It is also important to note that the difference in absolute \gls{gmix} values between the predicted and true curves is not relevant since it can not be uniquely determined from the target phase compositions alone, and only the relative shape of the \gls{gmix} curve determines the predicted equilibrium state.
The surrogate solver consistently displays substantially larger deviations from the specified target equilibrium compositions and frequently produces thermodynamically inconsistent equilibrium composition predictions, with violations of the extremum principle, i.e., the predicted configuration is unstable, and not even metastable.
More examples of this failure are shown in the SI.

Notably, for several systems, the surrogate predictions appear to have converged, yielding surrogate predictions exactly on target, yet to entirely incorrect solutions, indicating that the learned model provides stable gradients but not ones that point toward the equilibrium state.
Similar limitations were also highlighted in~\cite{mitsos2009bilevel} for local methods optimizing iso-activity for fitting NRTL parameters.
By comparison, the DISCOMAX solver with Hessian loss produces qualitatively accurate (close to the target) and, as guaranteed, thermodynamically consistent predictions, also for the worst-performing instances.
Furthermore, we note a systematic difficulty in capturing very narrow miscibility gaps with high precision, which is a common problem also in traditional parametric \gls{gE} models~\cite{de1988thermodynamics}.
% new comment here
Furthermore, we observe that the softmax predictions are generally close to the hard argmin predictions but with a better agreement for systems with sharper minima and worse agreement for systems with flatter minima, i.e., a smaller concave region in the \gls{gmix} curve.
This behavior follows directly from the softmax weighting and our use of a fixed temperature schedule for $\tau$. For a state with Gibbs energy $\Delta g^{\textrm{mix},(i,j)}$, its weight is proportional to
$p^{(i,j)} \sim \exp(-\Delta g^{\textrm{mix},(i,j)}/\tau)$.
If the discrete global minimum is separated from the remaining states by a large (energy) gap, the non-minimal states are exponentially suppressed, and the soft prediction concentrates strongly around the hard argmin, forcing agreement. Conversely, if the global minimum is only weakly separated from the remaining states, the soft prediction is a weighted average over many states and can deviate significantly from the hard argmin.

Taken together, these experiments demonstrate that the proposed DISCOMAX equilibrium solver provides a substantially more reliable and physically grounded approach for fitting individual \gls{lle} systems than the surrogate neural-network solver.
Figure~\ref{fig:single-sys-violin} shows consistently lower errors across a wide variety of systems, while the qualitative comparisons in Figure~\ref{fig:single-sys-systems-gap-width} highlight that surrogate predictions often violate basic thermodynamic principles and may converge to entirely incorrect equilibria.
In contrast, the DISCOMAX formulation preserves thermodynamic consistency in all cases, with deviations arising primarily in systems with extremely narrow miscibility gaps.
In addition to that, the DISCOMAX variant without auxiliary loss performs already well on its own and even better when paired with the Hessian loss.
These results establish the DISCOMAX solver as a robust foundation for end-to-end learning of thermodynamically consistent \gls{gE} models, motivating its use in the full multi-system training described in the following section.

\subsection{Generalizing to new systems}\label{sec:full_training}
We evaluate the generalization performance of the proposed thermodynamically consistent DISCOMAX equilibrium GNN using 10-fold cross-validation on the full dataset of binary \gls{lle} systems.
The method is compared against the surrogate-based approach under identical data splits.
The batch size was chosen to be 64 for the DISCOMAX~H and Surrogate~G variant and 256 for the DISCOMAX variant, based on a grid search on a single fold. Single-fold results for all batch sizes are given in the SI.  

\begin{table}[bpht]
\centering
\caption{Full cross-validation results for the phase-composition predictions in the generalization study, aggregated over ten folds. Metrics are shown as mean $\pm$ standard deviation.}
\begin{tabular}{l l c c c}
\toprule
\textbf{Model} & \textbf{Split} & MAE & RMSE & \textbf{$R^{2}$} \\
\midrule

\multirow{3}{*}{DISCOMAX}
& Test & 0.075 $\pm$ 0.003 & 0.098 $\pm$ 0.005 & 0.965 $\pm$ 0.003 \\
& Validation & 0.075 $\pm$ 0.003 & 0.098 $\pm$ 0.004 & 0.965 $\pm$ 0.003 \\
& Train & 0.035 $\pm<$0.001 & 0.040 $\pm<$0.001 & 0.993 $\pm<$0.001 \\

\midrule

\multirow{3}{*}{DISCOMAX~G}
& Test & 0.070 $\pm$ 0.002 & 0.091 $\pm$ 0.005 & 0.969 $\pm$ 0.003 \\
& Validation & 0.070 $\pm$ 0.003 & 0.091 $\pm$ 0.005 & 0.969 $\pm$ 0.003 \\
& Train & 0.024 $\pm<$0.001 & 0.032 $\pm$ 0.001 & 0.995 $\pm<$0.001 \\

\midrule

\multirow{3}{*}{DISCOMAX~H}
& Test & 0.068 $\pm$ 0.003 & 0.090 $\pm$ 0.005 & 0.970 $\pm$ 0.003 \\
& Validation & 0.068 $\pm$ 0.003 & 0.090 $\pm$ 0.005 & 0.970 $\pm$ 0.004 \\
& Train & 0.022 $\pm<$0.001 & 0.023 $\pm<$0.001 & 0.998 $\pm<$0.001 \\

\midrule

\multirow{3}{*}{DISCOMAX~H+G}
& Test & \textbf{0.067 $\pm$ 0.002} & \textbf{0.089 $\pm$ 0.005} & \textbf{0.971 $\pm$ 0.003} \\
& Validation & 0.069 $\pm$ 0.003 & 0.092 $\pm$ 0.006 & 0.969 $\pm$ 0.004 \\
& Train & 0.019 $\pm<$0.001 & 0.020 $\pm<$0.001 & 0.998 $\pm<$0.001 \\

\midrule

\multirow{3}{*}{Surrogate~G}
& Test & 0.080 $\pm$ 0.002 & 0.098 $\pm$ 0.004 & 0.965 $\pm$ 0.003 \\
& Validation & 0.080 $\pm$ 0.003 & 0.097 $\pm$ 0.005 & 0.965 $\pm$ 0.004 \\
& Train & 0.041 $\pm<$0.001 & 0.047 $\pm<$0.001 & 0.991 $\pm<$0.001 \\

\midrule

\multirow{3}{*}{Surrogate~H}
& Test & 0.072 $\pm$ 0.002 & 0.091 $\pm$ 0.005 & 0.970 $\pm$ 0.003 \\
& Validation & 0.072 $\pm$ 0.003 & 0.091 $\pm$ 0.005 & 0.969 $\pm$ 0.004 \\
& Train & 0.027 $\pm<$0.001 & 0.027 $\pm<$0.001 & 0.997 $\pm<$0.001 \\

\midrule

\multirow{3}{*}{Surrogate~H+G}
& Test & 0.072 $\pm$ 0.003 & 0.093 $\pm$ 0.006 & 0.968 $\pm$ 0.004 \\
& Validation & 0.073 $\pm$ 0.003 & 0.092 $\pm$ 0.006 & 0.969 $\pm$ 0.004 \\
& Train & 0.025 $\pm<$0.001 & 0.025 $\pm<$0.001 & 0.997 $\pm<$0.001 \\
\bottomrule
\end{tabular}
\label{tab:cv}
\end{table}
 
\begin{figure}
    \centering
    \begin{overpic}[width=0.48\linewidth]{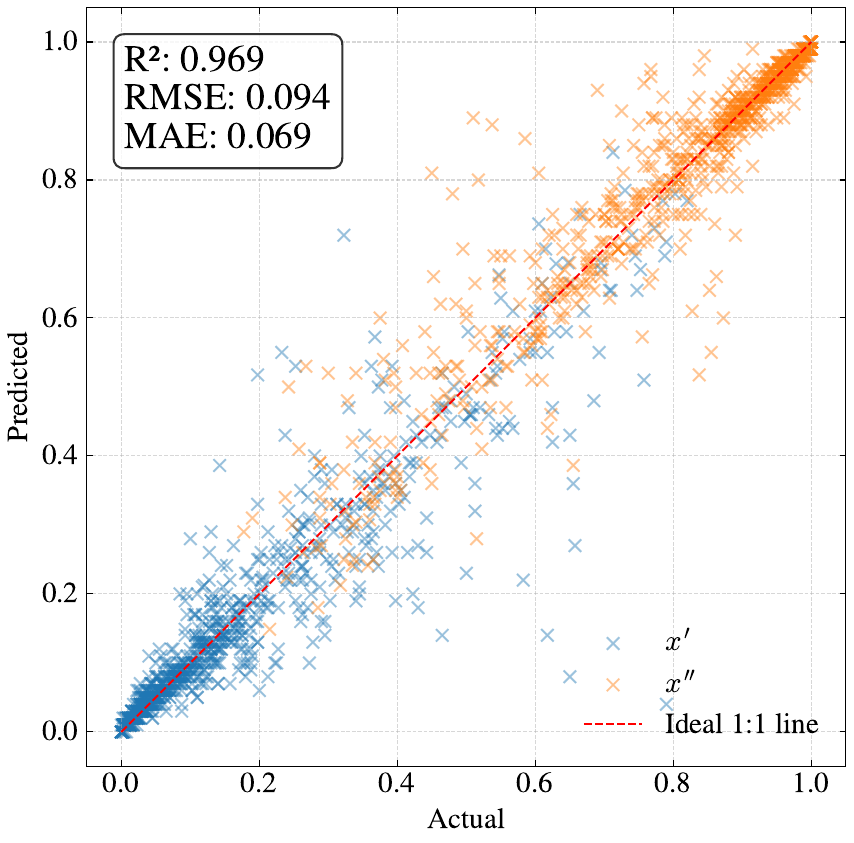}
        \put(13,100){a. DISCOMAX~H (Ours)}
    \end{overpic}
    \begin{overpic}[width=0.48\linewidth]{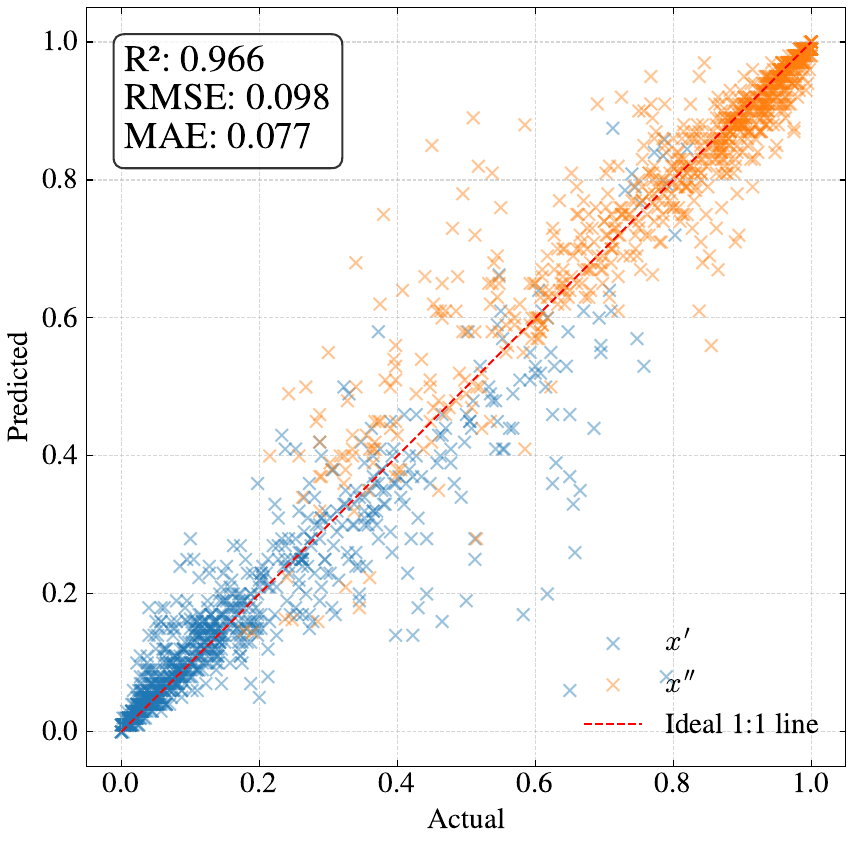}
        \put(13,100){b. Surrogate~G}
    \end{overpic}

    \vspace{1em}

    \begin{overpic}[width=0.48\linewidth]{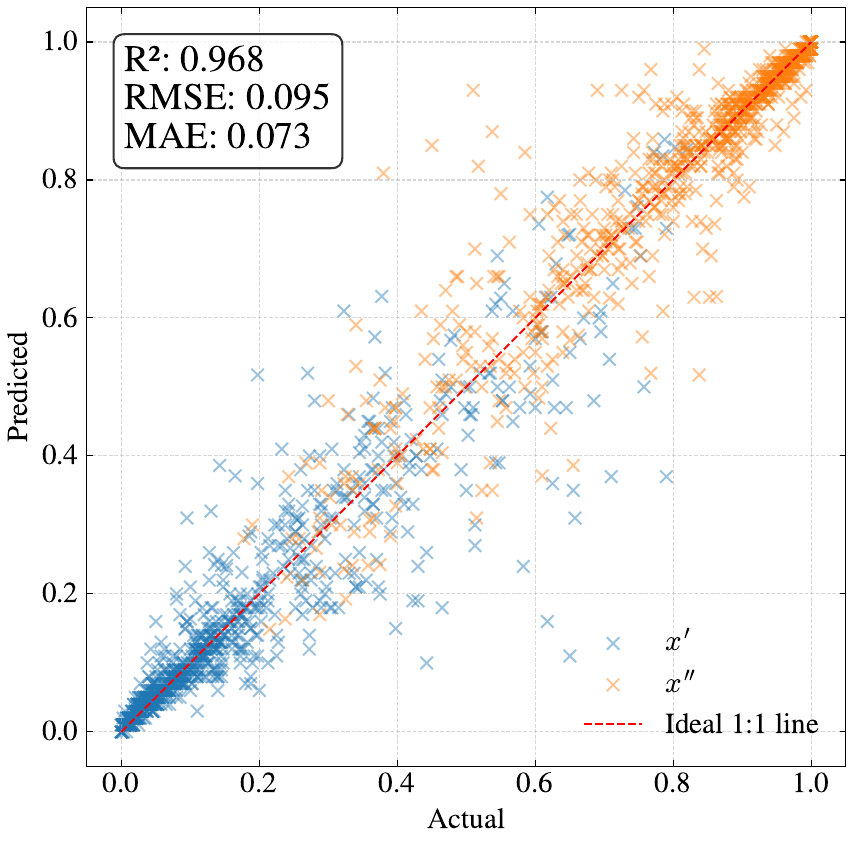}
        \put(13,100){c. DISCOMAX (Ours)}
    \end{overpic}

    \caption{Test-set parity plots of phase compositions for representative model variants: a. DISCOMAX~H, b. Surrogate~G, and c. DISCOMAX. Full cross-validation results for all other variants are reported in Table~\ref{tab:cv}.}
    \label{fig:parity_2x2_test}
\end{figure}

\begin{figure}
    \centering
    \begin{overpic}[width=0.48\linewidth]{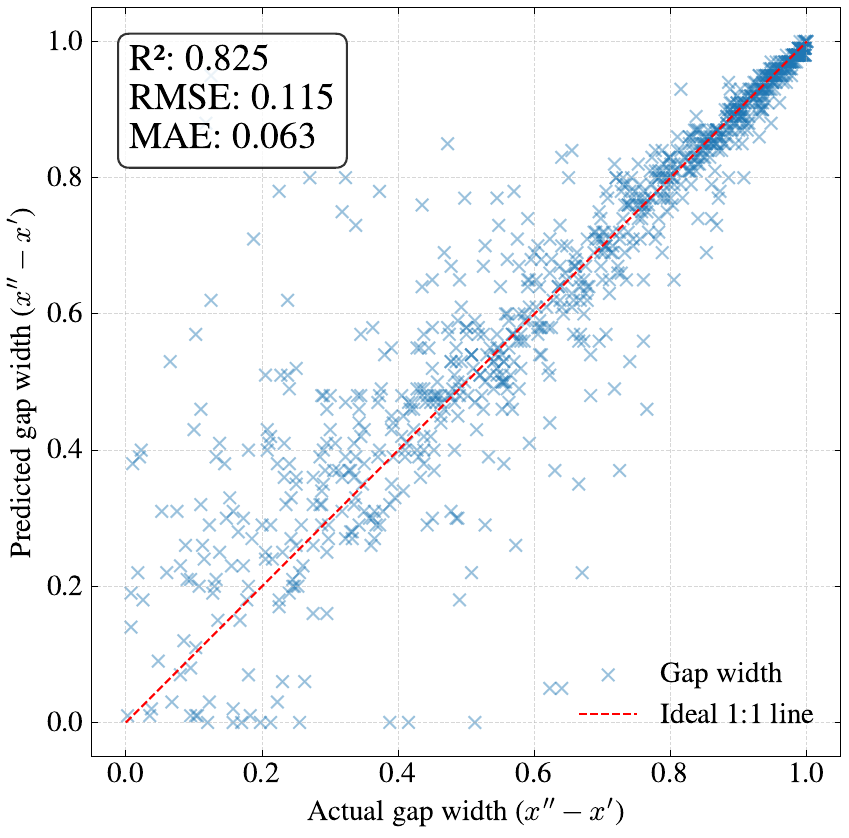}
        \put(13,100){a. DISCOMAX~H (Ours)}
    \end{overpic}
    \begin{overpic}[width=0.48\linewidth]{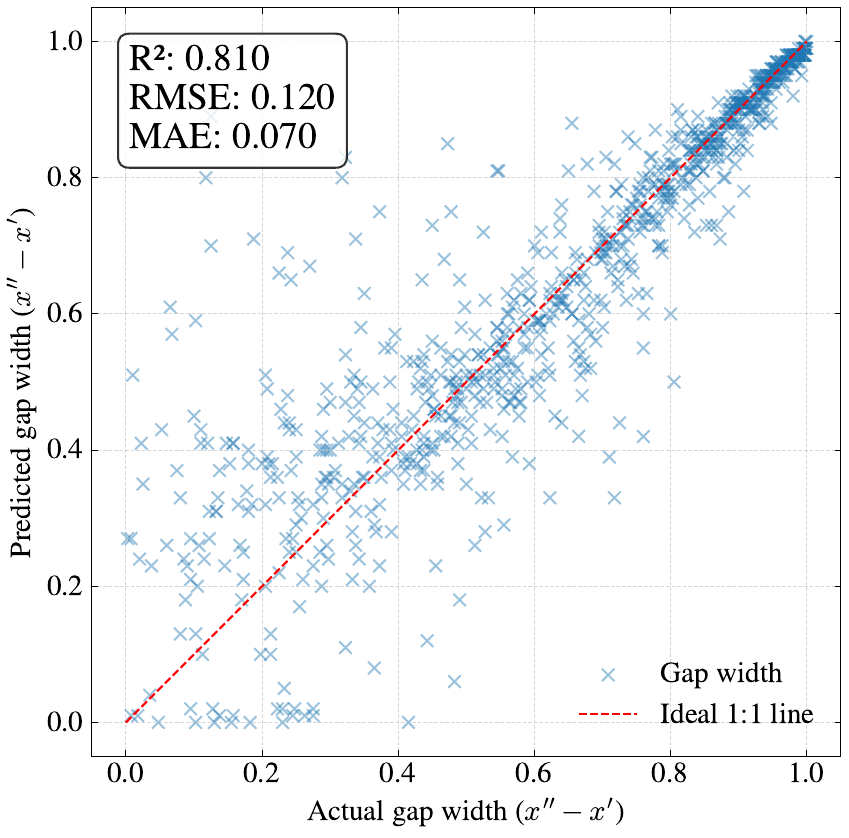}
        \put(13,100){b. Surrogate~G}
    \end{overpic}
    \caption{Test-set gap-width parity plots for DISCOMAX~H and Surrogate~G on a representative fold.}
    \label{fig:parity_gap_width_test}
\end{figure}

\begin{figure}
    \centering
    \begin{overpic}[width=0.48\linewidth]{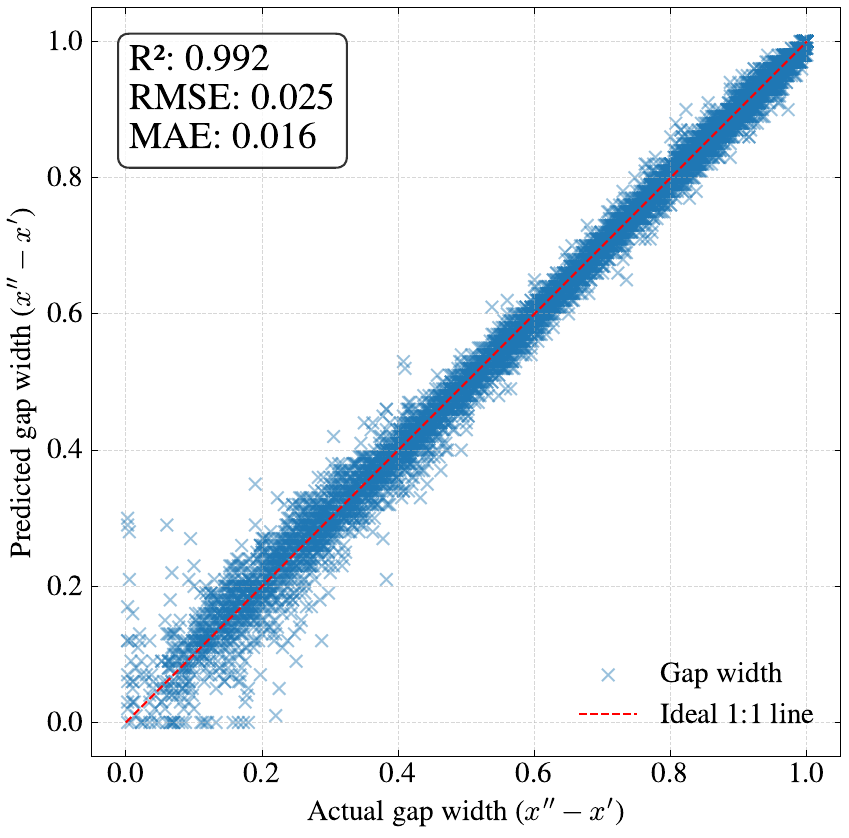}
        \put(13,100){a. DISCOMAX~H (Ours)}
    \end{overpic}
    \begin{overpic}[width=0.48\linewidth]{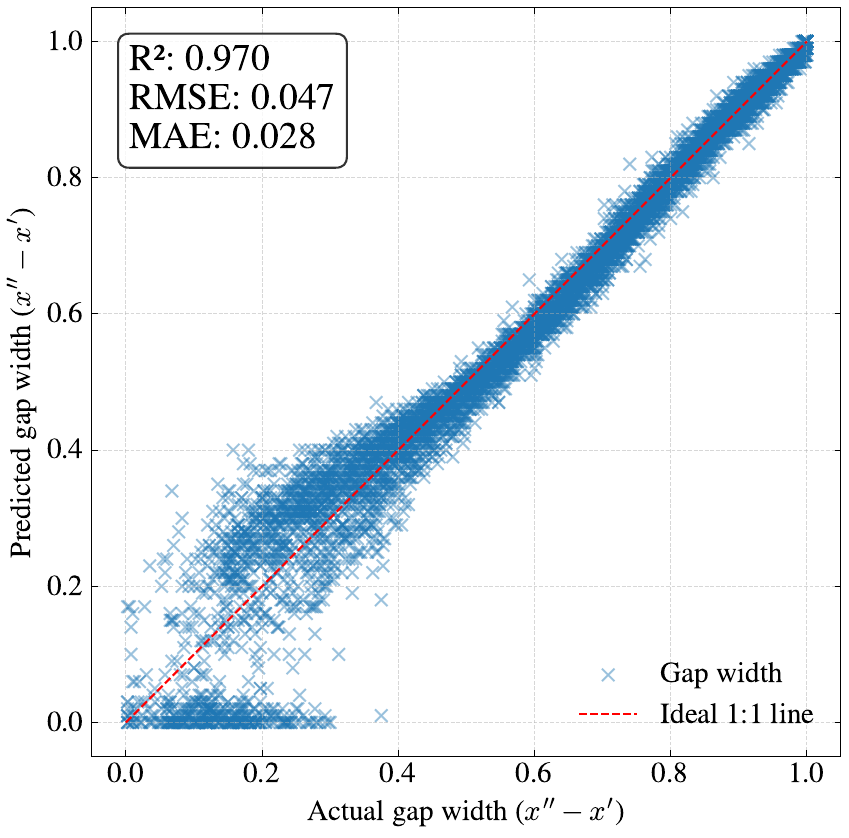}
        \put(13,100){b. Surrogate~G}
    \end{overpic}
    \caption{Training-set gap-width parity plots for DISCOMAX~H and Surrogate~G on the same fold as Figure~\ref{fig:parity_gap_width_test}.}
    \label{fig:parity_gap_width_train}
\end{figure}

\begin{figure}[htbp]
    \centering
    \includegraphics[width=.8\textwidth]{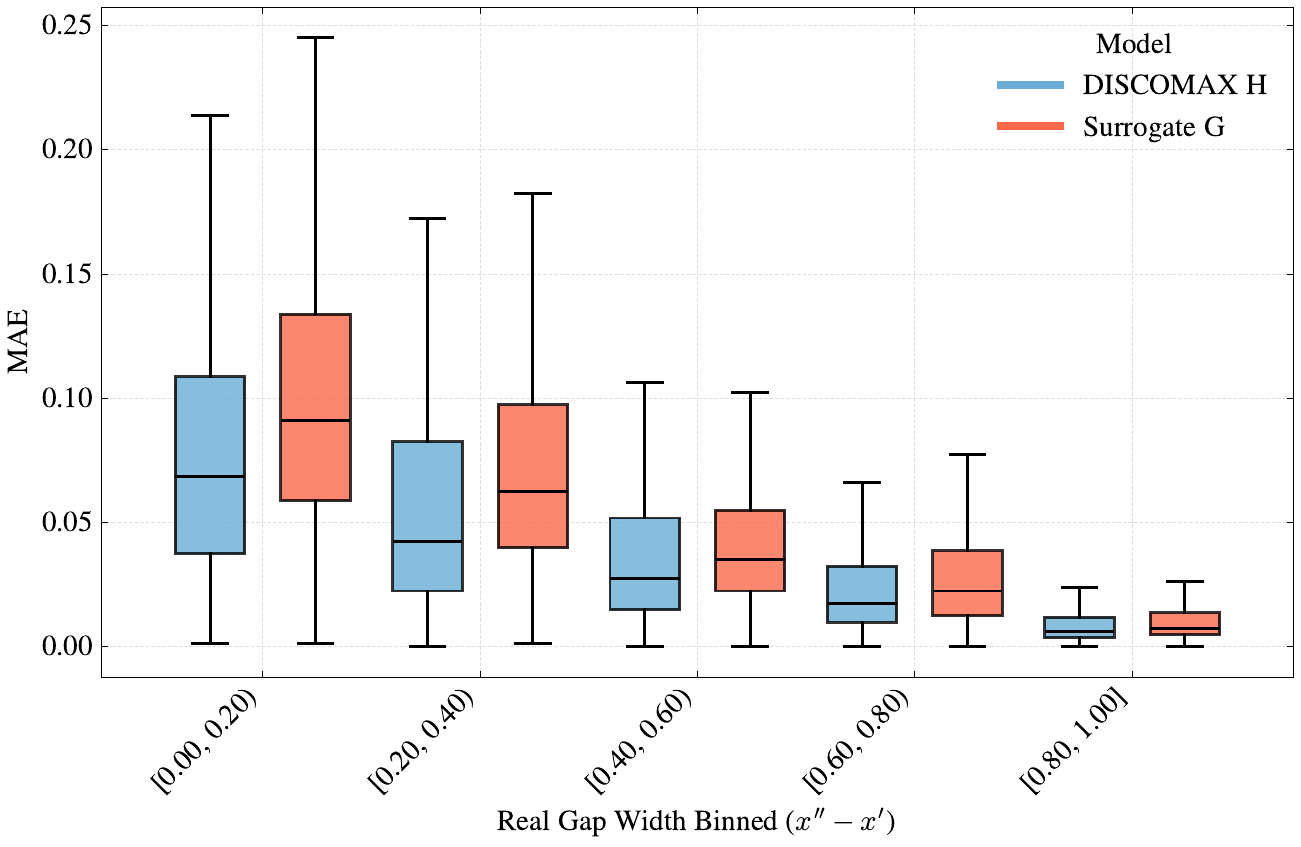}
    \caption{Cross-validation test-set MAE of phase-composition predictions binned by miscibility-gap width for the DISCOMAX~H and Surrogate~G models. Boxes show the interquartile range, black lines show medians, and whiskers extend to the default 1.5 interquartile range.}
    \label{fig:boxplot_dx}
\end{figure}

\begin{figure}[htbp]
    \centering
    \includegraphics[width=1.\textwidth]{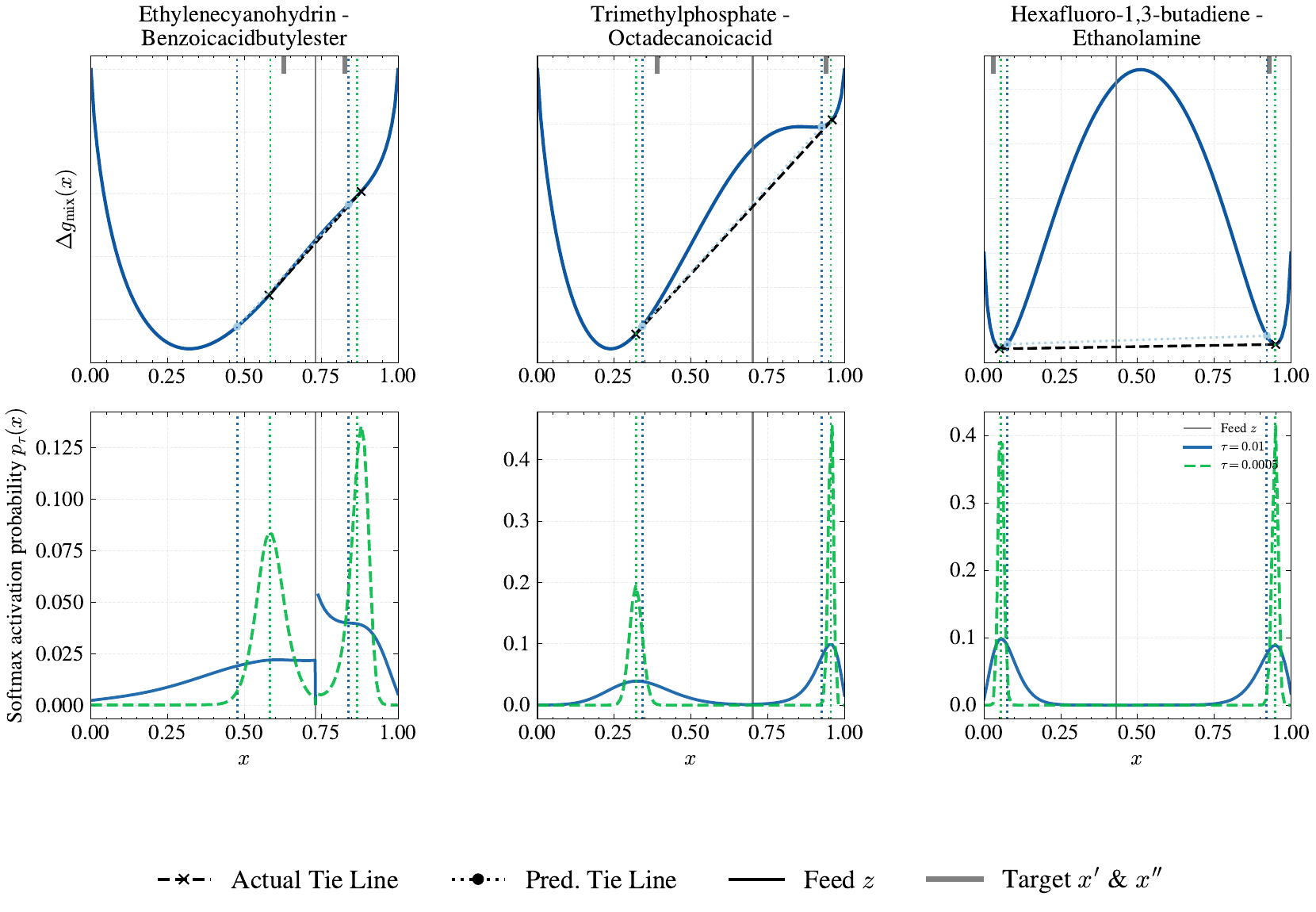}
    \caption{Representative DISCOMAX~H \gls{gmix} profiles and corresponding endpoint softmax activation probabilities for three held-out systems with increasing miscibility-gap width. The upper row shows the learned \gls{gmix} curves and predicted tie lines, while the lower row shows the marginal probabilities assigned to feasible $x'$ and $x''$ endpoints for two softmax temperatures.}
    \label{fig:gmix_profiles_probabilities}
\end{figure}

Figure~\ref{fig:parity_2x2_test} shows parity plots of predicted versus reference phase compositions on the test set.
Quantitative phase-composition results aggregated over all folds are reported in Table~\ref{tab:cv}.
Among all configurations, DISCOMAX~H+G achieves the best test performance, with a mean absolute error of $0.067 \pm 0.002$, a root mean squared error of $0.089 \pm 0.005$, and an $R^2$ of $0.971 \pm 0.003$, closely followed by DISCOMAX~H at $0.068 \pm 0.003$ test MAE.
Within the surrogate family, adding the Hessian loss reduces the test MAE from $0.080 \pm 0.002$ for Surrogate~G to $0.072 \pm 0.002$ for Surrogate~H and $0.072 \pm 0.003$ for Surrogate~H+G, but these variants remain less accurate than the best DISCOMAX models.
We continue to show Surrogate~G in the main parity plots because it corresponds most closely to the published HANNA 2 surrogate training setup. The cross-validation MAE distributions across all trained variants are further visualized in the SI.

% The differences between the approaches become more pronounced on the training set
% shown in Figure~\ref{fig:parity_2x2_train}.
% Both DISCOMAX-based models achieve substantially lower training errors than the surrogate solver, with MAEs of $0.022 \pm < 0.001$ (DISCOMAX~H) and $0.035 \pm 0.001$ (DISCOMAX), compared to $0.041 \pm 0.001$ for the Surrogate~G approach.
% This corresponds to an error reduction of roughly $46\%$ for DISCOMAX~H and about $15\%$ for DISCOMAX, indicating that the proposed differentiable equilibrium solver is able to fit the training data much more accurately.
% The surrogate solver, in contrast, is limited by the approximation error inherent in learning a direct mapping from discretized \gls{gmix} curves to phase compositions.
% The Surrogate~G parity plots also show a systematic tendency to narrow the predicted miscibility gap, i.e., to overestimate $x'$ and underestimate $x''$.
% The same tendency is present in the test set, but it is less visually prominent because the test split contains fewer points.
Figure~\ref{fig:parity_gap_width_test} provides a more detailed held-out view of the predicted versus target miscibility gap widths for DISCOMAX~H and Surrogate~G on a representative test fold.
For the test-set gap-width predictions, the Surrogate~G model has a higher difficulty in predicting narrow miscibility gaps accurately than the DISCOMAX~H model.
These systems are likely more difficult to fit since they are more likely to produce discrete jumps in the loss function from overshooting and oscillation between single-phase and two-phase \gls{gmix} curves.
Figure~\ref{fig:parity_gap_width_train} reports the corresponding training-split behavior.
On the training data, Surrogate~G shows a stronger tendency to fully compress narrower miscibility gaps than DISCOMAX~H.
This qualitative trend is not observed on the test set, where their performance seems similar, but the test data is also much sparser in the narrow-gap regime. 

To further investigate the differences in performance, Figure~\ref{fig:boxplot_dx} provides a binned analysis of the MAE as a function of the miscibility gap width, defined as $(x'' - x')$ for the DISCOMAX~H model and the Surrogate~G baseline.
As expected, the modeling of phase behavior becomes increasingly difficult as the system approaches the region of narrower gap widths.
Here, the DISCOMAX~H model demonstrates a clear advantage in these challenging regimes.
For gap widths $\le 0.40$, the proposed model significantly reduces both the median error and error variance compared to the Surrogate~G baseline.
As the gap width increases, the errors for both models converge towards the discretization precision of $0.01$.
Figure~\ref{fig:gmix_profiles_probabilities} complements this aggregate view by showing representative DISCOMAX~H \gls{gmix} profiles for three systems with narrow, intermediate, and wide miscibility gaps.
The top row shows the learned \gls{gmix} curves, the corresponding actual tie lines (DISOMAX forward pass), and the predicted soft tie lines (DISCOMAX backward pass), shown at a large $\tau$ of $0.01$ in order to clearly illustrate the deviation.
The test labels, i.e., the target phase compositions, are indicated with thick grey ticks.
The corresponding softmax activation plots show how probability mass is assigned only to feasible $x$.
For $tau = 0.01$, shown in blue, the softmax distribution is dispersed over a wider range of feasible states, especially for the first system, which is only weakly non-convex.
Lowering to $\tau = 0.0005$, shown in green, sharpens the distributions, moving the differentiable prediction $x_{\text{soft}}', x_{\text{soft}}''$ (dotted lines), used in the DISCOMAX backward pass, toward the discrete hard argmin tie line, shown with black crosses.
Additional auxiliary loss-combinations, batch-size, and \gls{gmix}-profile plots are provided in the SI.

Overall, the findings highlight both the strengths and shortcomings of surrogate-based equilibrium prediction, and they demonstrate that the proposed differentiable DISCOMAX solver provides a more robust and physically grounded foundation for learning equilibrium behavior directly from data.
\section{Discussion}
\label{sec:Discussion}
The results of this study yield several noteworthy observations.
We were able to show that we can train a machine learning model based on binary \gls{lle} data in a thermodynamically consistent manner, i.e., the predicted phase composition is the globally minimal Gibbs energy state, up to a chosen discretization.
This has, to the best of our knowledge, not been shown before.

On top of that, we achieve slightly better predictive performance in comparison to surrogate-based approaches, and we do not even require any auxiliary loss for this.
Not requiring an auxiliary loss was especially surprising to us, since the training of surrogate solver model variants diverges completely without it, even at the largest test batch size of 512.

While the surrogate solver performs unexpectedly poorly when fitted to individual systems, these deficiencies do not fully propagate to the full-dataset training scenario.
A likely explanation is that fitting each system in isolation exposes the surrogate solver to its own inaccurate equilibrium estimates, causing the optimization to stabilize at incorrect solutions.
In contrast, stochastic gradient descent over mini-batches, used in full training, reduces this effect by repeatedly varying the training signal, thereby preventing systematic convergence to these incorrect local minima.
Furthermore, the approximation errors of the surrogate solver might also act as a regularizer, which could explain the discrepancy between train and test set performance of both the DISCOMAX and surrogate solvers. 

The qualitative analysis for the single-system experiments presented above reveals that training with the surrogate solver often converges smoothly but to entirely wrong solutions.
This indicates that although the surrogate model provides stable gradients that facilitate optimization, and thus enable differentiation through the optimization problem, these gradients are biased and do not reliably point towards the true target equilibrium state.
Correcting such behavior would require supplying additional curated training data for every failure mode, which becomes increasingly impractical and non-scalable in higher-dimensional mixture spaces.

A key advantage of our proposed differentiable binary equilibrium solver, next to being thermodynamically consistent in the forward pass by construction, is that it is genuinely end-to-end trainable.
Unlike surrogate-based methods, it is not biased by the synthetically generated data used for training the surrogate solver (e.g., UNIFAC).
This is because the differentiable equilibrium calculation itself contains no trainable parameters.
This makes the method general and widely applicable to \emph{all} equilibrium properties, such as pure-component vapor–liquid equilibria and three-phase LLLE (shown in S2 in the SI), but also to solid–liquid equilibria~\cite{al2025accurately} and purely solid equilibrium structures.
In such settings, surrogate approaches would require application-specific training data and solvers, whereas the proposed method naturally transfers without modification.
\section{Limitations \& Future Steps}
\label{sec:FutWork}
The present work focuses exclusively on learning the \gls{gE} function from the equilibrium phase composition of binary liquid-liquid systems. As such, the training data contain only positive samples, i.e., mixtures that exhibit two-phase coexistence.
An important next step would be to also incorporate data on miscible systems into training and force \gls{gmix} to be convex for such samples.
Integrating both behaviors within a single framework would enable the model not only to predict phase-split compositions but also to determine whether a phase split occurs in the first place.
At present, no machine-learning-based method jointly addresses both tasks in a thermodynamically principled manner (aside from traditional rigorous solvers).
Although we did not investigate this explicitly, it is plausible that the surrogate approach of \cite{hoffmannJIRASEKMachinelearnedExpressionExcess2025} exhibits similar limitations: the model receives no supervision signal for systems in which no phase split should occur.
This suggests an interesting direction for future research, namely the use of miscibility or single-phase behavior as a complementary, weakly supervised learning signal.
An unexpected potential advantage of the surrogate solver baseline may be that it implicitly provides a form of regularization through the noise introduced by the surrogate solver itself.
We speculate this because the gap between training and test error is a lot smaller for this method compared to ours, although further investigation would be required to confirm this effect.
In our experiments, we did not apply dedicated regularization techniques beyond a small amount of weight decay in AdamW.
This choice allowed for a cleaner comparison between our model and the baseline without introducing any additional hyperparameters.
It is possible that the test performance of our method could be improved by incorporating stronger regularization, for example, by adding dropout layers.
Another potential strategy to mitigate overfitting would be the use of pretrained embeddings.
Exploring these directions remains an interesting avenue for future work.

While our case study has focused on binary \gls{lle} systems, the proposed framework is, in principle, applicable to pure-component VLE, multicomponent \gls{lle}, or SLE, for example.
Using it as a general-purpose equilibrium solver for mixtures with more than a few components would be challenging, since the computational cost of the current, largely brute-force, discretization-based algorithm scales poorly with the number of components. In particular, the number of groups that can be formed from a given set of states grows combinatorially, with an order of magnitude determined by the binomial coefficient.
It does not pose a practical obstacle in our intended use case of training neural \Gls{gE} models, since experimentally available phase-equilibrium datasets consist mostly of binary systems.
In this setting, the approach remains extremely efficient and fully applicable.
During deployment, practitioners can still simply rely on their method of choice for performing flash calculations. 
To improve the handling of multiple components, a promising direction would be to combine a coarse discretization with DISCOMAX, using the resulting solution as both an effective initial guess and a fallback gradient estimate. The solution could then be refined using a local optimization method, provided that the problem is sufficiently well-conditioned. Integrating this approach with implicit differentiation through the locally optimized solution appears particularly promising.

\section{Conclusion}
\label{sec:Conclusion}
We address the challenge of integrating rigorous thermodynamic constraints into machine-learning models whose target quantities are defined implicitly as solutions of an extremum principle.
Focusing on binary \gls{lle}, we introduce a differentiable equilibrium solver that reformulates the flash calculation as a discrete minimization problem over the \gls{gmix} surface.
Thermodynamic consistency is enforced exactly in the forward pass by evaluating the hard minimizer, while differentiability is maintained in the backward pass via a straight-through gradient estimator that propagates gradients via a Boltzmann-weighted aggregation over discretized compositions.

Building on this solver, we train GNN models for the excess Gibbs energy using only equilibrium phase-composition data, without requiring direct supervision on \gls{gE} or activity coefficients.
Both the single-system fitting and the generalization experiments demonstrate that the proposed approach reliably predicts phase splits across a wide range of miscibility-gap widths, with systematic deviations primarily confined to systems exhibiting a narrow two-phase region.

We benchmark our framework against a neural network surrogate solver trained on UNIFAC-generated data.
Although the surrogate solver provides smooth gradients, it frequently converges to thermodynamically inconsistent and incorrect equilibrium states when trying to fit individual systems.
In contrast, the proposed DISCOMAX solver preserves thermodynamic consistency at every step of training and inference and achieves superior quantitative performance in terms of mean absolute error on phase compositions and gap widths in full end-to-end GNN training.

Our results demonstrate that rigorous equilibrium thermodynamics can be embedded into modern machine-learning pipelines when the target properties arise from fairly small lower-level optimization problems.
The proposed framework establishes a principled foundation for learning thermodynamically grounded models directly from phase-equilibrium data and can, in principle, be extended to other types of phase behavior, including vapor-liquid and solid-liquid equilibria, as well as to more complex chemical systems.

\section*{Data and Software Availability}
The code of this work is available under the open-source Eclipse Public License at: \url{https://git.rwth-aachen.de/avt-svt/public/differentiable-thermo-eq-ml}.
The full dataset consisting of HANNA 2 predictions for all 138,601 possible binary systems and 8,597 \gls{lle} mixtures is available in the same repository.
\section*{Author contributions}
Karim K. Ben Hicham: Conceptualization, Methodology, Formal analysis, Investigation, Writing - original draft; development and implementation of the original algorithm.
Moreno Ascani: Conceptualization, Methodology, Writing - original draft; theoretical analysis and thermodynamic formulation of the algorithm.
Jan G. Rittig: Conceptualization, Writing - review \& editing.
Alexander Mitsos: Conceptualization, Supervision, Funding acquisition, Writing - review \& editing.
\section*{Supporting information}
The Supporting Information provides a statistical-thermodynamics-inspired derivation for the proposed equilibrium solver, together with illustrative case studies covering binary LLE, ternary LLLE, and vapor-pressure calculations.
In addition, it presents implementation details of the surrogate solver and supplementary results regarding the single-system fitting and the full-dataset training experiments.
\section*{Acknowledgments}
We acknowledge support of the
Werner Siemens Foundation in the frame of the WSS Research Center
``catalaix''.
\section*{Declaration of competing interest}
The authors declare that they have no known competing financial interests or personal relationships that could have appeared to influence the work reported in this paper.
\ifdefined\arxivcombined
\else
\printbibliography
\newpage

\section*{Graphical abstract}
\rule{0.05in}{1.75in}%
\begin{minipage}[b][1.75in]{3.25in}
  \sffamily
  \frenchspacing
  \includegraphics[width=\linewidth,height=1.75in,keepaspectratio]{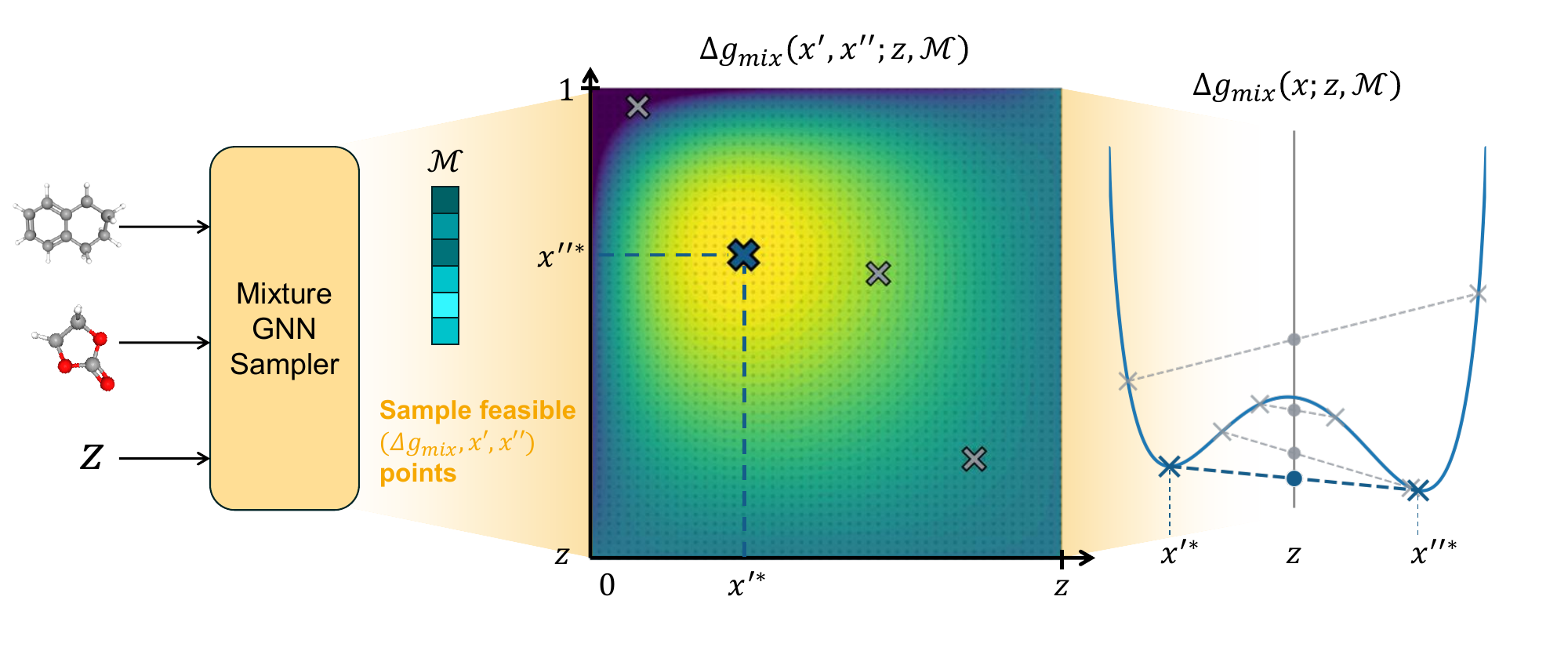}
\end{minipage}%
\rule{0.05in}{1.75in}

\fi

%% file: si/supplement.tex
\setcounter{page}{1}

\ifdefined\arxivcombined
\section*{Supporting Information}
\else
\maketitle
\tableofcontents
\fi
\section{Thermodynamically inspired derivation}
In this Section, we propose two complementary procedures for the same \gls{lle} $T,p$-Flash calculation, to which we refer as mass balance explicit (formulation 1) and mass balance implicit (formulation 2).
Both formulations, which are equivalent methods for performing the \gls{lle} $T,p$-Flash calculation, are derived in detail in this section by assuming random concentration fluctuations within a system at given temperature $T$, pressure $p$, and concentration $\bar{z}$, and appealing to statistical thermodynamics arguments.
First, a model system with random concentration fluctuations is defined and visualized, and relevant assumptions important for the subsequent mathematical derivation of the methods are discussed.
The resulting working equations, which in statistical thermodynamics terms can be interpreted as the probability distribution of the possible configurations of the system, yield the solution of a constrained convex optimization problem.
This optimization problem is first formulated based on the definition and visualization of the model system and the underlying assumptions.
From the stationary conditions of the Lagrangian of this optimization problem, the working equations are then derived and discussed.  \\

\begin{figure}[htbp]
    \centering
    \includegraphics[width=0.6\textwidth]{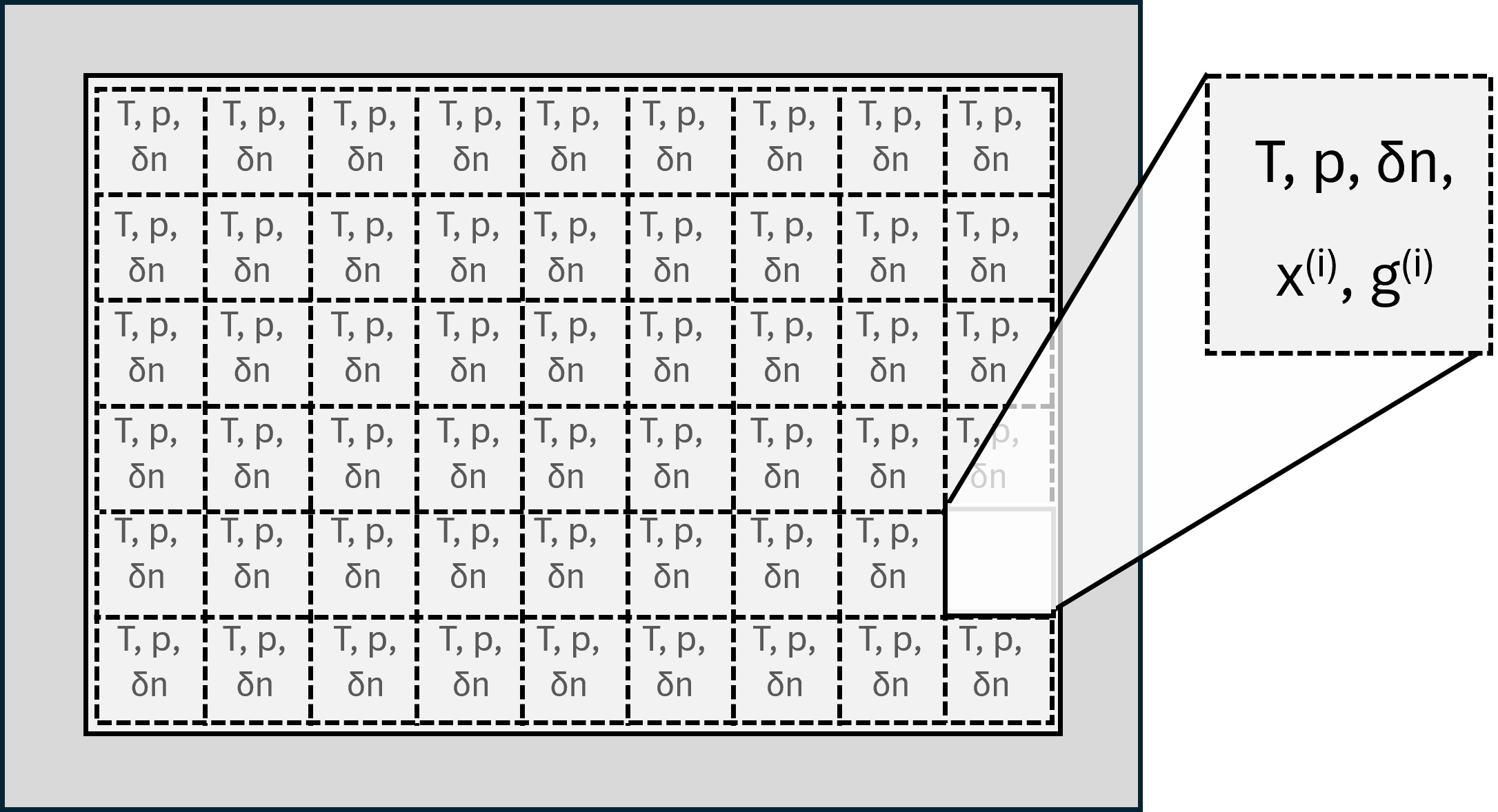}
    \caption{Visualization of the model system considered in this section. The whole system is divided into an (arbitrary) large number $M$ of subsystems, each of them having the same system $T$, $p$ with concentration and Gibbs energy values populated among a pre-defined, discrete set.}
    \label{fig:Ensemble_1}
\end{figure}

\noindent \textbf{Mass balance explicit (Formulation 1).} We consider a closed system at thermal and mechanical equilibrium with its environment.
The system is therefore at fixed temperature $T$, pressure $p$ and concentration $\bar{z} = (z_1, z_2, \ldots, z_n)^T$ and has an overall mole number $n_{tot}$.
Let us assume random fluctuations in the concentration within the system.
Further assuming random fluctuations implies that at any time $t_k$, the system can be divided into $M$ equimolar subsystems, with each subsystem $i=1,...,M$ having, at each time $k$, a local concentration $ \bar{x}^{(i,k)}$, a value of the molar Gibbs energy $g^{(i,k)}$ and mole number $\delta{n} = n_{tot}/M$.
We assume now that each subsystem is allowed to populate a discrete set of $N$ concentrations $ \bar{x}^{(1)},\bar{x}^{(2)},...,\bar{x}^{(N)} $ and respective value of molar Gibbs energy $g^{(1)},g^{(2)},...,g^{(N)}$.
The model system is depicted schematically in Figure~\ref{fig:Ensemble_1}.

The probability $p_{i}$ of finding one subsystem at one of the states $i=1,...,N$ is given by the number of subsystems $M(i,k)$ at the state $i$ at each time $k$ over the total number of subsystems $M$, averaged over time (for a total number of time steps $N_{t}$), according to:
\begin{equation}
p_{i}=\frac{1}{N_{t}}\sum_{k=1}^{N_{t}}{\frac{M(i,k)}{M}}
\label{eq:p_def}.
\end{equation}
For an arbitrary large number of subsystems $M$ or a large number of time steps $N_{t}$, the probability $p_{i}$ tends toward the value that maximizes the Shannon entropy $S$~\cite{Jaynes.1957,Jaynes.1957b}, which is given by Eq. \eqref{eq:shann_ent}~\cite{Shannon.1948}, subject to specific constraints of the considered system.
\begin{equation}
S = - \sum_{i=1}^{N} p_{i} \ln(p_{i})
\label{eq:shann_ent}
\end{equation}
The sought probability distribution must satisfy the normalization condition, mass balance, and a specific mean value of the Gibbs energy. The latter constraint means that the expectation value of the Gibbs energy with respect to the probability distribution equals the experimentally observed macroscopic mean Gibbs energy.
Equation \eqref{eq:shann_ent}, subject to the aforementioned conditions, can be translated into the constrained convex optimization problem given by:

\begin{align*}
\min_{{\bar{p}} \in \mathbb{R}^N} \ & -S(\bar{p})=\sum_{i=1}^{N} p_{i} \ln(p_{i}) \\
\text{s.t.} \ & \sum_{i=1}^{N} p_{i} = 1, \\
& \sum_{i=1}^{N} p_{i} x_{j}^{(i)} = z_{j}, \quad j=1,...,n-1 \\
& \sum_{i=1}^{N} p_{i} g^{(i)} = \bar{g} \\
\end{align*}
and the Lagrangian $\mathcal{L}:X\to\mathbb{R}$ given by the following equation:

\begin{equation}
\mathcal{L}\left(\bar{p},\lambda,\bar{\alpha},\beta\right) = \sum_{i=1}^{N} p_{i} \ln\left(p_{i}\right)+{\lambda}{\left(\sum_{i=1}^{N} p_{i}-1\right)}+\sum_{j=1}^{n-1}{\alpha_{j}\left(\sum_{i=1}^{N} p_{i} x_{j}^{(i)} - z_{j}\right)}+\beta\left(\sum_{i=1}^{N} p_{i} g^{(i)} - \bar{g}\right)
\label{eq:lag_shann}
\end{equation}
The feasible set $X^p \subset \mathbb{R}^N_+$ of the primal variables $p_i$ is convex as it is given as the Cartesian product of convex subsets. The objective function is convex, as it is a sum of $N$ convex terms $p_{i} \ln\left(p_{i}\right)$ in $X^p$, while the constraints are affine functions. The associated Lagrangian given by Eq. \ref{eq:lag_shann} is convex in the primal variables and affine in the Lagrange multipliers. Therefore, under the standard constraint qualifications, the global optimum is characterized by the KKT conditions (e.g. by the stationary condition $\nabla \mathcal{L}=0$) and is obtained as the solution of the system of equations given by Eq. \eqref{eq:lag_shann_pd1}-\eqref{eq:lag_shann_pd4}.

\begin{align}
\frac{\partial \mathcal{L}}{\partial p_{i}} &= 0 = \ln(p_{i})+1+\lambda+\sum_{j=1}^{n-1}{\alpha_{j}x_{j}^{(i)}}+\beta{g^{(i)}} \label{eq:lag_shann_pd1} \\
\frac{\partial \mathcal{L}}{\partial \lambda} &= 0 = \sum_{i=1}^{N} p_{i} - 1, \label{eq:lag_shann_pd2} \\
\frac{\partial \mathcal{L}}{\partial \alpha_{j}} &= 0 = \sum_{i=1}^{N} p_{i} x_{j}^{(i)} - z_{j}, \quad j=1,...,n-1 \label{eq:lag_shann_pd3} \\
\frac{\partial \mathcal{L}}{\partial \beta} &= 0 = \sum_{i=1}^{N} p_{i} g^{(i)} - \bar{g}, \label{eq:lag_shann_pd4}
\end{align}
In Eq. \eqref{eq:lag_shann_pd1}, the Lagrange multiplier $\lambda$ related to the summation condition (Eq. \eqref{eq:lag_shann_pd2}) can be obtained by inserting Eq. \eqref{eq:lag_shann_pd1} into Eq. \eqref{eq:lag_shann_pd2} and is given by 
\begin{equation}
e^{-\lambda - 1} = \sum_{i=1}^{N} {e^{\sum_{j=1}^{n-1}{\alpha_{j}x_{j}^{(i)}}}}{e^{\beta{g^{(i)}}}}
\end{equation}
The Lagrange multiplier $\beta$ is a negative-valued scaling factor which, in statistical thermodynamics, is equal to the negative value of the inverse thermal energy $k_BT$, $\beta = -1/k_BT$, and has the effect of increasing or decreasing the spread of the distribution.
For the scope of this work, the factor $\beta$ corresponds to the (negative valued) inverse of the temperature parameter $\tau$, $\beta = -\tau^{-1}$, which can be used to sharpen the peak of the resulting distribution. In our derivation (as in the classical derivation of the canonical partition function), the expected mean value of $\bar{g}$ is imposed by the value of $\tau$. The resulting $\bar{g}$, calculated as the expectation of the given distribution $\bar{p}_{\tau}$ decreases monotonically with decreasing $\tau$, and returns the (global) minimum of $\bar{g}$ in the limit of $\tau \xrightarrow[]{}0$.
Therefore, we are left with a probability distribution $p_i$ which is a function of the parameters $\tau$ and $\alpha_{i}$ and is given by: 
\begin{equation}
p_{i} = \frac{{e^{\sum_{j=1}^{n-1}{\alpha_{j}x_{j}^{(i)}}}}{e^{-\frac{g^{(i)}}{\tau}}}}{{\sum_{k=1}^{N}e^{\sum_{j=1}^{n-1}{\alpha_{j}x_{j}^{(k)}}}}{e^{-\frac{g^{(k)}}{\tau}}}} \quad i=1,...,N \label{eq:final_dist}
\end{equation}

While $\tau$ can be freely adjusted, the set of $n-1$ parameters $\alpha_{i}$ must be determined iteratively such that the distribution $p_i$ satisfies the mass balance given by Eq. \eqref{eq:lag_shann_pd3} for the given feed composition $\bar{z}$.
This is accomplished by solving the system of equations given by: 
\begin{equation*}
F_{m}(\bar{\alpha}) = \sum_{i=1}^{N}{\frac{x_{m}^{(i)}{e^{\sum_{j=1}^{n-1}{\alpha_{j}x_{j}^{(i)}}}}{e^{-\frac{g^{(i)}}{\tau}}}}{{\sum_{k=1}^{N}e^{\sum_{j=1}^{n-1}{\alpha_{j}x_{j}^{(k)}}}}{e^{-\frac{g^{(k)}}{\tau}}}}}-z_{m}=0 \quad m=1,...,n-1 \\
\end{equation*}
which for a binary system reduces to:
\begin{equation*}
F(\alpha) = \sum_{i=1}^{N}{\frac{x_{1}^{(i)}{e^{\alpha x_{1}^{(i)}}}{e^{-\frac{g^{(i)}}{\tau}}}}{{\sum_{k=1}^{N}e^{\alpha x_{1}^{(k)}}}{e^{-\frac{g^{(k)}}{\tau}}}}}-z_{1}=0
\end{equation*}

\begin{figure}[htbp]
    \centering
    \includegraphics[width=0.6\textwidth]{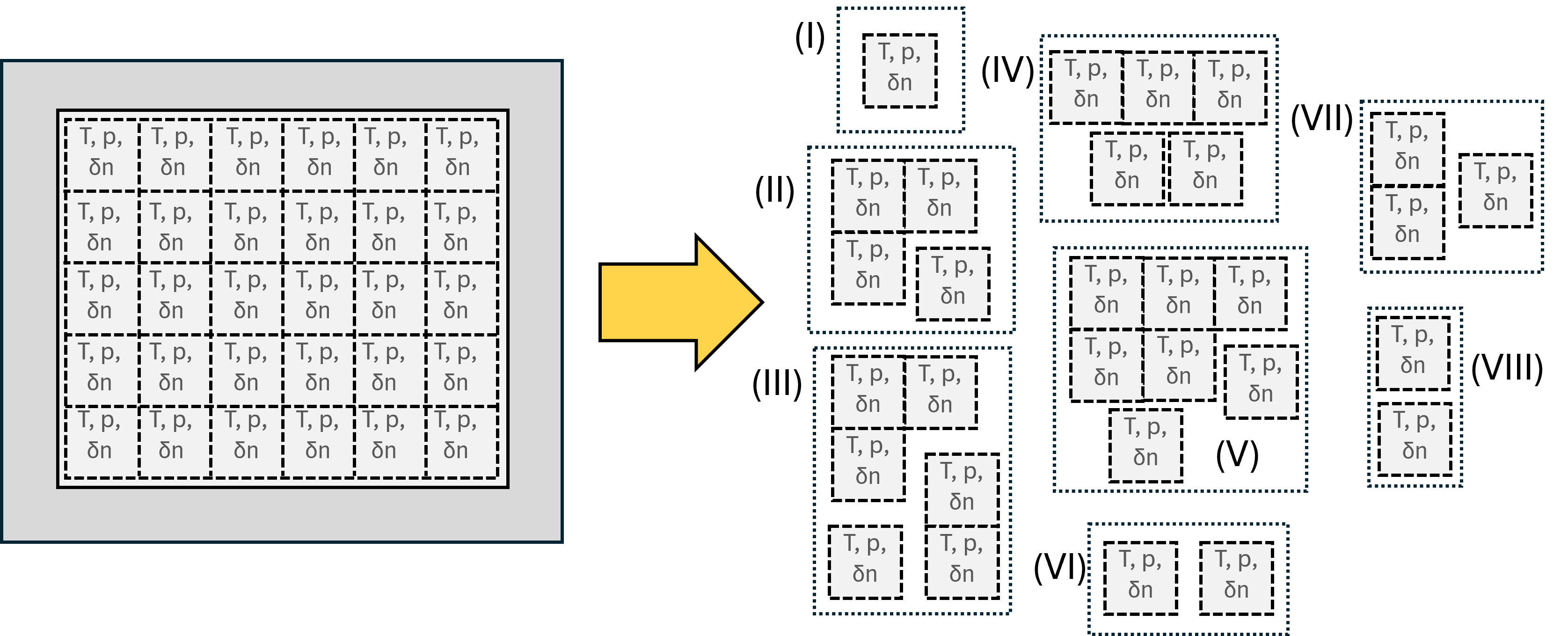}
    \caption{Visualization of the concept behind the second formulation of our method. The whole set of $M$ subsystems (left picture) can be constructed as a sum of groups consisting of fewer subsystems (right picture), with each group having the same overall composition $\bar{z}$ as the whole system.}
    \label{fig:Ens_decompos}
\end{figure}

\noindent \textbf{Mass balance implicit (Formulation 2).} To avoid the iterative step of determining the unknown parameter $\alpha$ from the mass balance condition, an alternative formulation of the problem depicted in Figure~\ref{fig:Ensemble_1}, which avoids the mass balance condition, can be defined by grouping subsystems together such that each group has the same concentration $\bar{z}$ as the entire system.
This procedure of decomposing the system into groups of composition $\bar{z}$ is depicted schematically in Figure~\ref{fig:Ens_decompos}, where each group is represented by roman numbers. 

In principle, any system composed of $M$ subsystems, each of which may occupy one of $N$ available states at fixed temperature $T$, pressure $p$, and composition–energy pairs $\bar{x}^{(i)},g^{(i)}=g(\bar{x}^{(i)})$ for $i=1,...,N$ can be constructed from a combination of any number of groups which belong to a class $\Gamma(\bar{z})$, according to:
\begin{equation*}
\Gamma(\bar{z})=\Gamma^{(1)}(\bar{z})+\Gamma^{(2)}(\bar{z})+...+\Gamma^{(n)}(\bar{z}), \\ %\label{eq:Class_Gamma} \\
\end{equation*}
i.e,, the class $\Gamma(\bar{z})$ consists of $n$ subclasses $\Gamma^{(k)}(\bar{z})$ indexed by $k=1,...,n$.
Each subclass $\Gamma^{(k)}(\bar{z})$ contains all possible combination of $k$ states that yields a system with overall composition $\bar{z}$, subject to the condition that no such $k$-tuple includes a sub-combination already represented in one of the lower-order subclasses $\Gamma^{(1)}(\bar{z})+...+\Gamma^{(k-1)}(\bar{z})$.
More explicitely, a group of $k$ states belongs to $\Gamma^{(k)}(\bar{z})$ if there exists a set of positive real numbers $\phi^{(k)}(\bar{z})$ depending on the system composition $\bar{z}$ and satisfying the following two Equations: 
\begin{equation}
\ \sum_{i=1}^{k}\phi^{(i)}=1  \label{eq:Phi_summ_cond} 
\end{equation}
\begin{equation}
\ \sum_{i=1}^{k}\phi^{(i)}\bar{x}^{(i)}=\bar{z}  \label{eq:Phi_mass_bal} 
\end{equation}
such that no subset of these $k$ states already qualifies as a member of a lower-order class under the same criterion.
As an example, consider a ternary system.
In this case a triplet of states $(\bar{x}^{(1)},g^{(1)}),(\bar{x}^{(2)},g^{(2)}),(\bar{x}^{(3)},g^{(3)})$ is excluded from $\Gamma^{(3)}(\bar{z})$ if the three compositions lie on a straight line through the feed composition $\bar{z}$, because at least two of the pairs $(1-2)$,$(1-3)$ or $(2-3)$ already belongs to the subclass $\Gamma^{(2)}(\bar{z})$.
Similarly, a binary system only contains the subclasses $\Gamma^{(1)}(\bar{z})$ (non-empty only if a state $k$ coincides with the feed composition $\bar{z}$) and the class $\Gamma^{(2)}(\bar{z})$.
As a procedure to verify whether a set of $k$ states $\bar{x}^{(1)},...,\bar{x}^{(k)}$ with $k<n$ belongs to the subclass $\Gamma^{(k)}(\bar{z})^{(k)}$ is first to proof that each composition is linearly independent, which means that the matrix $\bar{\bar{X}}^{(k)}$ containing each state composition $\bar{x}^{(j)}$ as columns and given by: 
\begin{equation*}
\bar{\bar{X}}^{(k)} =
\begin{bmatrix}
x_{1}^{(1)} & x_{1}^{(2)} & \cdots & x_{1}^{(k)} \\
x_{2}^{(1)} & x_{2}^{(2)} & \cdots & x_{2}^{(k)} \\
\vdots & \vdots & \ddots & \vdots \\
x_{n}^{(1)} & x_{n}^{(2)} & \cdots & x_{n}^{(k)} 
\end{bmatrix}
\end{equation*}
has $rank(\bar{\bar{X}})=k$.
If this is the case, then no subset of $k$ already belongs to a lower-order subclass.
As a second step, the set $\bar{\phi}(\bar{z})$ can be determined as solution of the linear least-square problem given by: 
\begin{equation*}
\ \bar{\phi}=(((\bar{\bar{X}})^{T})\bar{\bar{X}})^{-1}(\bar{\bar{X}})^{T}\bar{z}
\end{equation*}

If the array $\bar{\phi}$ satisfies Eq. \eqref{eq:Phi_summ_cond} - \eqref{eq:Phi_mass_bal} and the value of each element of $\bar{\phi}$ is in the interval $]0,1[$ then the set $k$ belongs to the subclass $\Gamma^{(k)}(\bar{z})$.
Starting from $N$ states, this procedure can be used to populate the subclasses by defining all possible groups in each subclass and then verifying if each group qualifies as subclass member.
For a subclass of order $k$, the number of possible groups is given by all the ways of choosing $k$ states from the $N$ total states, irrespective of their permutation, according to the general formula for the binomial coefficient given by: 
\begin{equation}
\ \binom{N}{k} = \frac{N!}{k!(N-k)!} \label{eq:Bin_coeff}
\end{equation}
which therefore represents an upper limit of the number of groups in subclass $k$.
Each group $j$ in each subclass $\Gamma^{(k)}(\bar{z})$ has a value of the Gibbs energy $g^{(j,k)}$ which is given as linear combination of the Gibbs energies $g^{(j)}$ of each of its states via the vector $\bar{\phi}^{(j,k)}$, according to: 
\begin{equation*}
\ g^{(j,k)}=\sum_{i=1}^{k}\phi^{(i,k)}g^{(i)} % \label{eq:g_groups} \\
\end{equation*}
The probability $p_S$ of finding group $S \in \Gamma(\bar{z})$ in the system is now given as the solution of the following optimization problem:
\begin{align*}
\min_{{\bar{p}} \in \mathbb{R}^{|\Gamma(\bar{z})|}} \ & -S(\bar{p})=\sum_{s \in \Gamma(\bar{z})} p_{s} \ln(p_{s}) \\
\text{s.t.} \ & \sum_{s \in \Gamma(\bar{z})} p_{s} = 1, \\
& \sum_{s \in \Gamma(\bar{z})} p_{s} g^{(s)} = \bar{g} \\
\end{align*}
The solution of the previous optimization problem is the Boltzmann distribution of each defined subgroup $m$ in the class $\Gamma (\bar{z})$, which is given by: 
\begin{equation}
\ p_{m}=\frac{e^{-\frac{g^{(m)}}{\tau}}}{\sum_{s \in \Gamma(\bar{z})}e^{-\frac{g^{(s)}}{\tau}}} \label{eq:Boltzmann_groups}
\end{equation}
with $s \in \Gamma(\bar{z})$ in the previous equation referring to any group in class $\Gamma (\bar{z})$.

\section{Single System Case Studies}
\label{sec:SingSysCS}
\subsection{Case Study 1: Binary LLE}
\begin{figure}[htbp]
    \centering
    \includegraphics[width=0.8\textwidth]{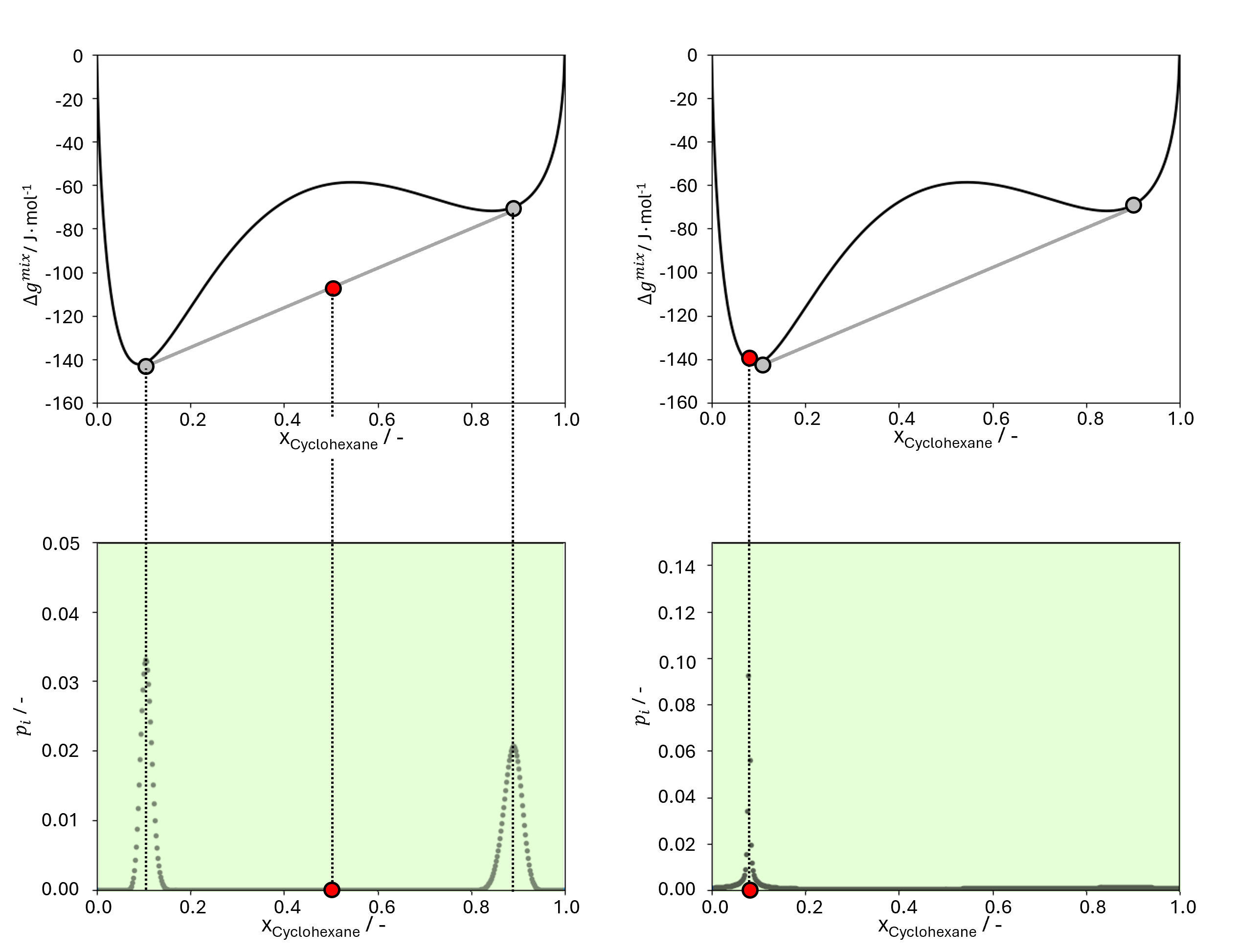}
    \caption{Predicted discrete probability distributions of the states defined by the discretization (bottom diagrams) for a homogeneous (right diagrams) and heterogeneous feed (left diagrams), and comparison with the $\Delta g^{\textrm{mix}}$ curve and \gls{lle} calculated by the PC-SAFT EoS (upper diagrams).}
    \label{fig:CS1_FinPict}
\end{figure}

The potential of the method will be shown against the simple case study of predicting the \gls{lle} of the binary system methanol-cyclohexane at T = 298.15 K and p = 5 bar.
The thermodynamic behavior of this system is described using the EoS PC-SAFT with pure-component and binary interaction provided by \cite{rehner2023feos}.
Figure ~\ref{fig:CS1_FinPict} shows the \gls{gmix} of the system as a function of the Cyclohexane concentration $x_{Cyclohexane}$ as well as the discrete state probability distribution $p_i$ over the Cyclohexane concentration as well.
For the calculation of the $\bar{p} = p_{1},...,p_{N}$, an equidistant grid of $N=500$ points was defined.
Each probability $p_i$ can be calculated either by Method 1 (directly by Eq.~\eqref{eq:final_dist}) or by Method 2 (indirectly from the probability of each group of states $p_m$ given by Eq.~\eqref{eq:Boltzmann_groups}), with only a marginal difference between the methods.

Figure~\ref{fig:CS1_FinPict} shows that the predicted discrete probability distribution is very sharply peaked around the (single or multiple) concentrations at equilibrium.
For concentrations outside the miscibility gap, the probability distribution is unimodal around the feed concentration (right diagram), while inside the miscibility gap, the predicted state distribution becomes bimodal (left diagram) with each distribution peaked at each phase composition and an area below each distribution which is proportional to the molar amount of each phase.
By using Formulation 1, the composition of each phase can be determined by clustering each distribution and calculating separate mean concentrations.
Formulation 2, on the other hand, allows direct evaluation of the concentration of both phases by directly applying Eq.~\eqref{eq:Boltzmann_groups} since each phase composition is already clustered in the definition of the binary groups $\Gamma^{(2)}(\bar{z})$.

\begin{figure}[htbp]
    \centering
    \includegraphics[width=0.5\textwidth]{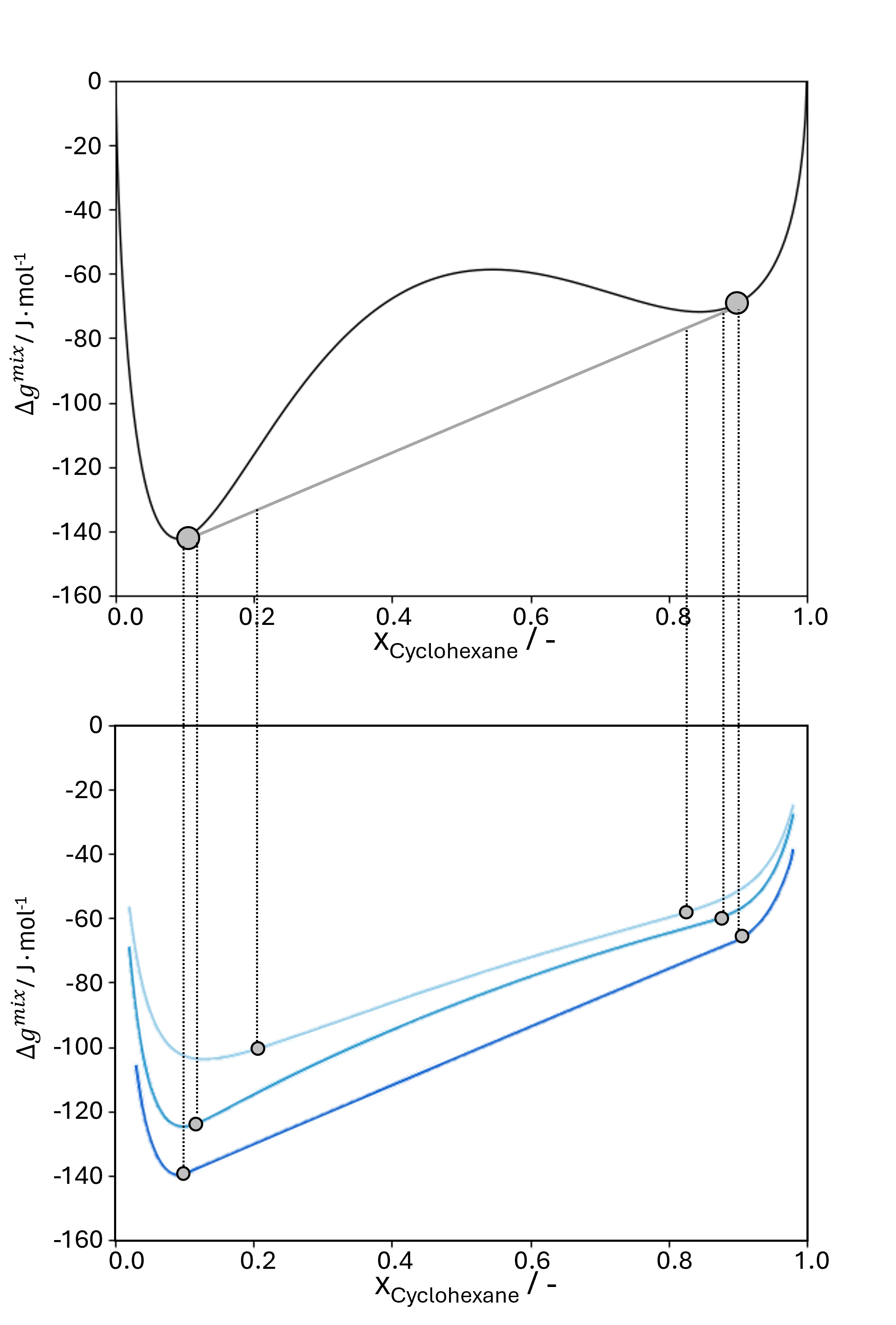}
    \caption{Effect of increasing the temperature parameter $\tau$ on the reconstructed $\Delta g^{\textrm{mix}}$ curve and on the predicted \gls{lle} (bottom diagram), compared with the $\Delta g^{\textrm{mix}}$ and \gls{lle} predicted by the PC-SAFT EoS (upper diagram).
Descriptions of the curves are reported in the main text.}
    \label{fig:CS1_FinPict_2}
\end{figure}
 
Figure~\ref{fig:CS1_FinPict_2} shows the effect of using different temperature parameters $\tau$ on the accuracy of the model in reconstructing the  $\Delta g^{\textrm{mix}}$ curve and on the accuracy of the calculated \gls{lle}.
The calculations are performed using 100 discretization points using the three temperature parameters $\tau = 0.0005, 0.005$ and $0.05$, which correspond, respectively, to the blue, cyan blue and pale blue curves.
Each curve is calculated, for a given feed concentration $z$, as mean-average of the $\Delta g^{\textrm{mix}}_k$ of each state according to Eq.~\eqref{eq:Deltag_mix_aver} by varying the feed concentration between $z=0$ (pure methanol) and $z=1$ (pure cyclohexane).

\begin{equation}
\Delta g^{\textrm{mix}} = \sum_{k=1}^{N} p_{k}\Delta g^{\textrm{mix}}_k\label{eq:Deltag_mix_aver}
\end{equation}

One interesting particularity of the method, which is exemplarily shown by Figure~\ref{fig:CS1_FinPict_2}, is that the re-calculated expectations of the $\Delta g^{\textrm{mix}}$ curve are always convex in the limit of $\tau \to 0$, regardless of the mathematical form of the original curve.
Within a miscibility gap, the calculated $\Delta g^{\textrm{mix}}$ scales almost linearly between both phases.
At the same time, one can observe an increase in accuracy at lower $\tau$ values, due to the fact that the contribution of states away from the mean value decreases.
The accuracy of the predicted \gls{lle} composition remains high even when the absolute value of the calculated $\Delta g^{\textrm{mix}}$ is low (as is the case at high $\Delta g^{\textrm{mix}}$ values) and increases with decreasing $\Delta g^{\textrm{mix}}$.

\subsection{Case Study 2: Three-Phase LLLE}
\label{sec:CS2}
\begin{figure}[htbp]
    \centering
    \includegraphics[width=0.7\textwidth]{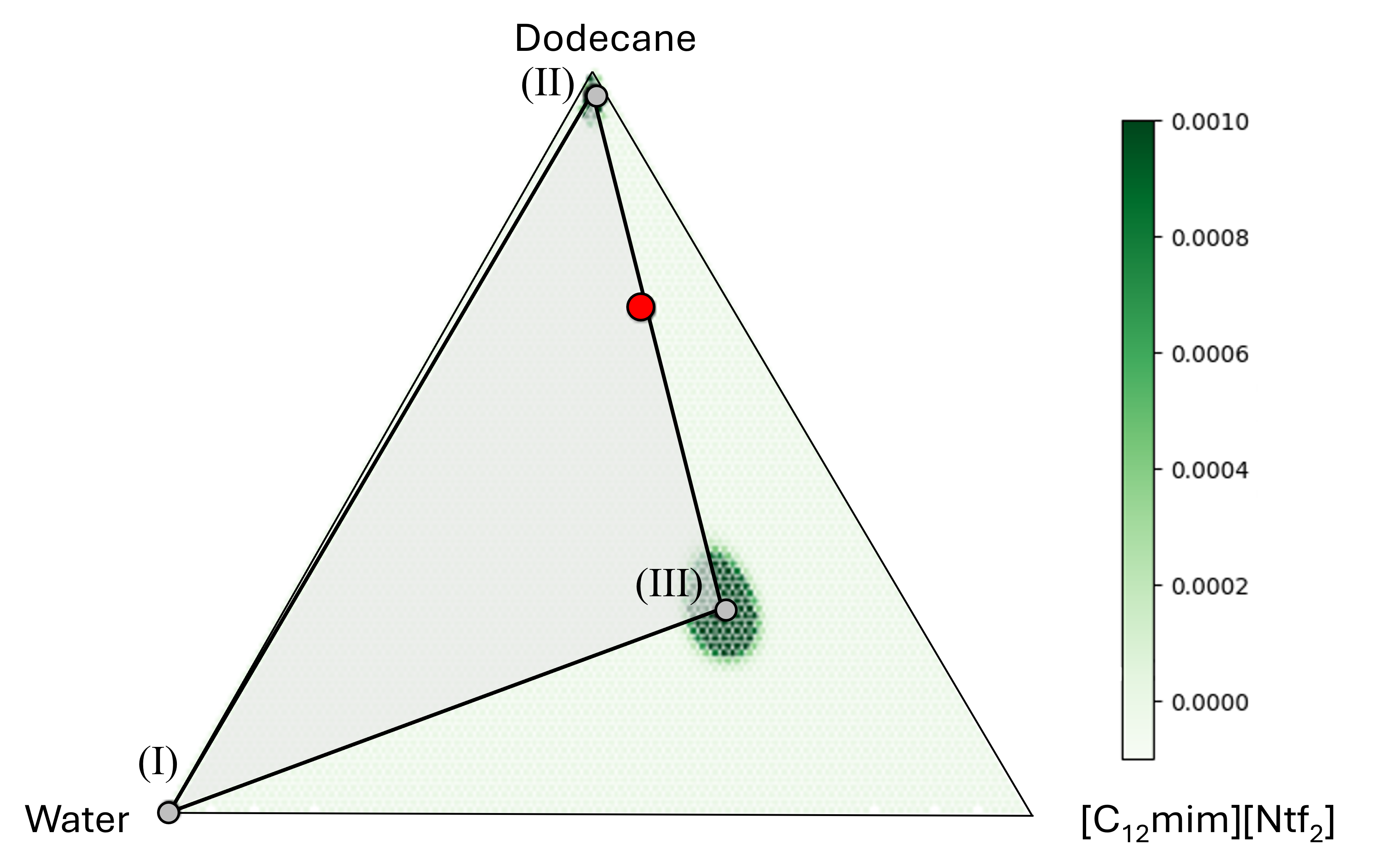}
    \caption{Ternary diagram showing the three-phase LLLE equilibrium of the system water-dodecane-[$\text{C}_{12}\text{mim}$][$\text{Ntf}_2$], calculated using our approach (method 1) as well as the resulting state distribution as a color map over the diagram.}
    \label{fig:CS2_LLE_p_srfc}
\end{figure}

In this example, we will show how our method can be used to predict the phase equilibrium in a ternary system that undergoes a three-phase liquid-liquid-liquid equilibrium (LLLE).
The ternary system under consideration is water-dodecane-[$\text{C}_{12}\text{mim}$][$\text{Ntf}_2$] with the latter component being an imidazolium-based ionic liquid.
The phase equilibrium of this system was measured by Rodríguez-Palmeiro \textit{et al.}~\cite{rodriguez2015measurement} at $T=298.15$ K and $p=1$ bar.
Under certain overall compositions, the mixture separates into three liquid phases: \text{(I)} an aqueous phase composed almost entirely of water,\text{(II)} an organic phase consisting predominantly of dodecane, and \text{(III)} an ionic-liquid-rich phase that also contains water and dodecane in comparable amounts.
The same authors also provided the pure components and binary interaction parameters of PC-SAFT required to calculate the phase behavior.
Using Method 1 and with a $100 \text{ x } 100$ discretization grid (corresponding to 5044 points), an estimate of the equilibrium composition of each phase was calculated.
For a feed composition of $\bar{z} = \text{(}0.11,0.68,0.21\text{)}$, a tri-modal probability distribution $p_i$ is predicted.
The composition of each phase was calculated by clustering each distribution and calculating separate average compositions.
Figure~\ref{fig:CS2_LLE_p_srfc} shows the probability distributions, depicted by different color intensity over the whole composition space, as well as the estimated compositions of each of the three phases \text{(I) - (III)} (represented by gray points) and the used feed composition (red point).

This example shows the versatility of the method in estimating different topologies of phase equilibrium by correctly recognizing the number of stable phases, without the necessity of defining case-specific formulations or performing separate stability checks.
On the other hand, the number of grid points sharply increases with increasing number of components, questioning the choice of equidistant grids with more than 2 components.

\subsection{Case Study 3: Vapor pressure calculation}
\label{sec:CS3}
The principles of our methods, explained in this section, maintain their validity when working in other ensembles, such as the $npT$.
Consider, for example, the problem of determining the vapor pressure of a pure fluid at given temperature $T$.
Working on the canonical ($nVT$) ensemble, the $TV$-flash is equivalent to finding the equilibrium condition (number of phases, usually one or two, and their density) which, at given temperature $T$ and molar volume $v$, minimizes the Helmholtz energy $A$ of the system~\cite{michelsen1999state}.
For a two-phase solution, the vapor pressure is then equal to the slope of the double tangent, which touches both phases in the $A-v$ plot and is given by Eq.~\eqref{eq:P_from_Tang}~\cite{winterBARDOWUnderstandingLanguageMolecules2025}.

\begin{equation}
\ p = -\frac{A(V_{L})-A(V_{V})}{V_{L}-V_{V}} \label{eq:P_from_Tang}
\end{equation}

\begin{figure}[htbp]
    \centering
    \includegraphics[scale=0.7]{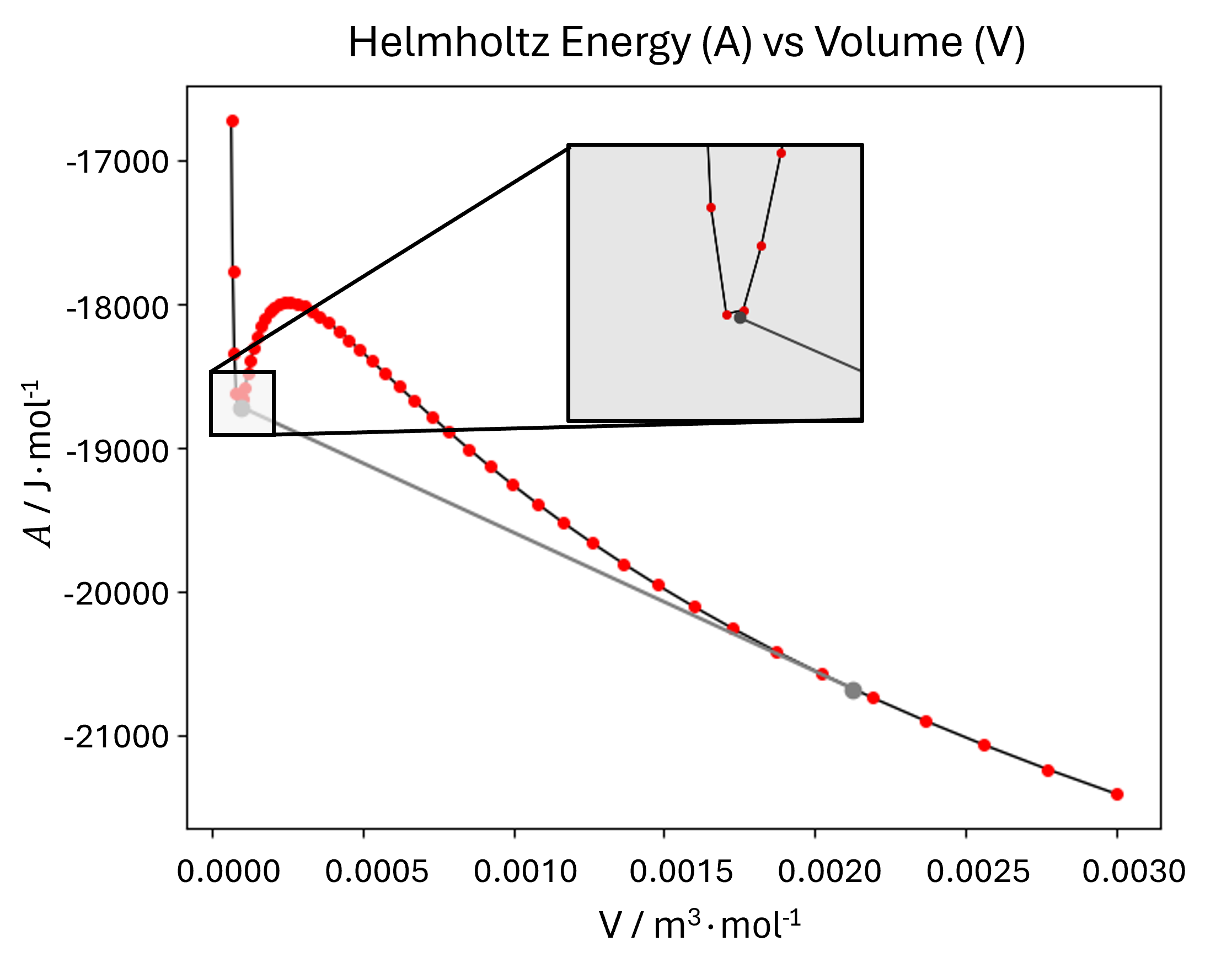}
    \caption{Helmholtz energy curve of subcritical propane, calculated with PC-SAFT, and the resulting phase coexistence and vapor pressure determined using our approach (gray points connected by a gray line). The vapor pressure is given by the negative slope of the gray line.}
    \label{fig:CS3_tot}
\end{figure}

The same derivation can be performed by considering fluctuations in the molar volume $v$ and the local Helmholtz energy $A$.
Discrete $(v_k,A_k)$ points can be defined, and the probability distribution $p_i$ can be defined by maximization of Eq.~\eqref{eq:shann_ent}.
Figure~\ref{fig:CS3_tot} shows the calculation of the vapor pressure of pure propane at $T=298.15 K$, employing our method, using PC-SAFT to describe the phase behavior of propane and parameters taken from~\cite{gross2001perturbed}. 100 discretization points have been defined, from which a set of binary states has been defined.

Using a temperature parameter $\tau=30$, the volume and Helmholtz energy of both phases can be estimated.
The resulting pressure, calculated using Eq.~\eqref{eq:P_from_Tang}, is $p=9.577 \text{ bar}$.
The rigorous calculation of the vapor pressure of propane, using the same set of PC-SAFT parameters, gives $p = 9.543 \text{ bar}$ at the same temperature, which represents a deviation of less than $0.4 \%$.
This simple calculation showcases the potential of this method for accurate estimation of VLE while avoiding numerical procedures.

\section{Dataset Generation}
To compute equilibrium composition labels for our list of binary systems, we compute discretized \gls{gmix} curves for each binary system and determine the corresponding phase splits using the DISCOMAX forward pass procedure.
For every system, a composition grid of 401 equally spaced points in the interval $[0+\epsilon,\,1-\epsilon]$ with $\epsilon=10^{-8}$ is constructed.  
At each composition $x$, \gls{gmix} is evaluated, with \gls{gE} predicted using the HANNA~2 model (with $H^{\text{ex}}$ data) at $T = 298.15\,\mathrm{K}$.

To estimate a representative feed composition, we compute the second derivative of the discretized \gls{gmix} curve w.r.t. composition ($\frac{\partial^2\Delta g_{\text{mix}}}{\partial x^2}$)
 using finite differences and identify regions of negative curvature~\cite{mitsos2009bilevel}.  
The equilibrium phase compositions $(x', x'')$ are then obtained by minimizing the discretized \gls{gmix} using the DISCOMAX algorithm described in the main text.  
If $|x' - x''| < 10^{-3}$, the system is considered single-phase and no phase split is recorded.
The method is simple and very fast, but not rigorous strictly speaking, since it does not use deterministic global optimization.
After visually inspecting many systems, we did not find any case where the method failed to identify the correct phase split.
The full code for this procedure is also available in the repository. 
The procedure is repeated for all $\sim$140,000 binary systems in our dataset, yielding the final set of equilibrium composition labels.

\section{Surrogate Solver Implementation}
Because the implementation details and training code of the surrogate solver presented by Hoffmann et~al.~\cite{hoffmannJIRASEKMachinelearnedExpressionExcess2025} are not publicly available, we re-implemented this baseline independently.
Their approach relies on a surrogate model trained on mod. UNIFAC-generated \gls{gmix} data. 
Even when the mod. UNIFAC data are split into distinct training and test partitions for training the surrogate; data leakage remains unavoidable: the mod. UNIFAC parameterization itself was fitted using a wide range of molecules, many of which also appear in the test set.
Consequently, the HANNA-style surrogate solver variants all possess prior information about the test systems before any training takes place, giving them an intrinsic advantage.

To further reduce the need for retraining the solver for every fold, we deliberately place the surrogate solver at an even stronger advantage by training it on \emph{all} mod. UNIFAC-accessible systems rather than attempting to construct an artificial split on an already leaked dataset.
This may yield overconfident test performance estimates, but is acceptable for this work, since the surrogate solver only serves as a baseline.

The dataset for training the surrogate solver is generated as follows.
We evaluate all ${\sim}140{,}000$ binary mixtures in our full dataset and identify those for which mod. UNIFAC predictions are feasible, i.e., systems composed exclusively of functional groups supported by the mod. UNIFAC model.
For each such mixture, we compute discretized \gls{gmix} curves over 401 equally spaced compositions in the interval $[0+\epsilon,1-\epsilon]$, and for temperatures ranging from $280\,\mathrm{K}$ to $380\,\mathrm{K}$.
Using the DISCOMAX discretized \gls{gmix} minimization procedure described in the main Algorithm, and relying solely on the thermodynamically exact discrete $\arg\min$, we obtain the corresponding phase splits for each system.
We first create a mask for where \gls{gmix} is negatively curved, and then place the feed composition in the center of that region, assuming that only one major two-phase region exists.
We used the same method for generating the training dataset.
The surrogate solver itself predicts binodal phase compositions directly from a \gls{gmix} curve and therefore does not require a feed composition for isolated tie-line prediction.
However, if these predicted phase compositions are used in a differentiable flash or process calculation (where the feed composition is specified), it can also not guarantee that the predicted phase compositions are mass balance-feasible.

From the initial set of UNIFAC-compatible mixtures, we are able to generate \gls{gmix} curves for 82,215 systems, of which 5,224 exhibit a liquid-liquid split.
This yields the complete surrogate-solver training dataset.

The surrogate model itself is a simple order-invariant multilayer perceptron (MLP).
Order invariance with respect to the 101 discretized compositions is enforced by evaluating the model on the original order and output $x$, and reversed order and output $1-x$, with the final prediction obtained by averaging these two outputs.
The network architecture closely follows~\cite{hoffmannJIRASEKMachinelearnedExpressionExcess2025}, consisting of three hidden layers with ReLU activations and a final sigmoid output layer.
Training is performed using the Adam optimizer together with a OneCycle learning rate schedule, with a maximum learning rate of $0.01$.
The model is trained for up to 200 epochs with early stopping.
Using this setup, we obtain a mean absolute error of approximately $\mathrm{MAE} = 0.008$, $\mathrm{RMSE} = 0.024$, and $R^2 = 0.996$ on the validation dataset.
This is comparable to the reported $\mathrm{MAE} = 0.004$ by ~\cite{hoffmannJIRASEKMachinelearnedExpressionExcess2025}, but their surrogate dataset is not publicly available, so a direct comparison is not possible.

\section{Supplementary Results}
\subsection{Single-system fitting diagnostics}
\begin{figure}[htbp]
\centering
    \includegraphics[width=1.\textwidth]{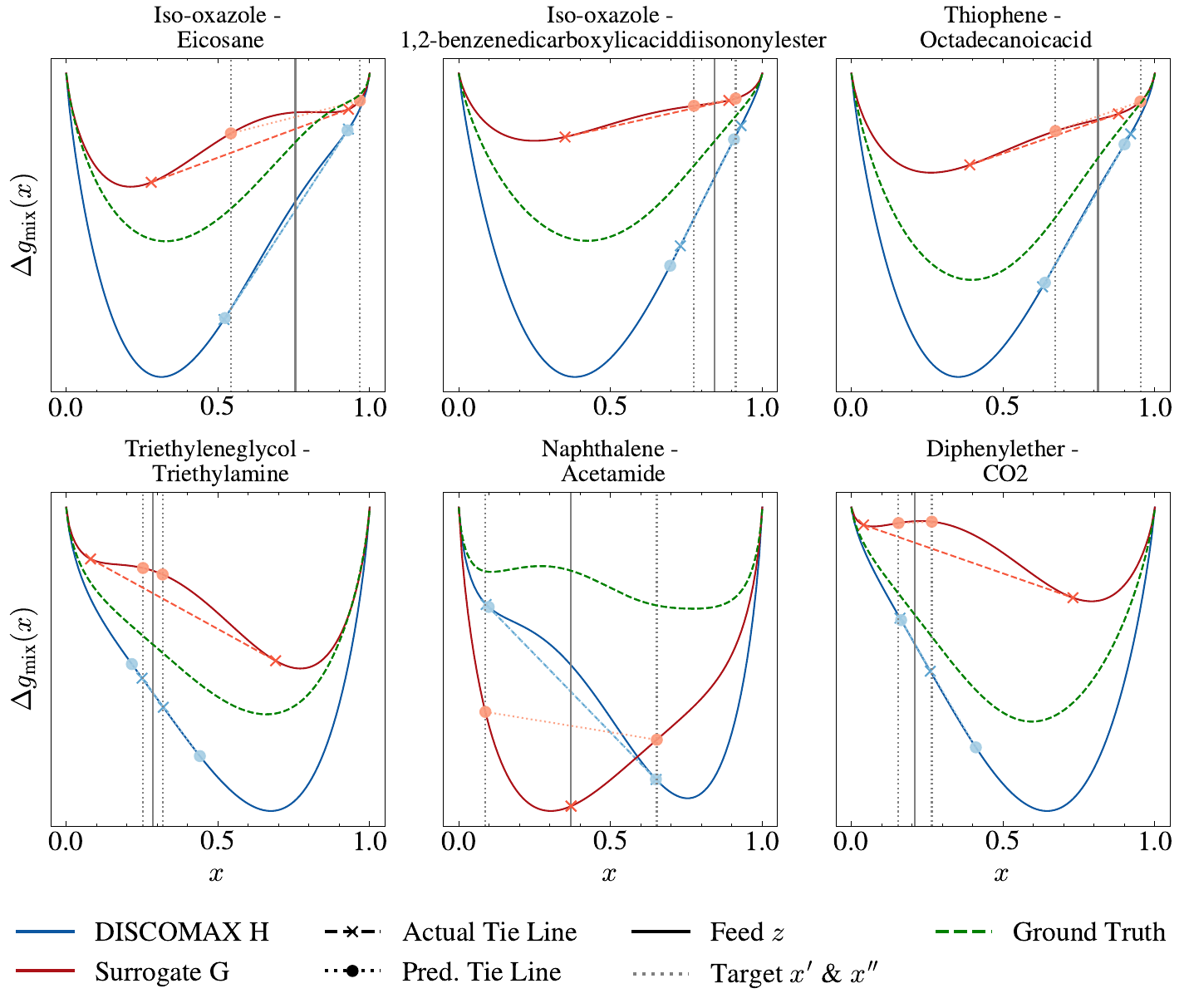}
        \caption{Three largest-error single-system fits for DISCOMAX~H (top row) and Surrogate~G (bottom row), with the prediction of the other solver overlaid on the same system. Blue and red markers denote DISCOMAX~H and Surrogate~G predictions, respectively, and green markers denote the target phase compositions.}
    \label{fig:single-sys-sytems-worst}
\end{figure}

Figure~\ref{fig:single-sys-sytems-worst} provides a qualitative comparison of the fitted phase-composition profiles for the best-performing configuration of each solver on the most difficult systems for each model.
The DISCOMAX~H predictions are selected through the differentiable discretized minimization, whereas the Surrogate~G baseline directly regresses the phase-compositions from the predicted \gls{gmix} curve.

\begin{figure}[htbp]
    \centering
    \begin{minipage}{0.49\textwidth}
        \centering
        \includegraphics[width=\textwidth]{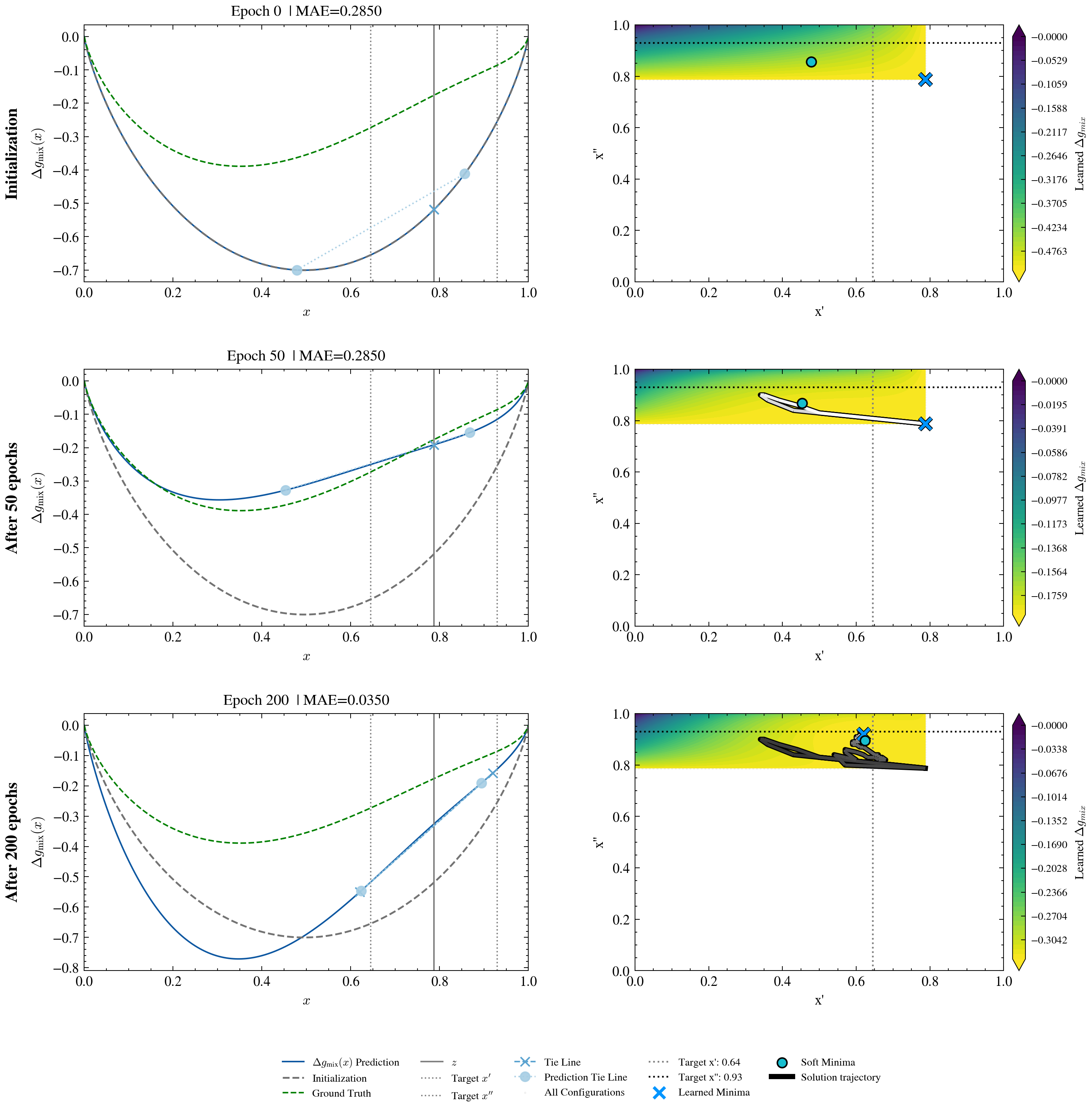}
    \end{minipage}
    \hfill
    \begin{minipage}{0.49\textwidth}
        \centering
        \includegraphics[width=\textwidth]{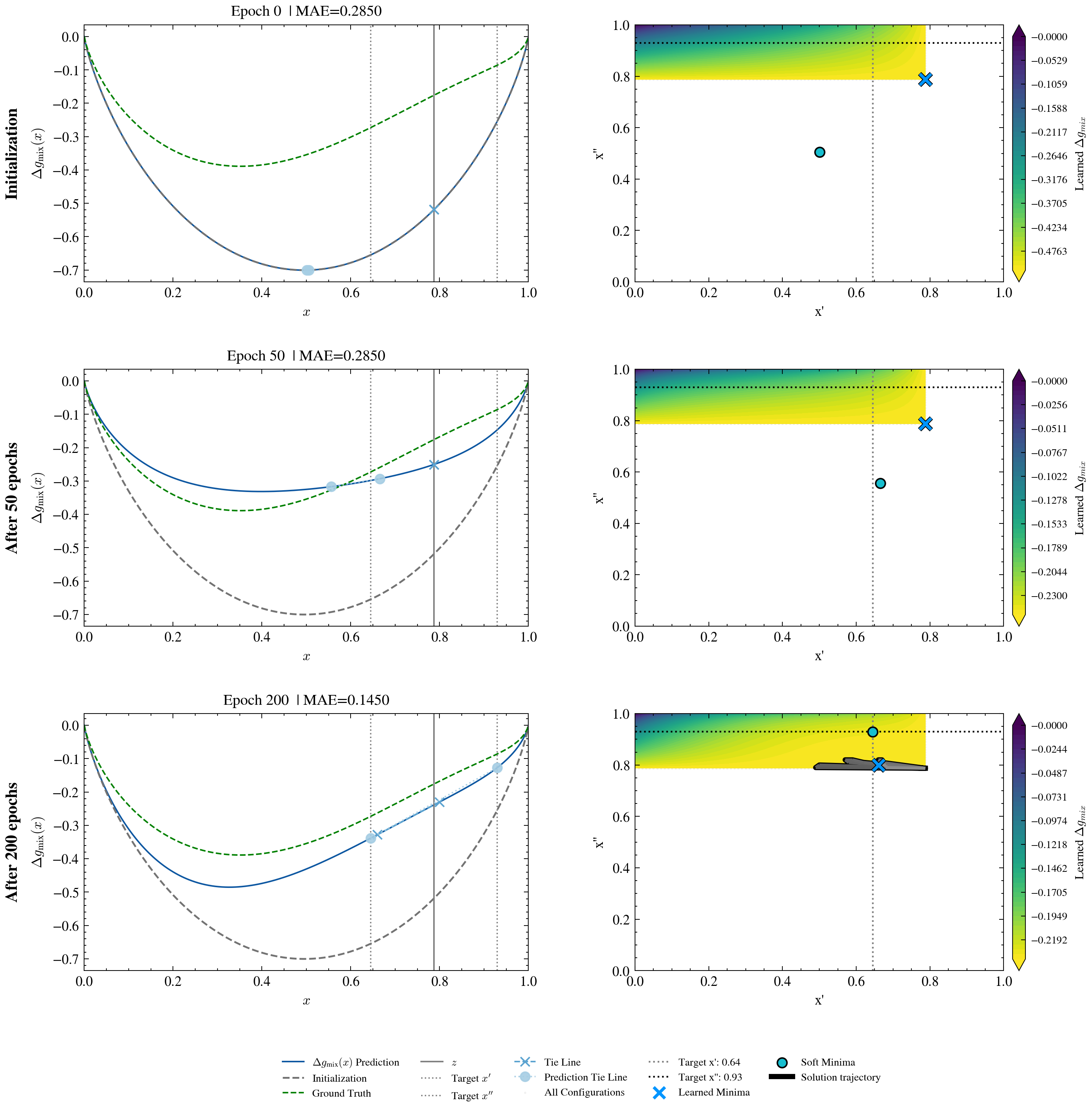}
    \end{minipage}
    \caption{Representative challenging single-system optimization trajectory for DISCOMAX~H (left) and Surrogate~G (right). Left panels show the learned \gls{gmix} profile and the corresponding phase-composition predictions during training. Right panels show all corresponding mass balance-feasible phase splits color-coded by the predicted \gls{gmix}. Hard argmin is shown with a cross, and soft argmin with a circle. The target phase compositions are shown as dotted lines.}
    \label{fig:single-system-trajectory}
\end{figure}

Figure~\ref{fig:single-system-trajectory} shows representative challenging single-system optimization trajectories for DISCOMAX~H and Surrogate~G.

\begin{figure}[htbp]
    \centering
    \includegraphics[width=1.\textwidth]{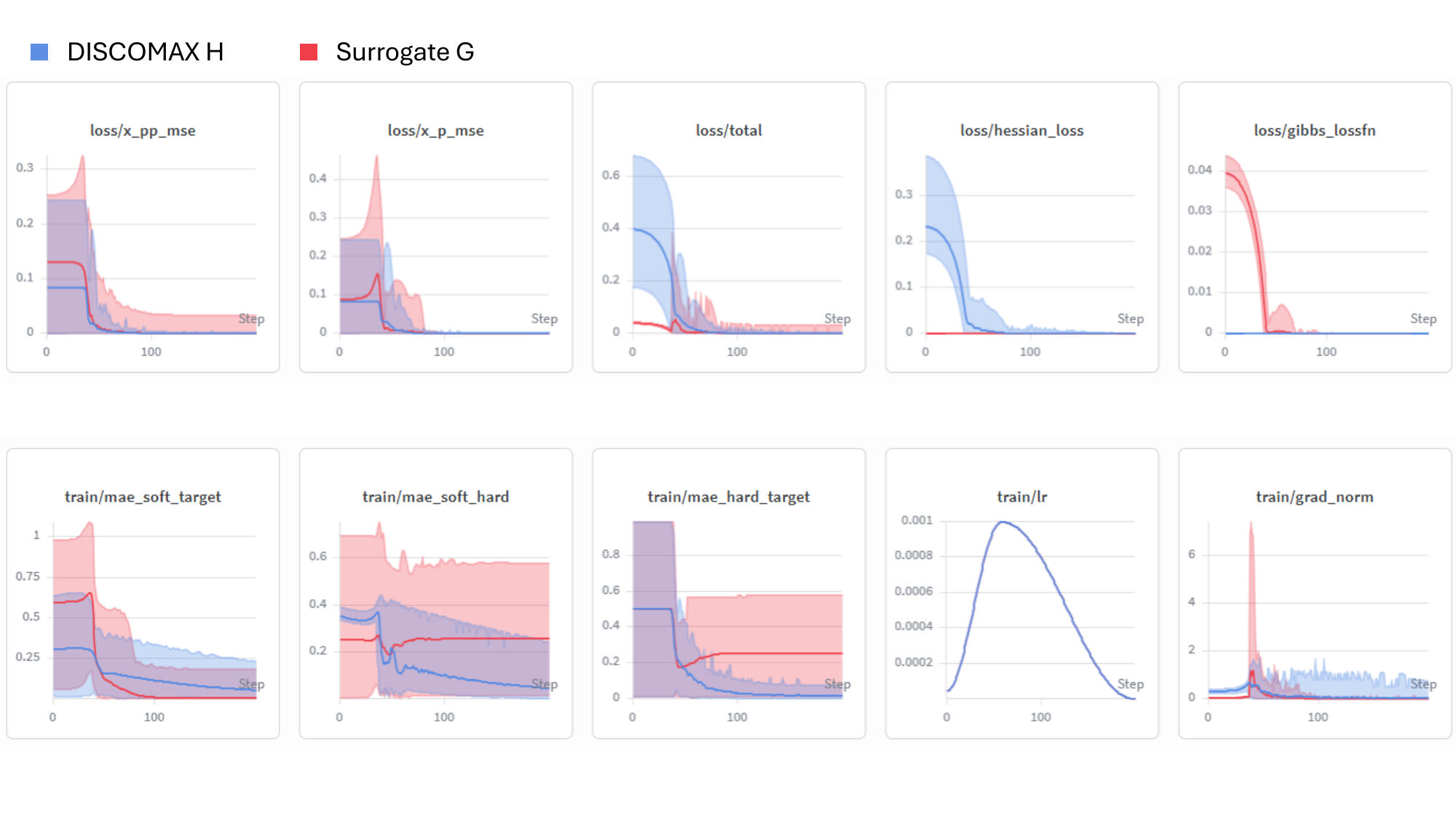}
    \caption{Single-system training diagnostics for DISCOMAX~H and Surrogate~G, including differentiable-output--target MSE terms ($x',x''$), Gibbs and Hessian losses, hard-target MAE, differentiable-output--hard-argmin discrepancy, learning rate, and gradient norm of the model weights. For DISCOMAX~H, the differentiable output is the soft argmin; for Surrogate~G, it is the learned surrogate output. Plots show the mean, max, and min across the 50 single-system fits.}
    \label{fig:single-system-diagnostics}
\end{figure}

Figure~\ref{fig:single-system-diagnostics} reports relevant metrics aggregated over all 50 single-system fits.
The hard-argmin--target MAE and differentiable-output--hard-argmin discrepancy make the different optimization behavior visible.
For DISCOMAX~H, the differentiable output is the soft argmin; for Surrogate~G, it is the output of the learned surrogate solver.

\begin{figure}[htbp]
    \centering
    \includegraphics[width=1.\textwidth]{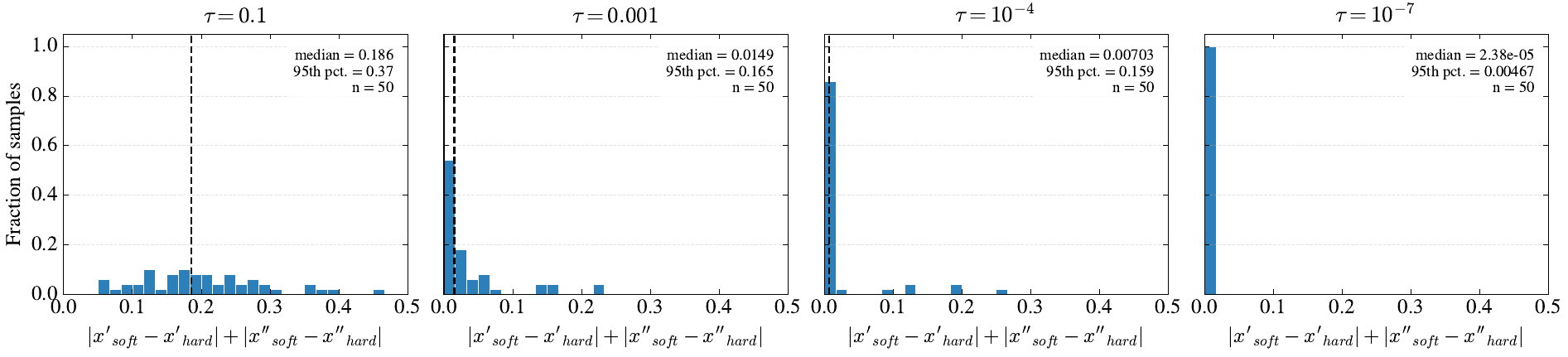}
    \caption{Final DISCOMAX~H soft--hard composition-discrepancy distributions for all 50 single-system fits at fixed softmax temperatures. The dashed line marks the median in each panel; annotations report the median and 95th percentile. All panels use common bins and horizontal limits.}
    \label{fig:soft-hard-discrepancy-single-system}
\end{figure}

\begin{figure}[htbp]
    \centering
    \includegraphics[width=1.\textwidth]{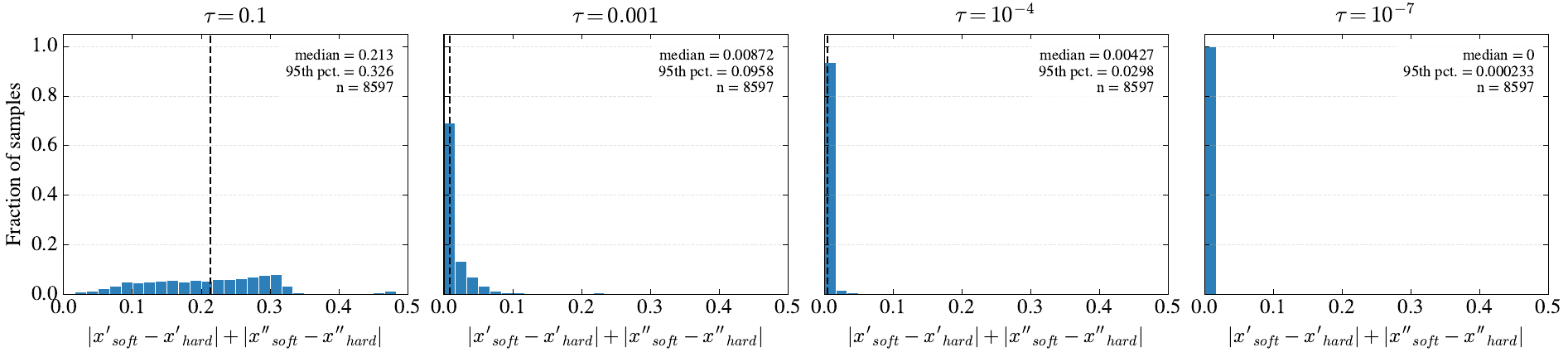}
    \caption{Final DISCOMAX~H soft--hard composition-discrepancy distributions for all 8597 held-out samples pooled across ten cross-validation folds. Each sample is evaluated by the model for the fold in which it was held out. The dashed line marks the median in each panel; annotations report the median and 95th percentile. All panels use common bins and horizontal limits.}
    \label{fig:soft-hard-discrepancy-generalization}
\end{figure}

To assess the final soft--hard displacement over complete experiment distributions, Figures~\ref{fig:soft-hard-discrepancy-single-system} and~\ref{fig:soft-hard-discrepancy-generalization} recompute the DISCOMAX~H soft argmin on each final learned \gls{gmix} landscape at four fixed temperatures.
The discrepancy is $|x'_{\mathrm{soft}}-x'_{\mathrm{hard}}|+|x''_{\mathrm{soft}}-x''_{\mathrm{hard}}|$.
The single-system row contains all 50 final fits, while the generalization row contains all 8597 samples pooled from the ten held-out cross-validation folds; each dataset sample occurs in exactly one held-out fold.
As expected, decreasing $\tau$ concentrates the soft argmin around the hard minimizer.
At $\tau=10^{-7}$, the median discrepancy is $2.38\times10^{-5}$ for the single-system fits and numerically zero for held-out cross-validation, with 95th percentiles of $4.67\times10^{-3}$ and $2.33\times10^{-4}$, respectively.

\begin{figure}[htbp]
    \centering
    \includegraphics[width=1.\textwidth]{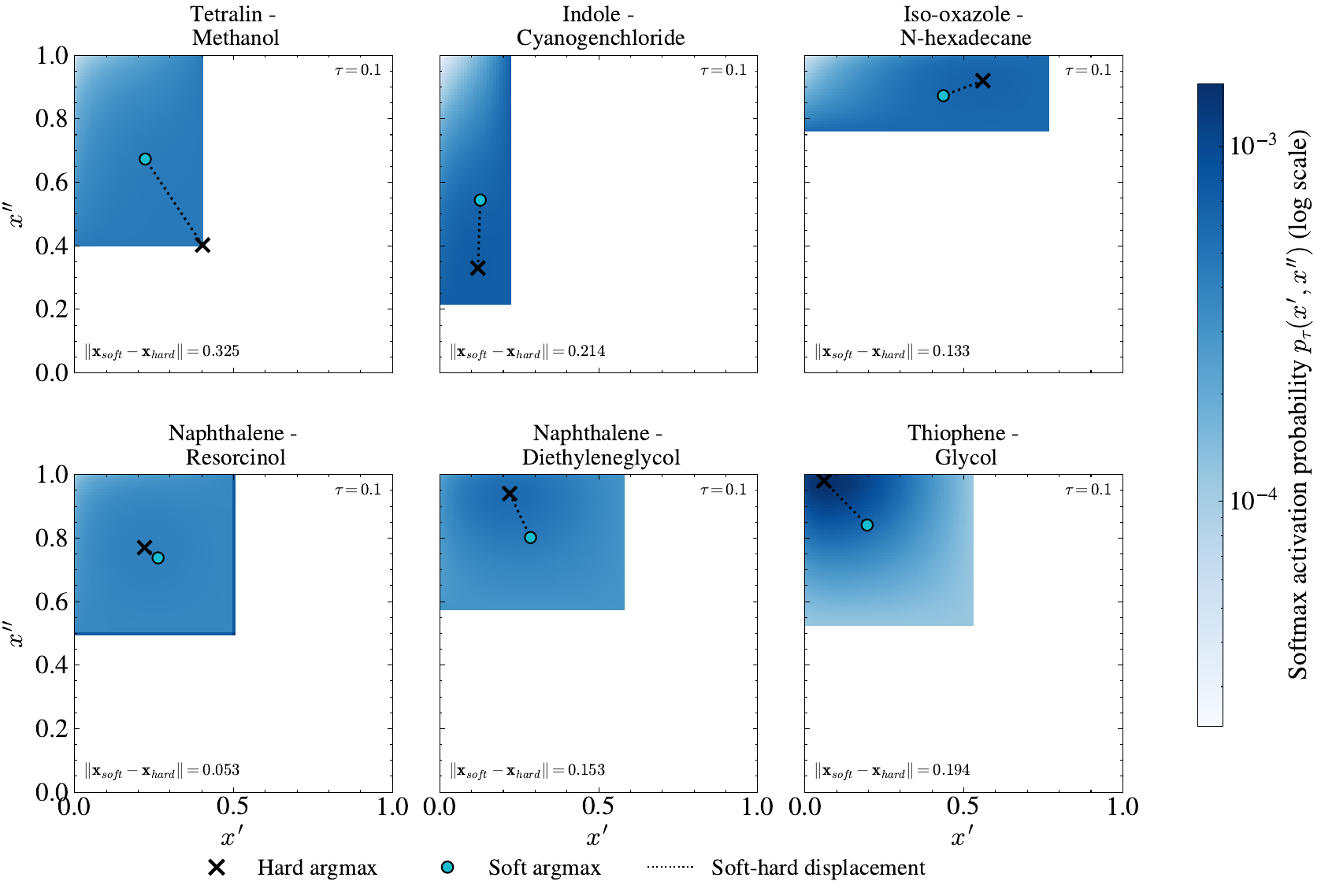}
    \caption{Final DISCOMAX~H softmax activation landscapes for six representative single-system fits at fixed $\tau=0.1$. Crosses and circles denote the hard and soft argmax, respectively, and dotted lines show their displacement. Activation probabilities share a logarithmic color scale across all six systems.}
    \label{fig:softmax-activation-tau-1e-1}
\end{figure}

\begin{figure}[htbp]
    \centering
    \includegraphics[width=1.\textwidth]{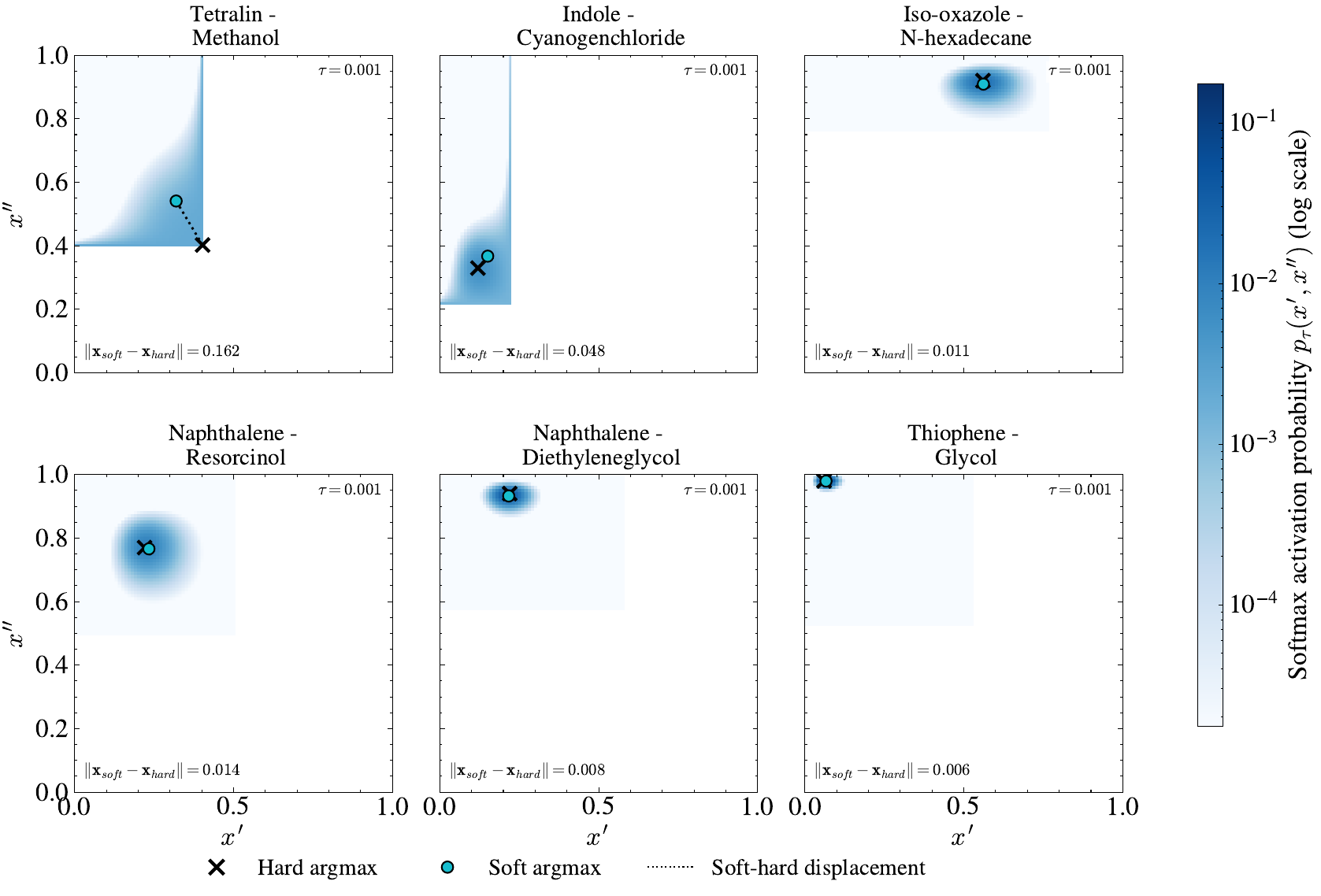}
    \caption{Final DISCOMAX~H softmax activation landscapes for the same six single-system fits at fixed $\tau=10^{-3}$. The activation distributions become more concentrated around the hard argmax than at $\tau=0.1$.}
    \label{fig:softmax-activation-tau-1e-3}
\end{figure}

\begin{figure}[htbp]
    \centering
    \includegraphics[width=1.\textwidth]{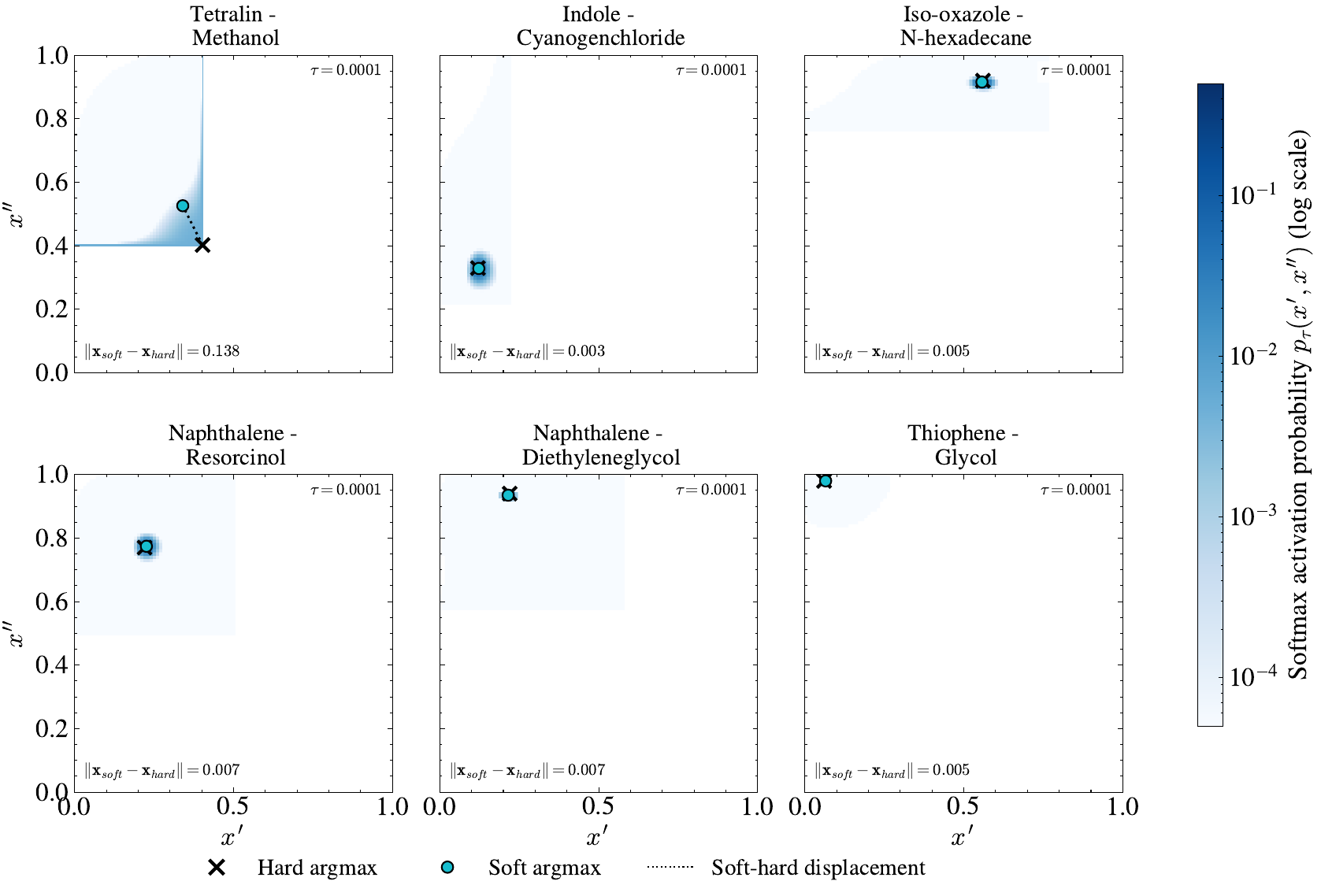}
    \caption{Final DISCOMAX~H softmax activation landscapes for the same six single-system fits at fixed $\tau=10^{-4}$.}
    \label{fig:softmax-activation-tau-1e-4}
\end{figure}

\begin{figure}[htbp]
    \centering
    \includegraphics[width=1.\textwidth]{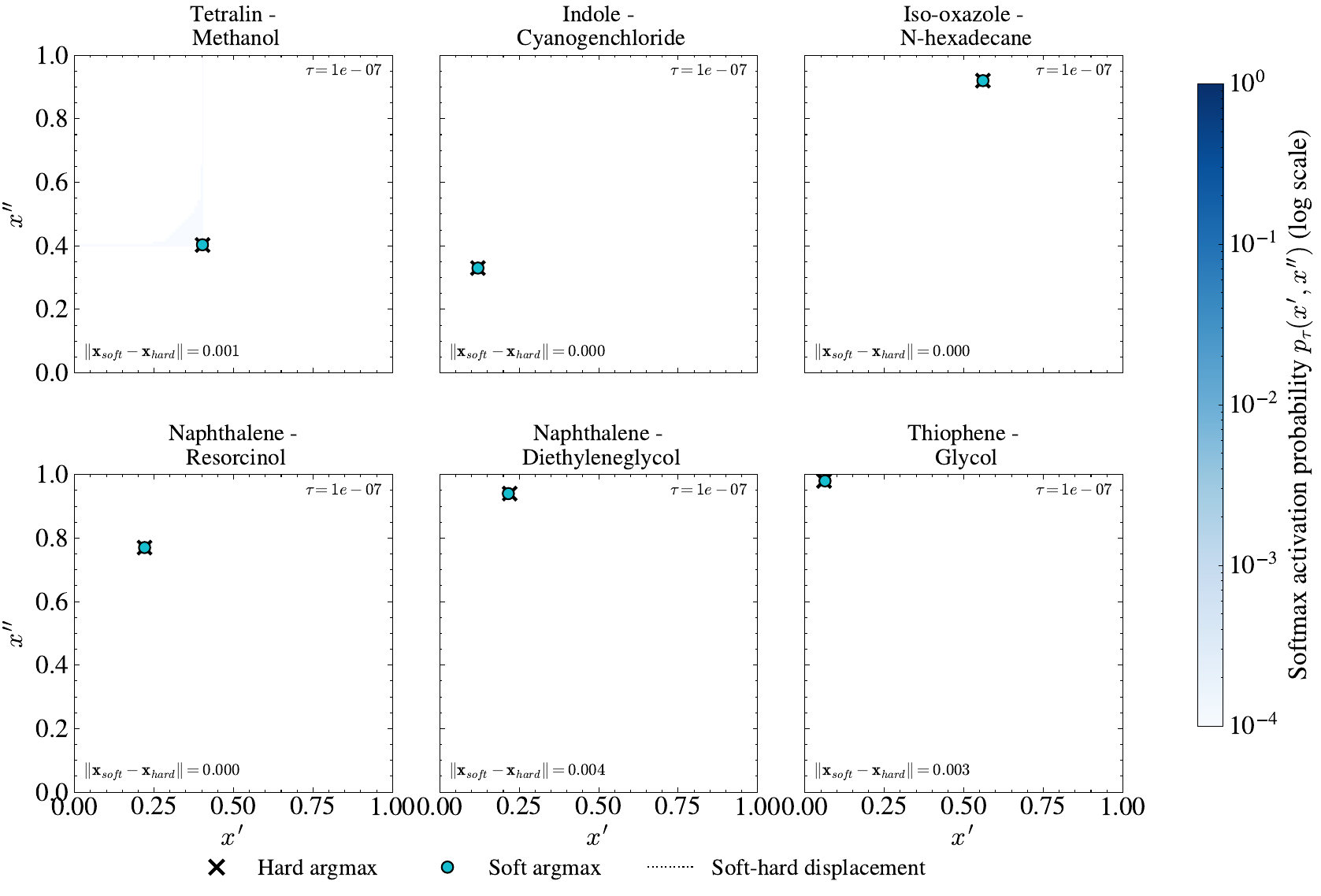}
    \caption{Final DISCOMAX~H softmax activation landscapes for the same six single-system fits at fixed $\tau=10^{-7}$, where the soft and hard argmax nearly coincide.}
    \label{fig:softmax-activation-tau-1e-7}
\end{figure}

Figures~\ref{fig:softmax-activation-tau-1e-1}, \ref{fig:softmax-activation-tau-1e-3}, \ref{fig:softmax-activation-tau-1e-4}, and \ref{fig:softmax-activation-tau-1e-7} visualize this progressive concentration of the final DISCOMAX~H softmax activation landscapes for the same six representative single-system fits with increasing gap width as $\tau$ decreases.
We can observe nicely that for fixed $\tau$, the systems with smaller gap-width, which correspond to flatter minima, e.g., the first and second system, exhibit a larger soft--hard displacement than the systems with sharper minima.
The reason for that is that the equilibrium compositions are very close to the feed composition, which makes the \gls{gmix} landscape very flat in the $x'$ and $x''$ directions, this then moves the softargmax into the center of the feasible composition space.

Table~\ref{tab:hpo-grid} summarizes the auxiliary hyperparameter scan used for the single-system fitting experiments. DISCOMAX is relatively robust to different hyperparameter choices.

\input{tables/hpo_grid_table.tex}

\begin{figure}[htbp]
    \centering
    \includegraphics[width=1.\textwidth]{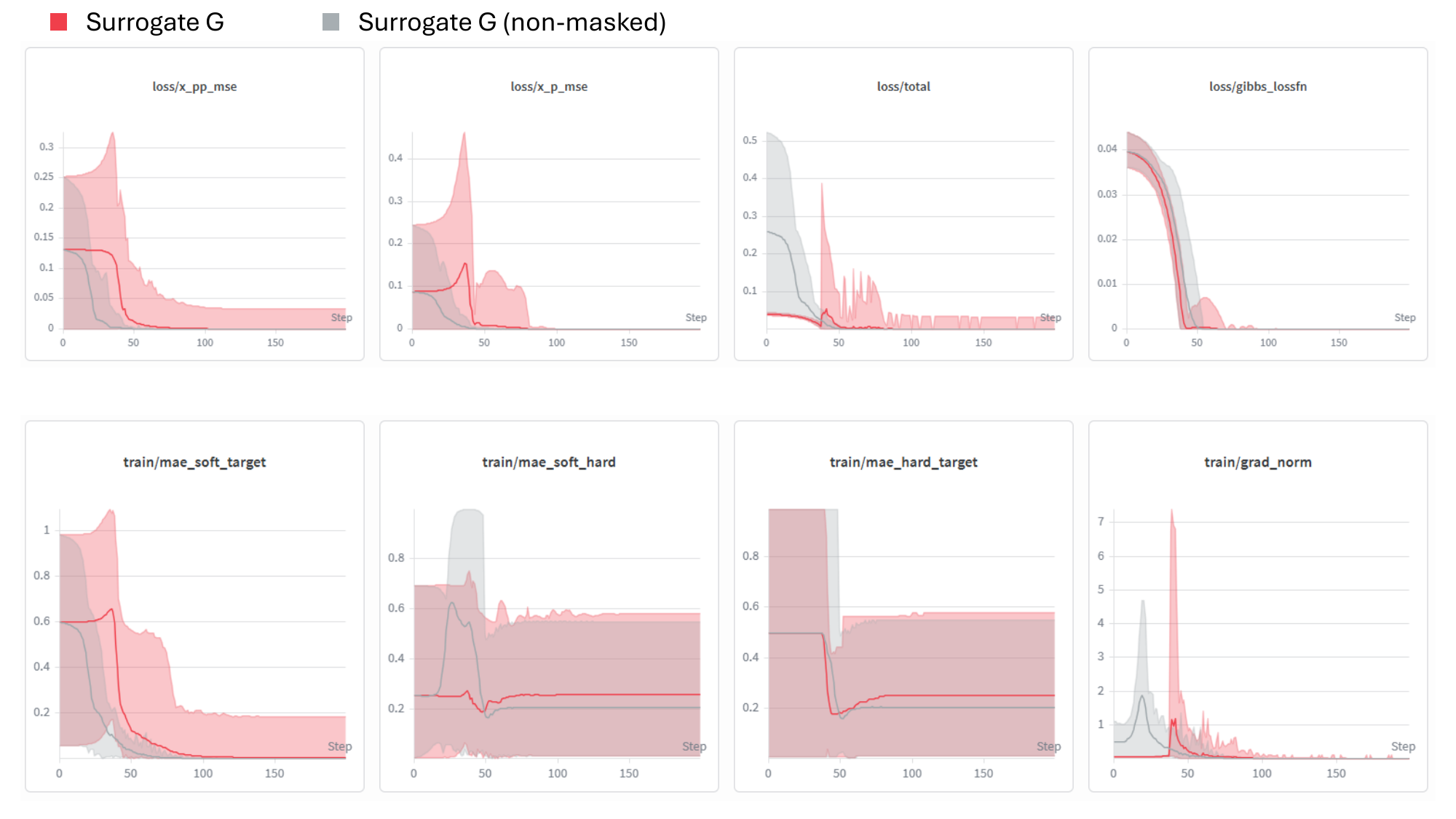}
    \caption{Single-system training diagnostics for Surrogate~G in red and Surrogate~G (non-masked) in grey, including surrogate-output--target MSE terms ($x',x''$), Gibbs loss, hard-target MAE, surrogate-output--hard-argmin discrepancy, and gradient norm of the model weights. The surrogate output does not involve a softmax. Plots show the mean, max, and min across the 50 single-system fits.}
    \label{fig:single-system-diagnostics-nonmasked}
\end{figure}

A comparison of the training diagnostics for the non-masked Surrogate~G baseline and the masked Surrogate~G version (used in all core experiments) is shown in Figure~\ref{fig:single-system-diagnostics-nonmasked}.
The non-masked version of Surrogate~G applies the surrogate solver prediction loss to all systems regardless of the presence of a phase split.
Surprisingly, this version does not perform significantly different to the masked version, and even slightly better than the hierarchical masked counterpart. This may be due to unstable gradients in the hierarchical version, as visible in the higher gradient norm spike, once the necessary condition for the surrogate loss to be active is met.

\begin{figure}[htbp]
    \centering
    \includegraphics[width=1.\textwidth]{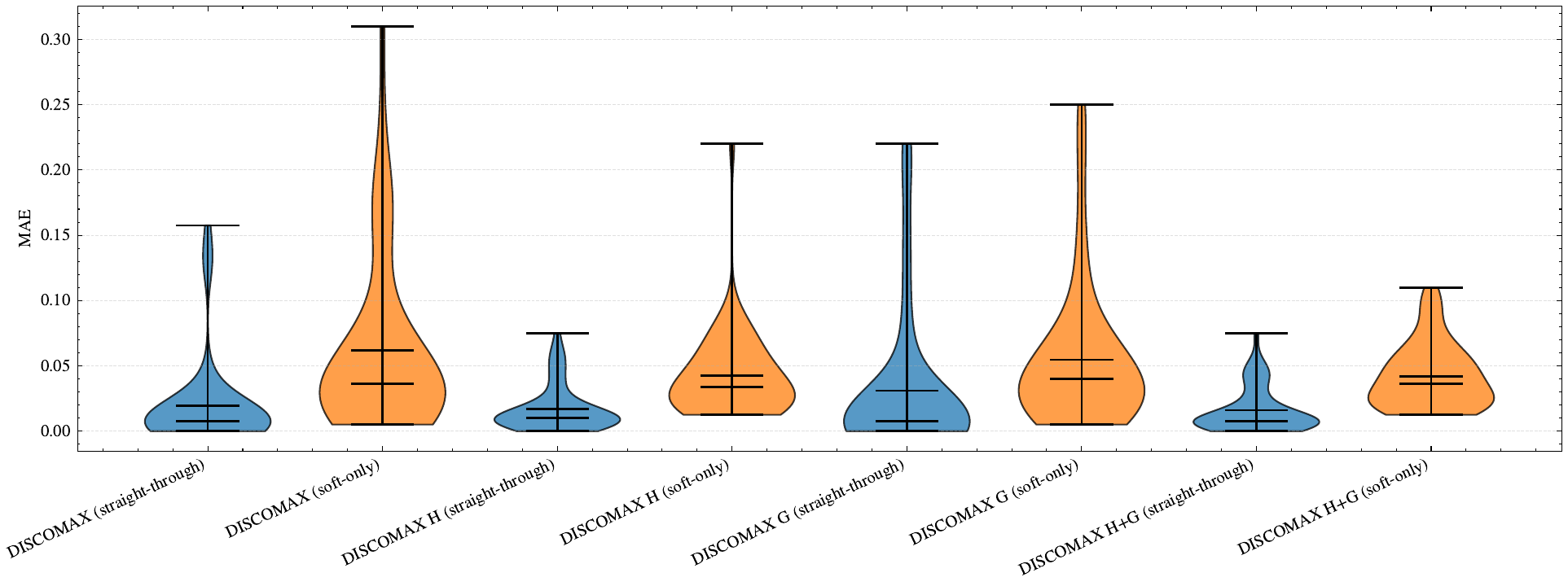}
    \caption{Single-system phase-composition MAE distributions across all 50 systems for DISCOMAX variants trained with the straight-through estimator and directly on the soft phase-composition outputs.}
    \label{fig:single-system-soft-only}
\end{figure}

Figure~\ref{fig:single-system-soft-only} shows that removing the straight-through estimator consistently increases the phase-composition MAE across all four DISCOMAX loss configurations in the 50 single-system fits.

\subsection{Full-dataset training diagnostics}
\begin{figure}[htbp]
    \centering
    \includegraphics[width=0.75\linewidth]{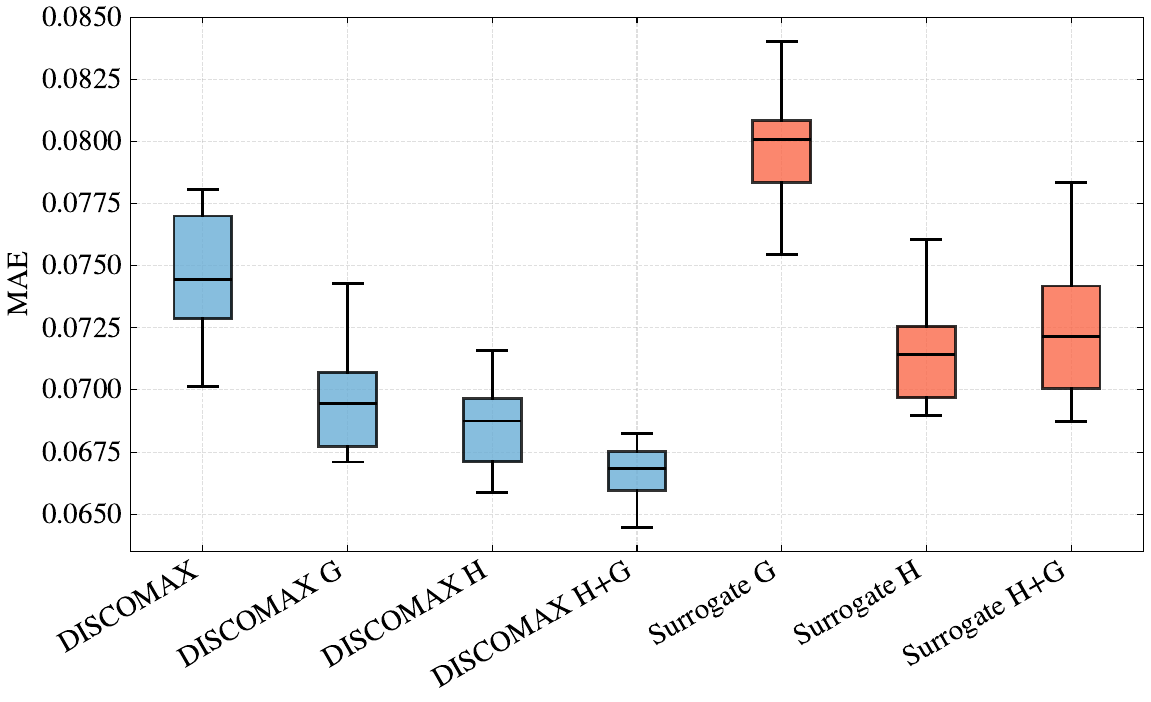}
    \caption{Cross-validation results of the test-set of ten folds for all full-dataset configurations (pure Surrogate not shown).}
    \label{fig:boxplot_cv}
\end{figure}

Figure~\ref{fig:boxplot_cv} summarizes the ten-fold cross-validation distribution of the test-set phase-composition MAE for all full-dataset configurations.
The boxes show the interquartile range across folds, and the center line shows the median.

Table~\ref{tab:cv-all-methods-gap-width} reports the corresponding cross-validation performance for predicting the miscibility-gap width across all model variants and data splits.

The miscibility-gap width error is defined as $(x''^\ast - x'^\ast) - (x'' - x')$, where $x'^\ast$ and $x''^\ast$ are the target phase compositions, and $x'$ and $x''$ are the predicted phase compositions.

\input{tables/cv_metrics_table_all_methods_from_preds_gap-width.tex}

\begin{figure}[htbp]
    \centering
    \includegraphics[width=1.\textwidth]{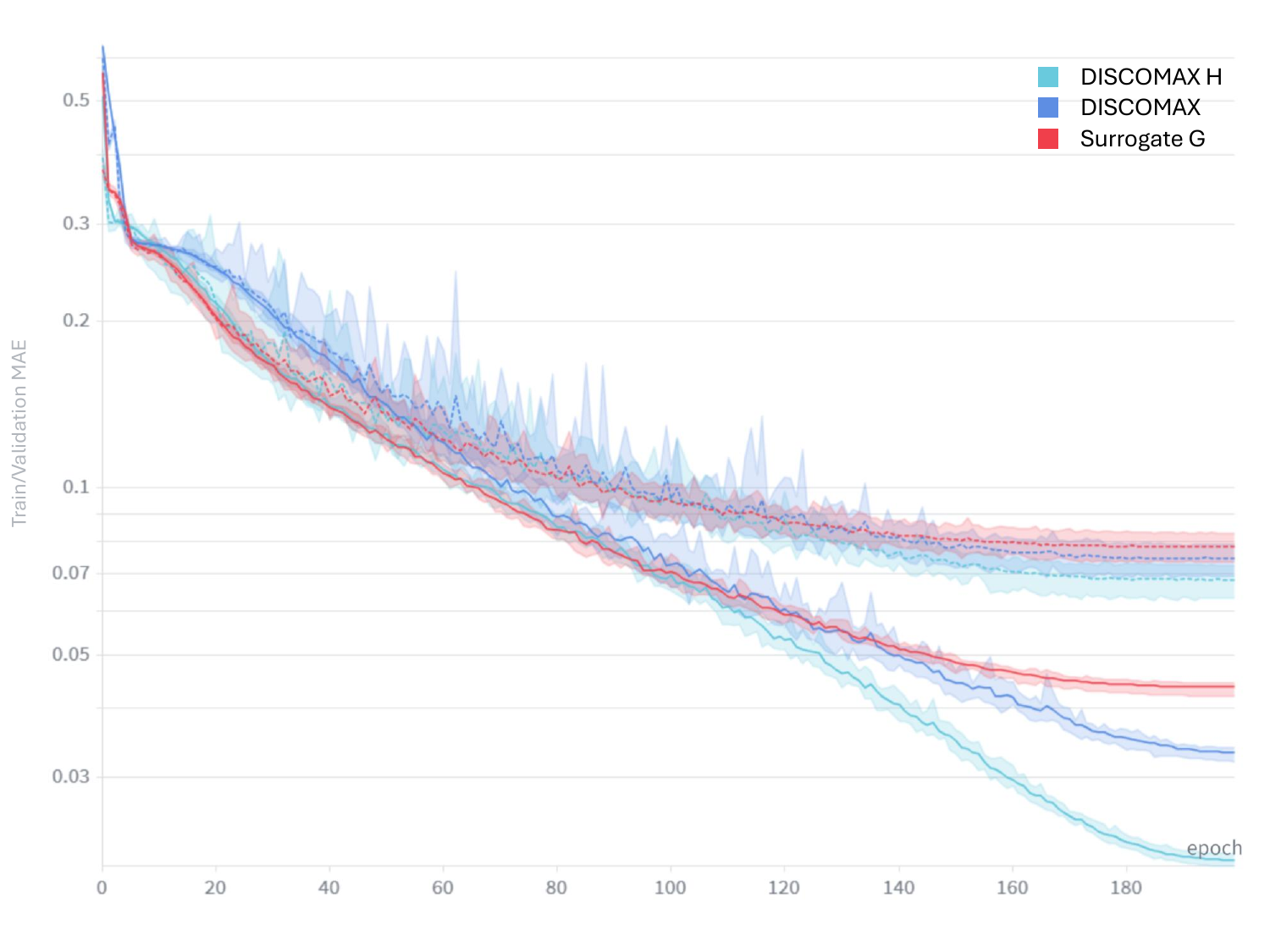}
    \caption{Training and validation MAE curves for DISCOMAX~H (cyan), DISCOMAX (blue), and Surrogate~G (red) across the cross-validation training runs.}
    \label{fig:cv-training-curves}
\end{figure}

\begin{figure}[htbp]
    \centering
    \includegraphics[width=1.\textwidth]{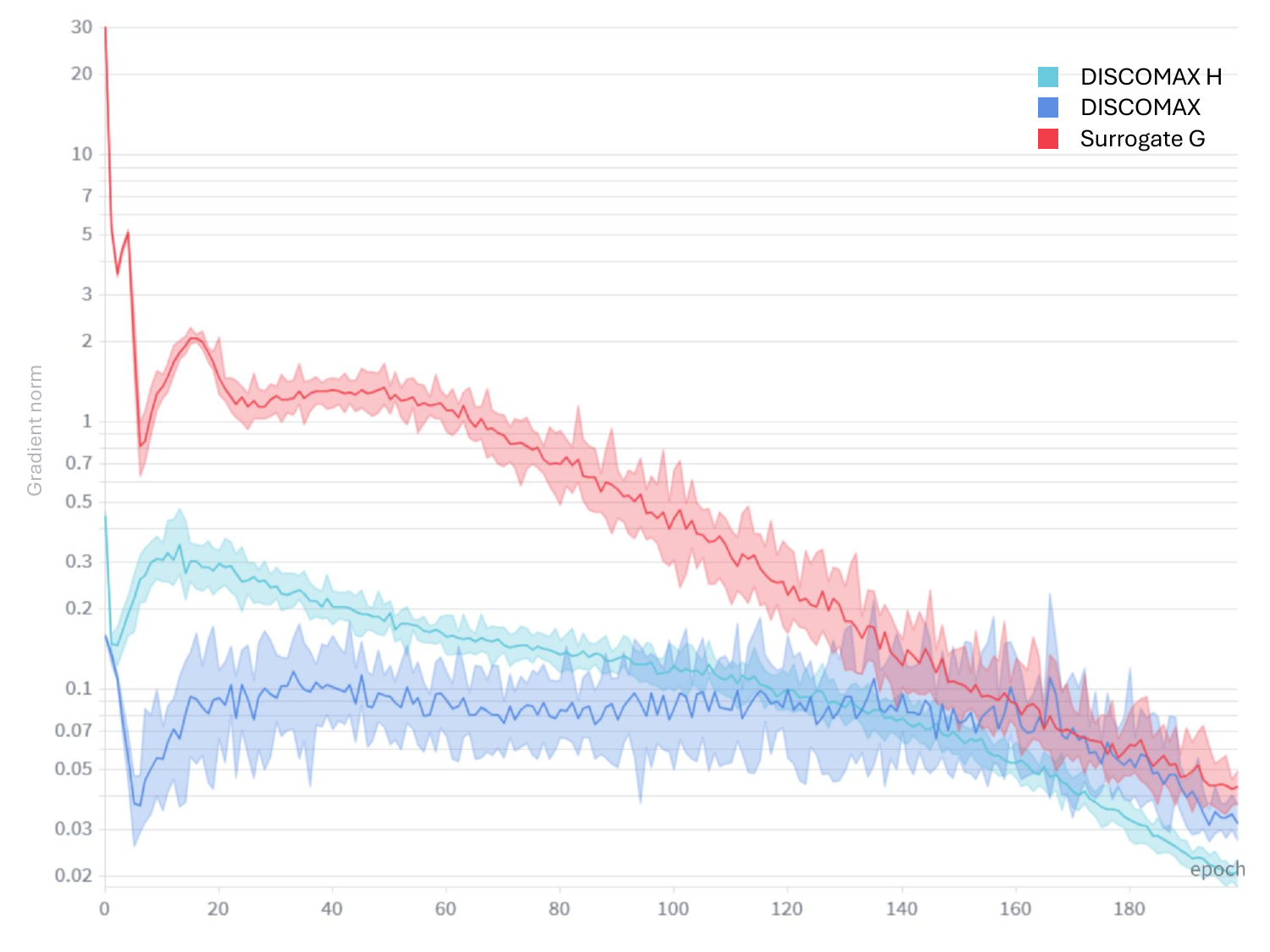}
    \caption{Gradient-norm traces for DISCOMAX~H (cyan), DISCOMAX (blue), and Surrogate~G (red) during cross-validation training.}
    \label{fig:cv-gradient-norms}
\end{figure}

Figures~\ref{fig:cv-training-curves} and~\ref{fig:cv-gradient-norms} show the corresponding training dynamics for the cross-validation runs.
The DISCOMAX-based variants reach lower training and validation MAE than Surrogate~G, and also the gradient norms are more stable initially and lower in variance.

\begin{figure}[htbp]
    \centering
    \includegraphics[width=.8\textwidth]{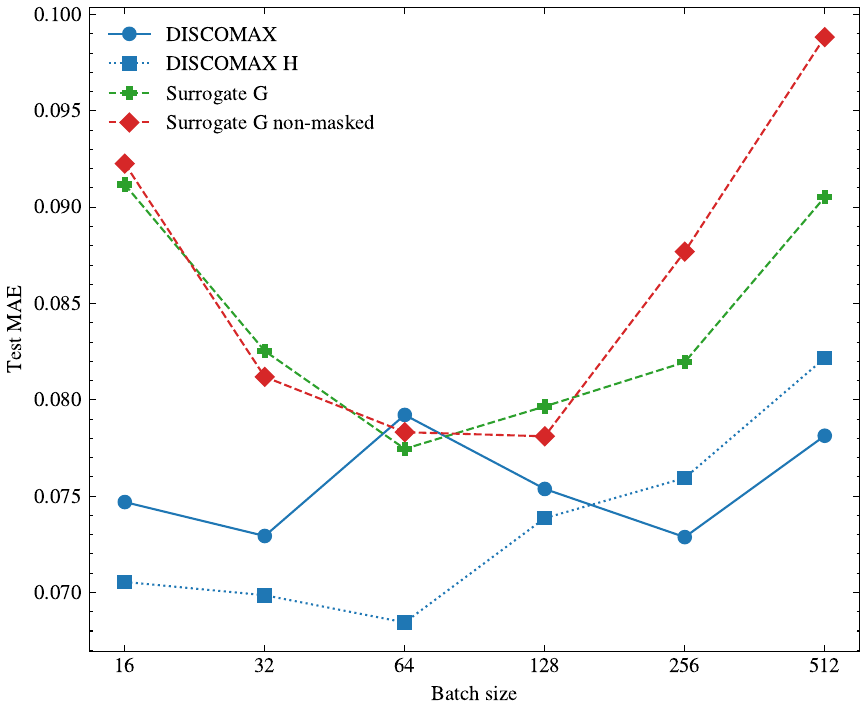}
    \caption{Test-set phase-composition MAE as a function of batch size for the compact comparison between DISCOMAX, DISCOMAX~H, Surrogate~G, and the non-masked Surrogate~G baseline on a single fold.}
    \label{fig:batch_comp_vs_test_mae_hard_target}
\end{figure}

The influence of the batch size on test-set performance is illustrated in Figure~\ref{fig:batch_comp_vs_test_mae_hard_target}.
The most important comparison of the work is between the proposed DISCOMAX and DISCOMAX~H variant and the baseline Surrogate~G model.
The corrected masked Surrogate~G baseline, used everywhere throughout this work, does not perform significantly different to a non-masked Surrogate~G version, which does not mask the surrogate loss for single-phase systems, and always applies the surrogate loss to all systems, regardless of the presence of a phase split.
A degradation in performance is observed for all models at larger batch sizes, at fixed learning rate.
Both proposed methods are more robust than the surrogate baseline with respect to changes in batch size.

\begin{figure}[htbp]
    \centering
    \includegraphics[width=1.\textwidth]{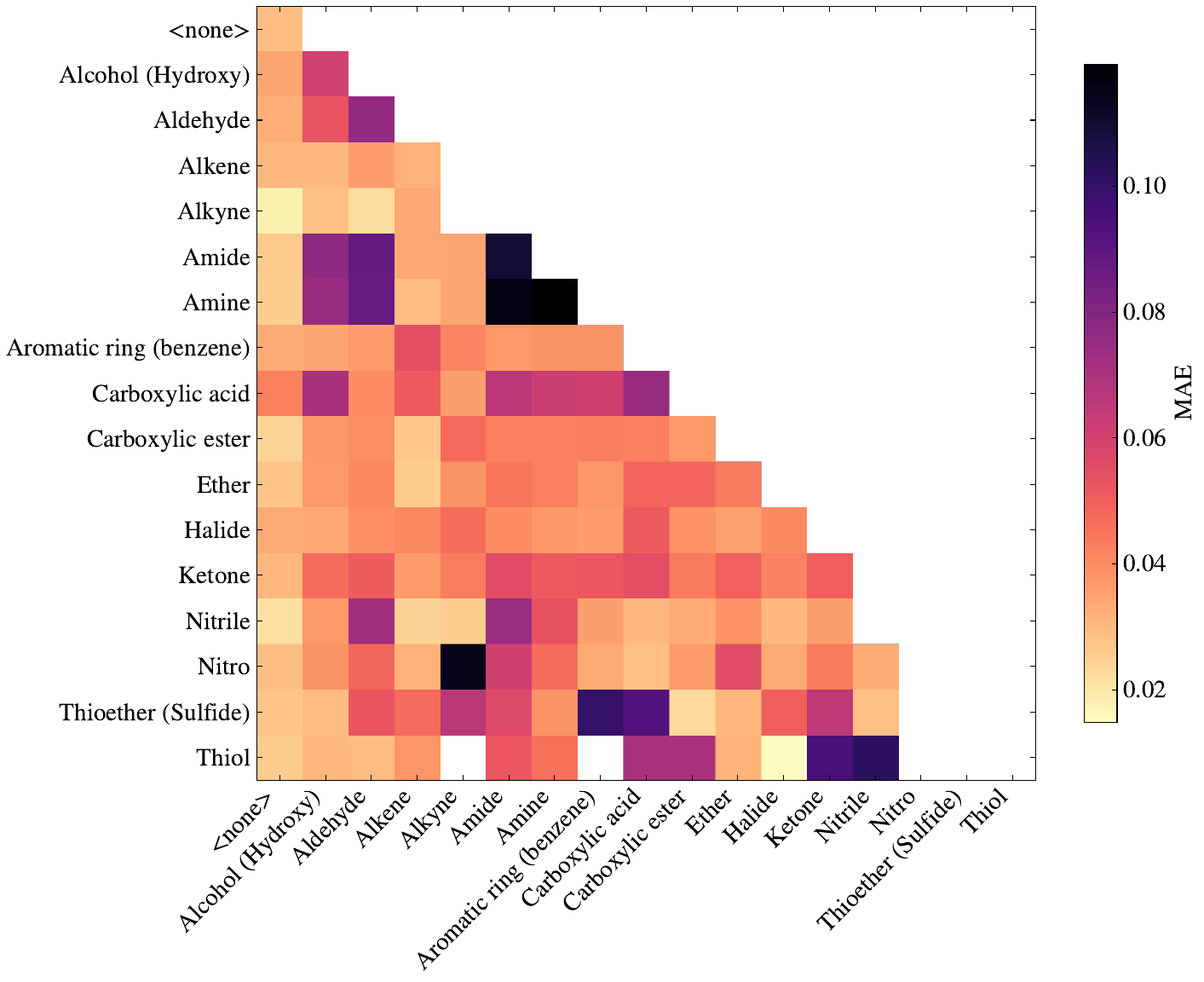}
    \caption{Cross-validation test-set MAE of phase-composition predictions grouped by the presence of common functional groups within the binary mixtures.
    Error patterns reveal strong chemical dependencies. Functional-group frequencies in the train and test sets are reported in Table~\ref{tab:fgs}.}
    \label{fig:lle_fg_mae}
\end{figure}

To investigate potential chemical sources of error, Figure~\ref{fig:lle_fg_mae} groups the prediction MAE by the presence of common functional groups in the mixtures.
Table~\ref{tab:fgs} lists the frequencies of these groups in the dataset.
The model performs well across most chemical classes, but exhibits degraded accuracy for mixtures involving carboxylic acids and, to a lesser extent, nitro-containing compounds.
These trends likely reflect intrinsic difficulties of modeling highly nonideal systems rather than artifacts of the solver itself.

\begin{figure}[htbp]
    \centering
    \includegraphics[width=.8\textwidth]{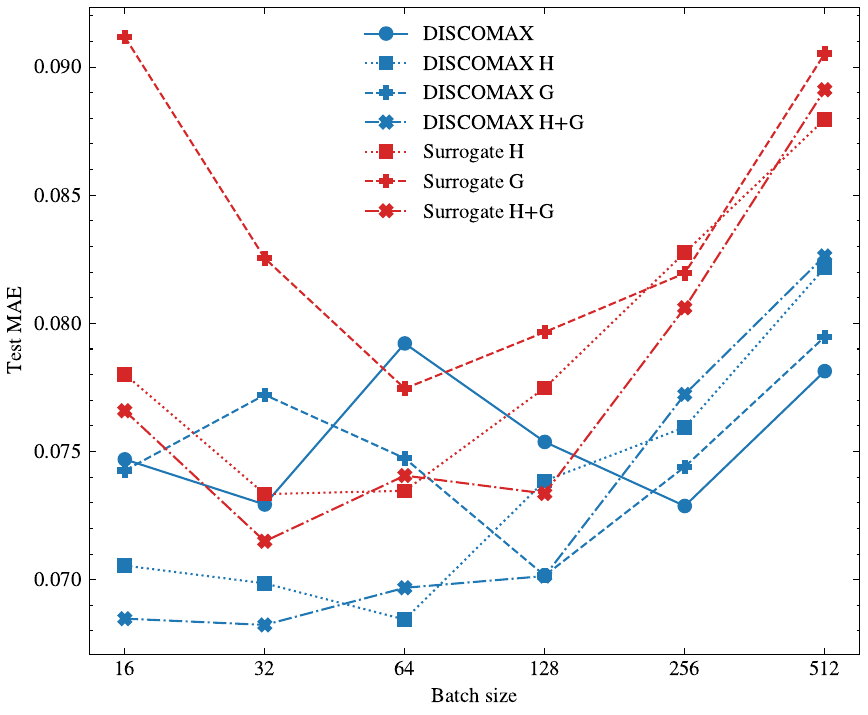}
    \caption{Test-set MAE phase-composition predictions as a function of batch size for different solver and loss configurations for a single fold.
    DISCOMAX~H and DISCOMAX~H+G provide the strongest overall performance.
    Smaller batch sizes yield the best overall performance, whereas large batch sizes substantially degrade accuracy, highlighting a strong interaction between mini-batch statistics and equilibrium-model training dynamics.}
    \label{fig:batch_size_vs_test_mae_hard_target}
\end{figure}

Figure~\ref{fig:batch_size_vs_test_mae_hard_target} depicts the test set MAE of several solver-loss configurations across a range of batch sizes.
All models suffer from higher batch sizes for the chosen learning rate of 0.001.
The training was not stable for higher learning rates for some of the surrogate variants.
The DISCOMAX~H variant seems most robust to varying batch sizes.

\begin{figure}[htbp]
    \centering
    \includegraphics[width=.8\textwidth]{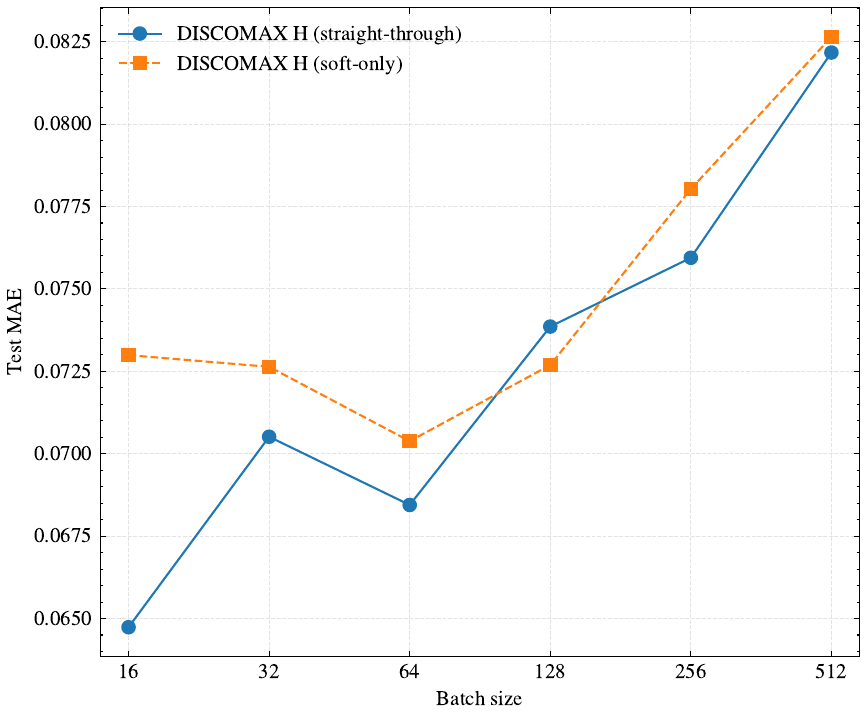}
    \caption{Test-set phase-composition MAE as a function of batch size for DISCOMAX~H trained with the straight-through estimator and directly on the soft phase-composition outputs on a single fold.}
    \label{fig:batch-size-soft-only}
\end{figure}

Figure~\ref{fig:batch-size-soft-only} shows that removing the straight-through estimator slightly degrades test performance, particularly at small batch sizes.

\begin{figure}[p]
    \centering
    \includegraphics[width=0.95\textwidth]{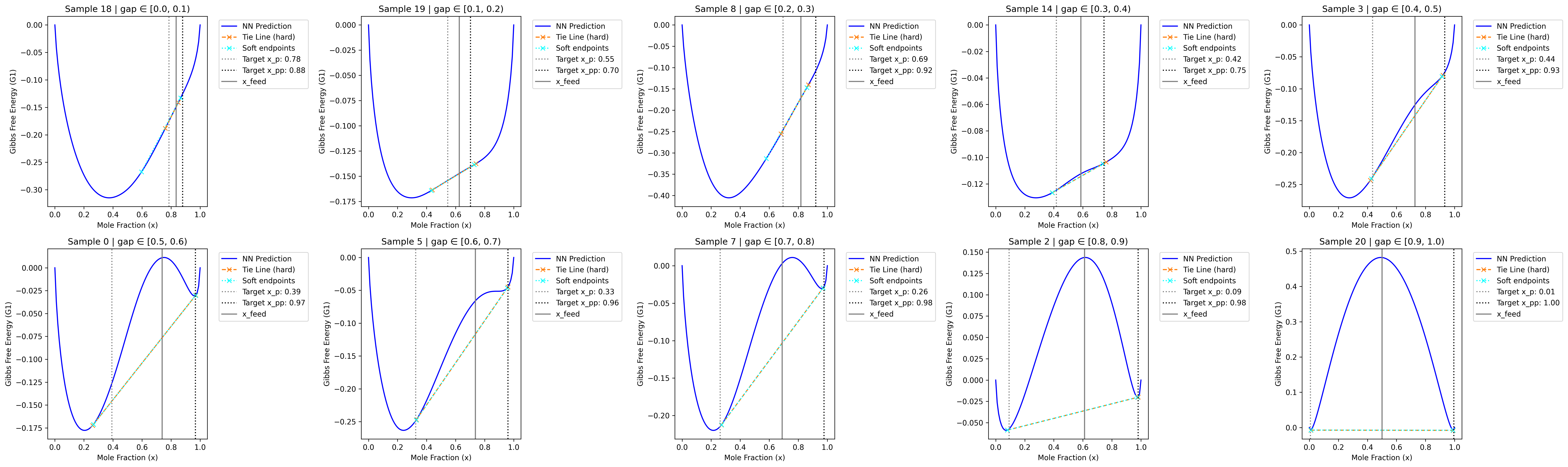}
    \vfill
    \vspace{.5cm}
    \includegraphics[width=0.95\textwidth]{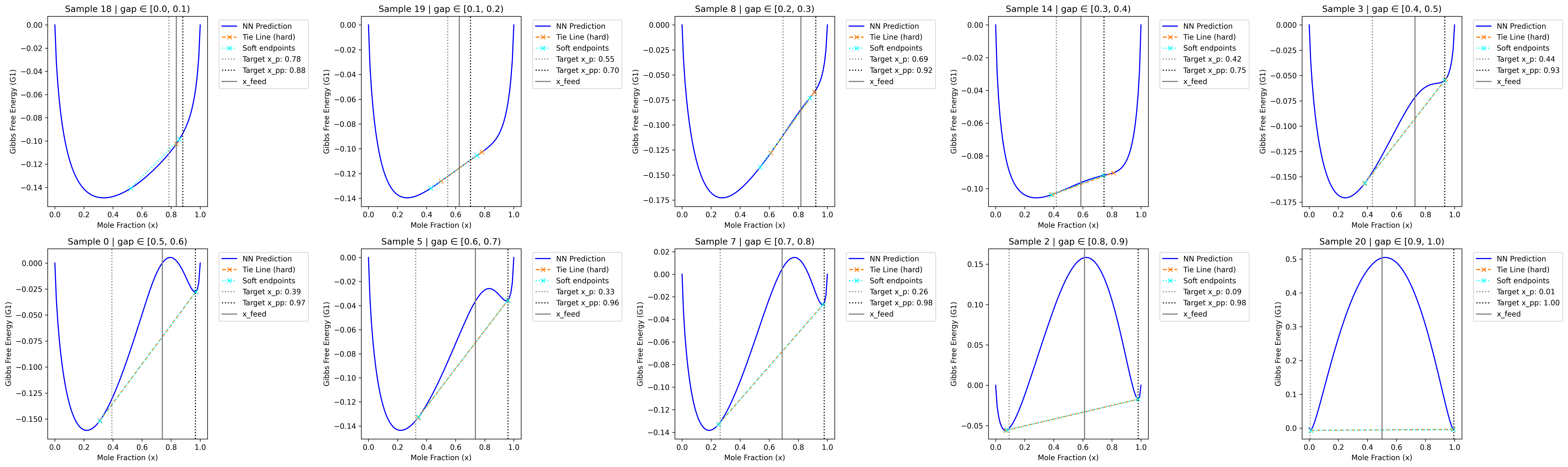}
    \vfill
    \vspace{.5cm}
    \includegraphics[width=0.95\textwidth]{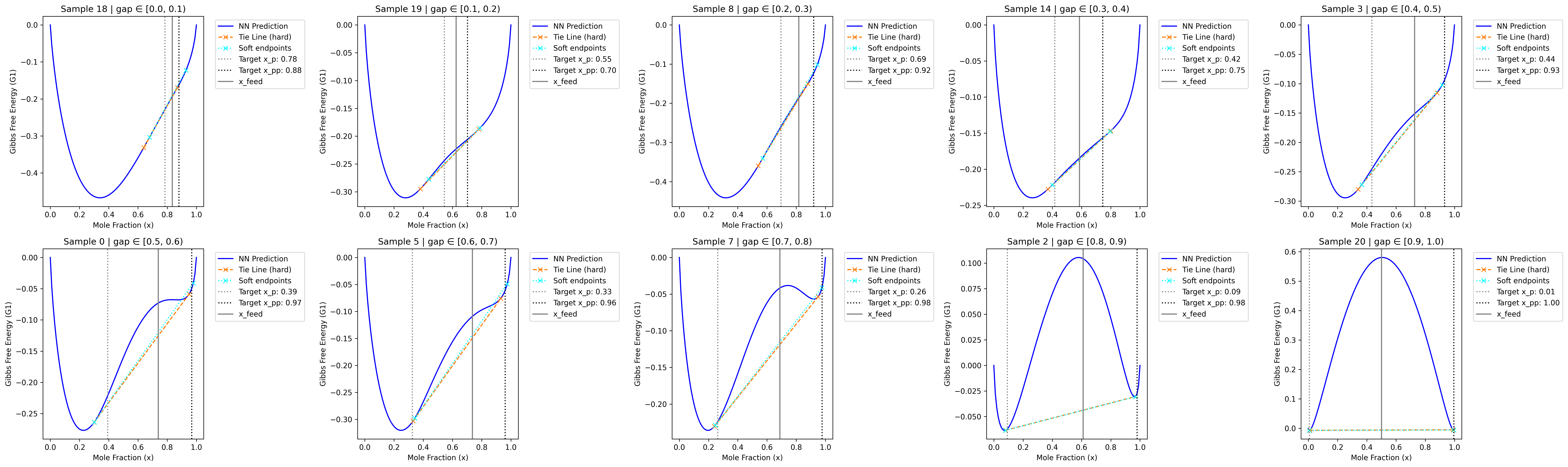}
    \caption{Representative test-set \gls{gmix} profiles and phase-composition predictions for DISCOMAX~H (top), DISCOMAX (middle), and Surrogate~G (bottom). Examples are ordered by miscibility-gap width. Solid curves denote the learned \gls{gmix} profile; vertical markers indicate the feed composition, target phase compositions, and both the hard and soft predictions.}
    \label{fig:phase-profile-test}
\end{figure}

\begin{figure}[p]
    \centering
    \includegraphics[width=0.95\textwidth]{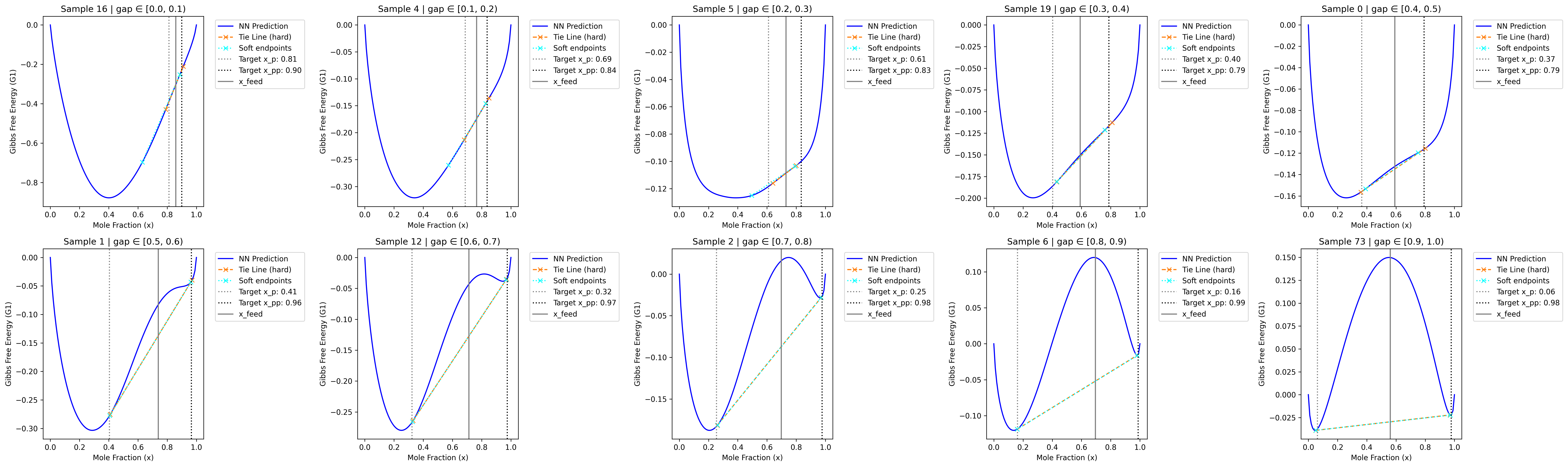}
    \vfill
    \vspace{.5cm}
    \includegraphics[width=0.95\textwidth]{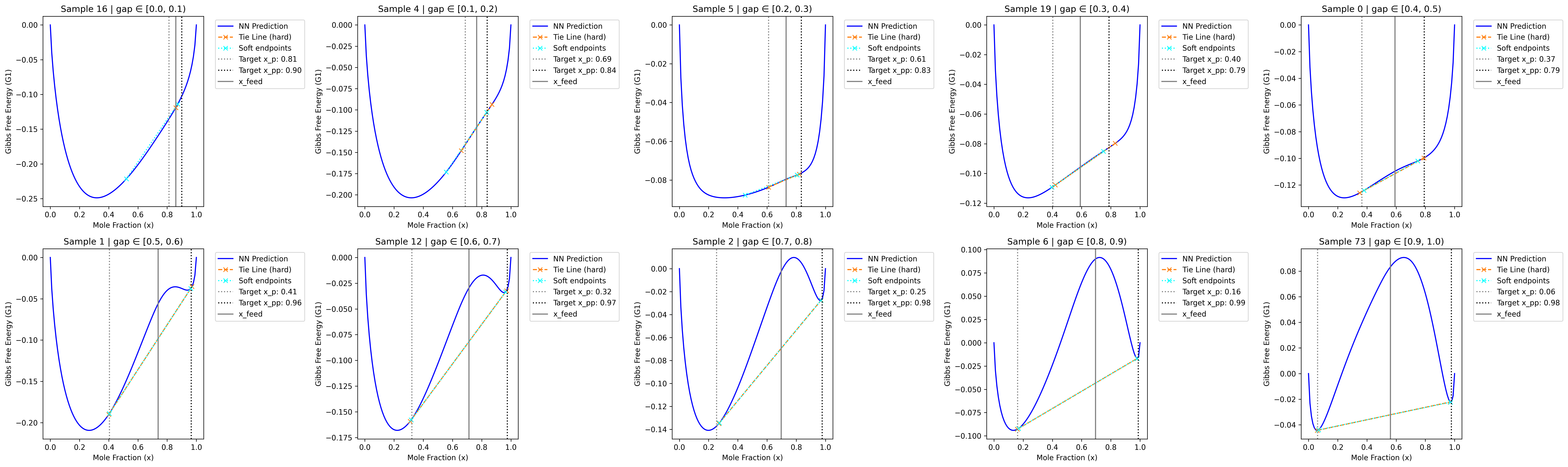}
    \vfill
    \vspace{.5cm}
    \includegraphics[width=0.95\textwidth]{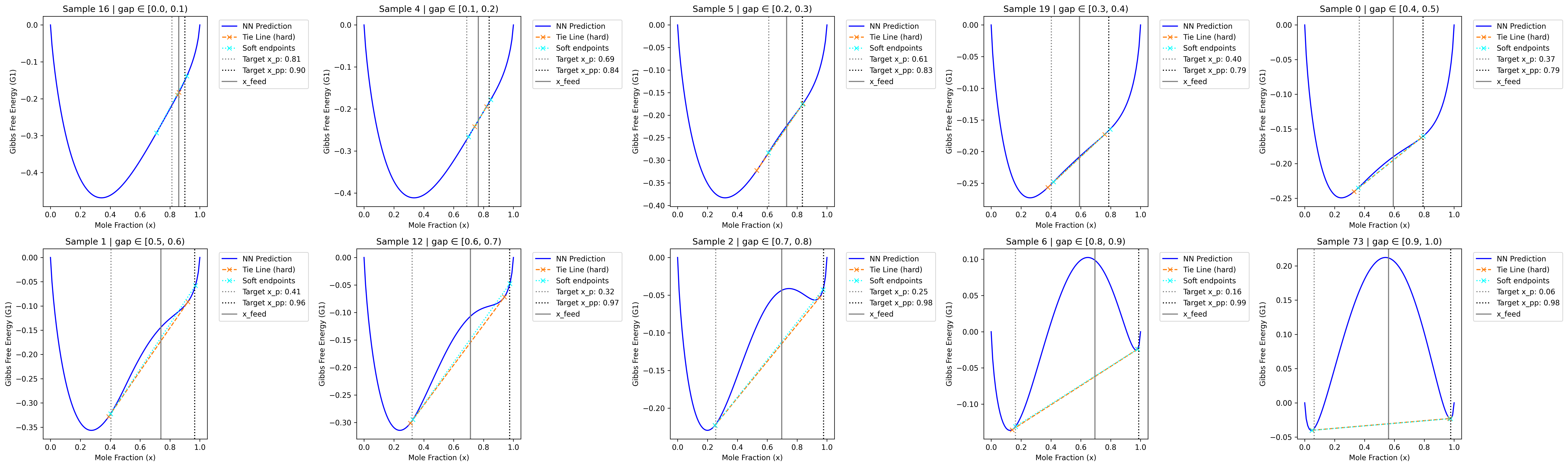}
    \caption{Representative training-set \gls{gmix} profiles and phase-composition predictions for DISCOMAX~H (top), DISCOMAX (middle), and Surrogate~G (bottom), using the same visual encoding as Figure~\ref{fig:phase-profile-test}.}
    \label{fig:phase-profile-train}
\end{figure}

Figures~\ref{fig:phase-profile-test} and~\ref{fig:phase-profile-train} show representative learned \gls{gmix} profiles for held-out test and training systems, respectively. The learned \gls{gmix} profiles are smooth, physically reasonable, and closely match the target equilibrium phase compositions, for both the training and test set.

\begin{table}[bpht]
\centering
\caption{Full cross-validation results for the test, validation, and train set aggregated over ten folds for the best performing models on the COSMO-RS dataset DISCOMAX batch-size 128 and DISCOMAX H batch-size 128. Metrics are shown as mean $\pm$ standard deviation.}
\label{tab:cosmo-cv}
\begin{tabular}{lccc}
\toprule
 & Test MAE & Test RMSE & Test $R^2$ \\
Model &  &  &  \\
\midrule
DISCOMAX H & 0.046 $\pm$ 0.003 & 0.047 $\pm$ 0.003 & 0.986 $\pm$ 0.002 \\
DISCOMAX & 0.045 $\pm$ 0.003 & 0.047 $\pm$ 0.003 & 0.986 $\pm$ 0.002 \\
\bottomrule
\end{tabular}
\end{table}

Table~\ref{tab:cosmo-cv} reports the full ten-fold COSMO-RS cross-validation 
results for the best-performing configurations of DISCOMAX~H and DISCOMAX.

\input{tables/functional_group_counts_table.tex}

\begin{figure}[htbp]
    \centering
    \includegraphics[width=0.8\linewidth]{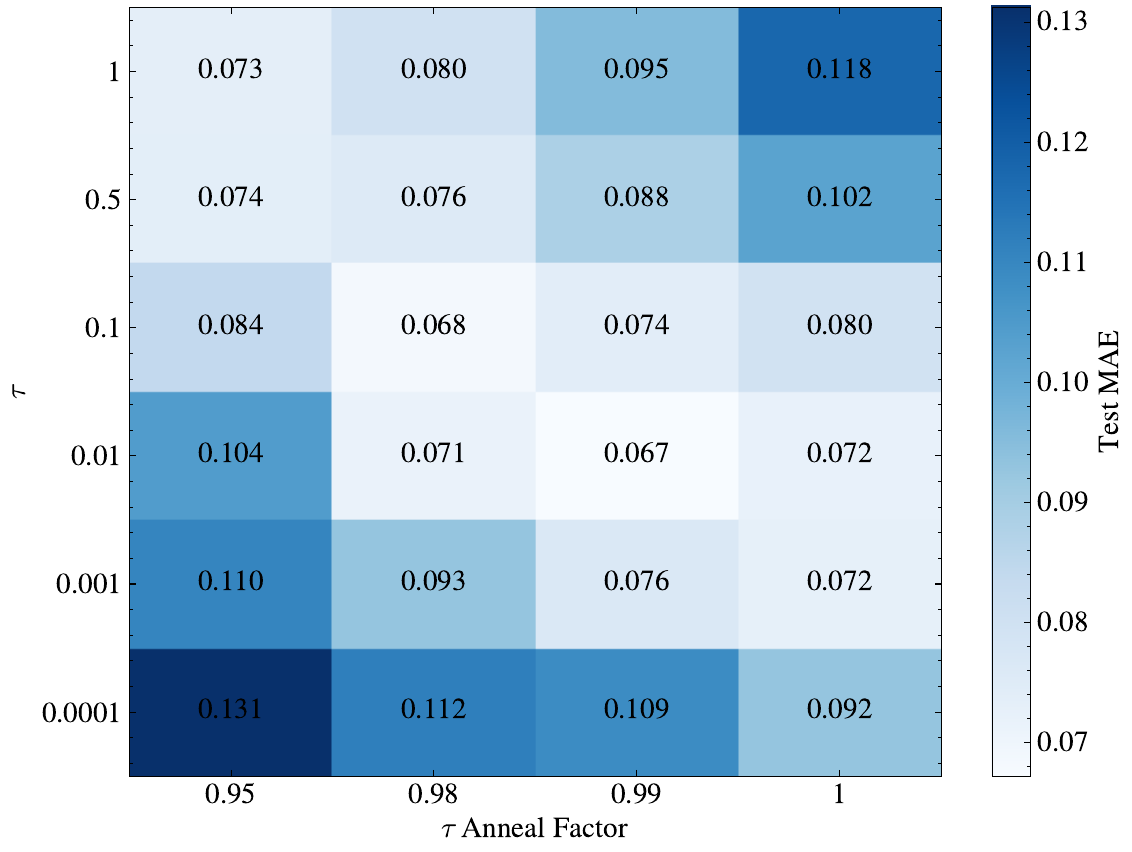}
    \caption{Influence of the initial temperature parameter $\tau$ and multiplicative annealing factor on the single-fold test-set MAE for DISCOMAX~H. Moderate smoothing and gradual annealing give the best results; the lowest observed MAE is 0.067 at $\tau=0.01$ with annealing factor 0.99.}
    \label{fig:dk_schedule}
\end{figure}

Figure~\ref{fig:dk_schedule} shows that the equilibrium-layer temperature must be chosen in an intermediate range.
Very low initial temperatures or overly strong annealing degrade the final test MAE, while $\tau=0.1$ with annealing factor 0.98 gives a similar MAE of 0.068.
A reason for this could be that the gradients of the soft argmin at too low $\tau$ values are locally accurate, but almost zero if the current phase-composition predictions are far from the target, and hence there is a trade-off between local accuracy and global convergence.  
The single-system diagnostics in Figure~\ref{fig:single-system-diagnostics} show that the soft--hard composition discrepancy decreases during training, and Figures~\ref{fig:soft-hard-discrepancy-single-system} and~\ref{fig:soft-hard-discrepancy-generalization} quantify its final distribution for the single-system and held-out generalization studies, respectively.
A direct gradient-level comparison between the softmax gradient and local finite-difference estimates around the hard minimizer would be useful future work, but is outside the scope of the present study.

\begin{figure}[htbp]
    \centering
    \includegraphics[width=0.8\linewidth]{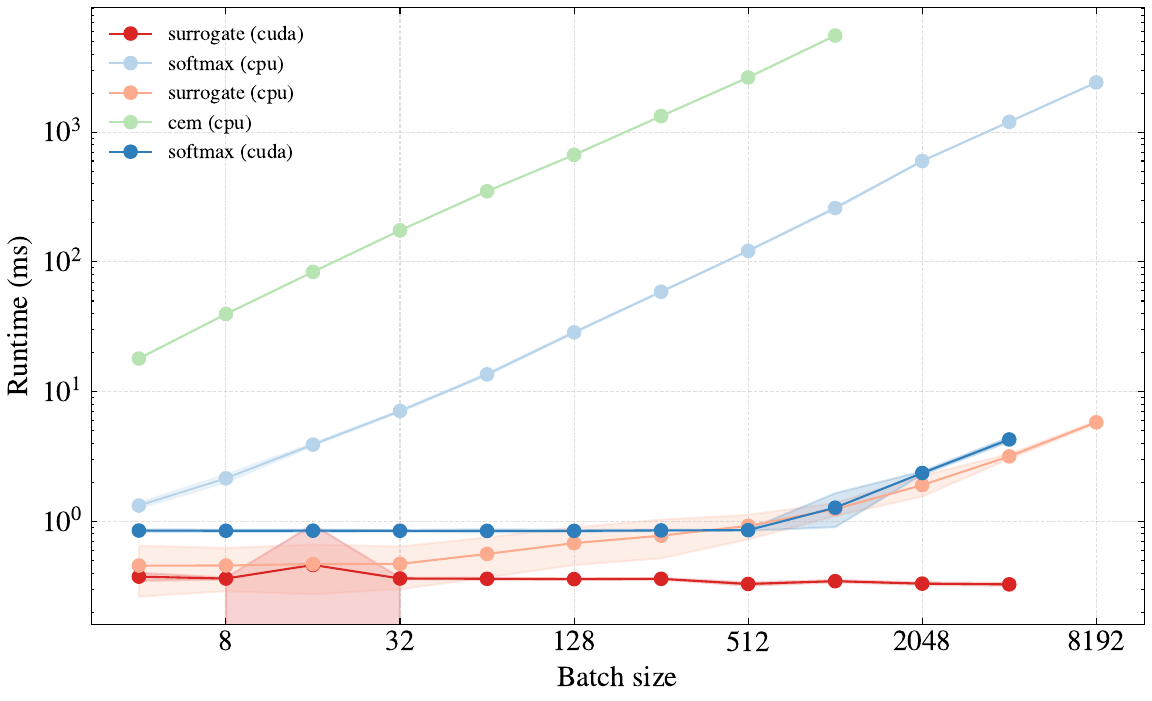}
    \caption{Run times of the respective algorithms to complete the full predictions for \gls{lle} based on a precomputed \gls{gmix} array of 101 entries. CEM is only shown for an approximate comparison and is separately computed for repeatedly solving a binary system from the corresponding work. The major differences in the CEM method make a fair comparison difficult.}
    \label{fig:runtimes}
\end{figure}
Figure~\ref{fig:runtimes} compares prediction-time costs for the different equilibrium solvers on precomputed \gls{gmix} arrays.
The DISCOMAX implementations remain competitive with the surrogate solver while preserving the thermodynamic structure of the discrete minimization problem.
The CEM is slower and not differentiable, but it is also capable of identifying the number and compositions of stable phases and has favorable scaling properties for multicomponent mixtures.
\clearpage
\ifdefined\arxivcombined
\else
\printbibliography
\fi

%% file: si/tables/hpo_grid_table.tex
\begin{table}[htbp]
\centering
\caption{Single system fitting hyperparameter grid search.}
\label{tab:hpo-grid}
\small
\textbf{Tested parameter sets}\\
\texttt{solver}: $\{$DISCOMAX, Surrogate$\}$\\
\texttt{lr}: $\{0.001, 0.005, 0.01\}$\\
\texttt{$\lambda_H$}: $\{0.01, 0.05, 0.1\}$\\
\texttt{$\lambda_G$}: $\{0.01, 0.05, 0.1\}$\\
\vspace{0.5em}
\begin{tabular}{lllll}
\toprule
solver & lr & $\lambda_H$ & $\lambda_G$ & MAE \\
\midrule
DISCOMAX & 0.005 & 0.05 & - & 0.0113 $\pm$ 0.0129 \\
DISCOMAX & 0.001 & 0.01 & - & 0.0113 $\pm$ 0.0164 \\
DISCOMAX & 0.005 & 0.01 & - & 0.0143 $\pm$ 0.0232 \\
DISCOMAX & 0.005 & 0.1 & - & 0.0150 $\pm$ 0.0201 \\
DISCOMAX & 0.01 & 0.1 & - & 0.0160 $\pm$ 0.0180 \\
DISCOMAX & 0.01 & 0.01 & - & 0.0161 $\pm$ 0.0345 \\
DISCOMAX & 0.001 & 0.05 & - & 0.0171 $\pm$ 0.0178 \\
DISCOMAX & 0.01 & 0.05 & - & 0.0182 $\pm$ 0.0223 \\
DISCOMAX & 0.001 & 0.1 & - & 0.0228 $\pm$ 0.0230 \\
Surrogate & 0.001 & - & 0.01 & 0.2527 $\pm$ 0.1527 \\
Surrogate & 0.001 & - & 0.05 & 0.2981 $\pm$ 0.1655 \\
Surrogate & 0.001 & - & 0.1 & 0.3246 $\pm$ 0.1825 \\
Surrogate & 0.01 & - & 0.01 & 0.3330 $\pm$ 0.1853 \\
Surrogate & 0.005 & - & 0.01 & 0.3405 $\pm$ 0.2099 \\
Surrogate & 0.005 & - & 0.05 & 0.3923 $\pm$ 0.2253 \\
Surrogate & 0.01 & - & 0.05 & 0.3980 $\pm$ 0.2281 \\
Surrogate & 0.01 & - & 0.1 & 0.4138 $\pm$ 0.2386 \\
Surrogate & 0.005 & - & 0.1 & 0.4180 $\pm$ 0.2302 \\
\bottomrule
\end{tabular}
\end{table}

%% file: si/tables/cv_metrics_table_all_methods_from_preds_gap-width.tex
\begin{table}[bpht]
\centering
\caption{Full cross-validation results for the gap-width metric on the test, validation, and train set, aggregated over ten folds for all model variants. Metrics are shown as mean $\pm$ standard deviation.}
\begin{tabular}{l l c c c}
\toprule
\textbf{Model} & \textbf{Split} & MAE & RMSE & \textbf{$R^{2}$} \\
\midrule

\multirow{3}{*}{DISCOMAX}
& Test & 0.068 $\pm$ 0.002 & 0.120 $\pm$ 0.005 & 0.803 $\pm$ 0.018 \\
& Validation & 0.068 $\pm$ 0.003 & 0.120 $\pm$ 0.006 & 0.804 $\pm$ 0.021 \\
& Train & 0.027 $\pm<$0.001 & 0.045 $\pm$ 0.001 & 0.972 $\pm$ 0.002 \\

\midrule

\multirow{3}{*}{DISCOMAX~G}
& Test & 0.064 $\pm$ 0.003 & 0.113 $\pm$ 0.006 & 0.825 $\pm$ 0.020 \\
& Validation & 0.065 $\pm$ 0.003 & 0.113 $\pm$ 0.006 & 0.828 $\pm$ 0.018 \\
& Train & 0.019 $\pm<$0.001 & 0.040 $\pm$ 0.002 & 0.978 $\pm$ 0.002 \\

\midrule

\multirow{3}{*}{DISCOMAX~H}
& Test & 0.062 $\pm$ 0.003 & 0.112 $\pm$ 0.006 & 0.829 $\pm$ 0.017 \\
& Validation & 0.063 $\pm$ 0.003 & 0.111 $\pm$ 0.006 & 0.832 $\pm$ 0.019 \\
& Train & 0.016 $\pm<$0.001 & 0.025 $\pm<$0.001 & 0.992 $\pm<$0.001 \\

\midrule

\multirow{3}{*}{DISCOMAX~H+G}
& Test & \textbf{0.062 $\pm$ 0.002} & \textbf{0.110 $\pm$ 0.006} & \textbf{0.834 $\pm$ 0.019} \\
& Validation & 0.064 $\pm$ 0.003 & 0.113 $\pm$ 0.007 & 0.827 $\pm$ 0.020 \\
& Train & 0.013 $\pm<$0.001 & 0.022 $\pm<$0.001 & 0.994 $\pm<$0.001 \\

\midrule

\multirow{3}{*}{Surrogate~G}
& Test & 0.069 $\pm$ 0.003 & 0.117 $\pm$ 0.005 & 0.814 $\pm$ 0.015 \\
& Validation & 0.069 $\pm$ 0.003 & 0.114 $\pm$ 0.005 & 0.824 $\pm$ 0.015 \\
& Train & 0.033 $\pm$ 0.001 & 0.056 $\pm$ 0.002 & 0.958 $\pm$ 0.003 \\

\midrule

\multirow{3}{*}{Surrogate~H}
& Test & 0.065 $\pm$ 0.002 & 0.111 $\pm$ 0.005 & 0.831 $\pm$ 0.017 \\
& Validation & 0.066 $\pm$ 0.003 & 0.111 $\pm$ 0.006 & 0.832 $\pm$ 0.019 \\
& Train & 0.021 $\pm<$0.001 & 0.030 $\pm<$0.001 & 0.988 $\pm<$0.001 \\

\midrule

\multirow{3}{*}{Surrogate~H+G}
& Test & 0.066 $\pm$ 0.003 & 0.114 $\pm$ 0.006 & 0.823 $\pm$ 0.020 \\
& Validation & 0.066 $\pm$ 0.003 & 0.112 $\pm$ 0.007 & 0.830 $\pm$ 0.017 \\
& Train & 0.020 $\pm<$0.001 & 0.030 $\pm$ 0.001 & 0.988 $\pm$ 0.001 \\
\bottomrule
\end{tabular}
\label{tab:cv-all-methods-gap-width}
\end{table}

%% file: si/tables/functional_group_counts_table.tex
\begin{table}[htbp]
    \caption{Common functional groups and the number of binary mixtures in which they are present.}
    \label{tab:fgs}
    \centering
        \begin{tabular}{lr}
        \toprule
        Functional Group & Count \\
        \midrule
        Alcohol (Hydroxy) & 4161 \\
        Halide & 3486 \\
        Ether & 2777 \\
        Ketone & 2569 \\
        Alkene & 2357 \\
        Amine & 2217 \\
        Aromatic ring (benzene) & 2094 \\
        Carboxylic ester & 1438 \\
        Nitrile & 1386 \\
        Amide & 899 \\
        Aldehyde & 782 \\
        Carboxylic acid & 481 \\
        Nitro & 301 \\
        Alkyne & 248 \\
        Thioether (Sulfide) & 241 \\
        Thiol & 117 \\
        \bottomrule
        \end{tabular}
\end{table}